\documentclass[10pt,twocolumn,letterpaper]{article}

\usepackage[pagenumbers]{cvpr} %

\def\cvprPaperID{928} %
\def\confName{CVPR}
\def\confYear{2022}

\usepackage[accsupp]{axessibility}  %

\usepackage[hyphens]{url}

\usepackage[toc,page]{appendix}

\usepackage[pagebackref,breaklinks,colorlinks]{hyperref}

\usepackage[dvipsnames]{xcolor}  %
\usepackage{colortbl}

\usepackage{graphicx}
\usepackage{amsmath}
\usepackage{amssymb}
\usepackage{booktabs}

\usepackage{multirow}

\newcommand{\qheading}[1]{\noindent\textbf{#1}:}

\newcommand{\zheading}[1]{\textbf{#1:}}

\usepackage{cuted}

\usepackage{bm}
\usepackage{xspace}

\usepackage{caption}
\usepackage{balance}

\usepackage{pifont}

\newcommand{\TODO}[1]{\xspace{\color{red} #1}\xspace}
\renewcommand{\TODO}[1]{\xspace{\color{black} #1}\xspace}

\newcommand{\lookat}[1]{\xspace{\color{black} #1}\xspace}

\usepackage[normalem]{ulem} %
\usepackage{arydshln}

\usepackage{lipsum}

\newcommand{\modelCOLOR}{black}
\newcommand{\modelname}{{\color{\modelCOLOR}SHAPY}\xspace}

\newcommand{\colorattr}{\color{PineGreen}}
\newcommand{\colorheight}{\color{Bittersweet}}
\newcommand{\colorweight}{\color{Orange}}
\newcommand{\colorcirc}{\color{OrangeRed}}

\newcommand{\heightVar}[0]{{\colorheight{H}}}
\newcommand{\attrVar}[0]{{\colorattr{A}}}

\newcommand{\circVar}[0]{{\colorcirc{C}}}

\newcommand{\colorAFMtoS}{\text{\mbox{{\colorattr{A}}{\colorheight{H}}{\colorweight{W}}{\colorcirc{C}}2S}}\xspace}
\newcommand{\colorAHWCtoS}{\text{\mbox{{\colorattr{A}}{\colorheight{H}}{\colorweight{W}}{\colorcirc{C}}}}\xspace}
\newcommand{\colorAHWC}{\text{\mbox{{\colorattr{A}}{\colorheight{H}}{\colorweight{W}}{\colorcirc{C}}2S}}\xspace}

\newcommand{\colorAHWtoS}{\text{\mbox{{\colorattr{A}}{\colorheight{H}}{\colorweight{W}}2S}}\xspace}
\newcommand{\colorAHCtoS}{\text{\mbox{{\colorattr{A}}{\colorheight{H}}{\colorcirc{C}}2S}}\xspace}

\newcommand{\colorAHtoS}{\text{\mbox{{\colorattr{A}}{\colorheight{H}}2S}}\xspace}
\newcommand{\colorAH}{\text{\mbox{{\colorattr{A}}{\colorheight{H}}}}\xspace}
\newcommand{\colorACtoS}{\text{\mbox{{\colorattr{A}}{\colorcirc{C}}2S}}\xspace}

\newcommand{\colorFMtoS}{\text{\mbox{{\colorheight{H}}{\colorweight{W}}{\colorcirc{C}}2S}}\xspace}
\newcommand{\colorHWtoS}{\text{\mbox{{\colorheight{H}}{\colorweight{W}}2S}}\xspace}
\newcommand{\colorHCtoS}{\text{\mbox{{\colorheight{H}}{\colorcirc{C}}2S}}\xspace}
\newcommand{\colorHtoS}{\text{\mbox{{\colorheight{H}}2S}}\xspace}
\newcommand{\colorCtoS}{\text{\mbox{{\colorcirc{C}}2S}}\xspace}

\newcommand{\vcaliper}{``Virtual Caliper''\xspace}
\newcommand{\virtualcaliper}{\vcaliper}

\newcommand{\vanilla}{\mbox{base}\xspace}

\newcommand{\StoM}{VM\xspace}
\newcommand{\AtoS}{\text{\mbox{A2S}}\xspace}

\newcommand{\ItoA}{\mbox{I2A}\xspace}

\newcommand{\polynomialweights}{W}

\newcommand{\numattr}{K}
\newcommand{\lineartarget}{\mathbf{Y}}

\newcommand{\StoA}{\text{\mbox{S2A}}\xspace}

\newcommand{\HtoS}{\text{\mbox{H2S}}\xspace}
\newcommand{\CtoS}{\text{\mbox{C2S}}\xspace}
\newcommand{\HWtoS}{\text{\mbox{HW2S}}\xspace}
\newcommand{\HCtoS}{\text{\mbox{HC2S}}\xspace}
\newcommand{\HWCtoS}{\text{\mbox{HWC2S}}\xspace}

\newcommand{\AHtoS}{\text{\mbox{AH2S}}\xspace}
\newcommand{\ACtoS}{\text{\mbox{AC2S}}\xspace}
\newcommand{\AHWtoS}{\text{\mbox{AHW2S}}\xspace}
\newcommand{\AHCtoS}{\text{\mbox{AHC2S}}\xspace}

\newcommand{\AHWCtoS}{\text{\mbox{AHWC2S}}\xspace}
\newcommand{\AFMtoS}{\text{\mbox{AHWC2S}}\xspace}

\newcommand{\shapyH}{\mbox{\modelname-{\colorheight{H}}}\xspace}
\newcommand{\shapyA}{\mbox{\modelname-{\colorattr{A}}}\xspace}
\newcommand{\shapyHA}{\mbox{\modelname-{\colorheight{H}\colorattr{A}}}\xspace}

\newcommand{\websiteURL}{\mbox{\href{https://shapy.is.tue.mpg.de}{shapy.is.tue.mpg.de}}}

\newcommand{\pixie}{\mbox{PIXIE}\xspace}
\newcommand{\expose}{\mbox{ExPose}\xspace}

\newcommand{\straps}{\mbox{STRAPS}\xspace}
\newcommand{\tuch}{\mbox{TUCH}\xspace}
\newcommand{\hybrik}{\mbox{HybrIK}\xspace}

\newcommand{\retinanet}{\mbox{RetinaNet}\xspace}
\newcommand{\arcface}{\mbox{ArcFace}\xspace}

\newcommand{\tpose}{\mbox{T-pose}\xspace}
\newcommand{\apose}{\mbox{A-pose}\xspace}
\newcommand{\twoD}{2D\xspace}

\newcommand{\threeD}{3D\xspace}

\newcommand{\mmts}{MMTS\xspace}
\newcommand{\cmts}{CMTS\xspace}

\newcommand{\ssp}{\mbox{SSP-\threeD}\xspace}
\newcommand{\ssplong}{\mbox{Sports Shape and Pose \threeD}\xspace}

\newcommand{\ourTitle}{Accurate 3D Body Shape Regression using Metric and Semantic Attributes}

\newcommand{\vspaceTABaboveCaption}{-0.0 em}

\newcommand{\hps}{\mbox{HPS}\xspace}
\newcommand{\HPS}{\hps}

\newcommand{\hbw}{\mbox{HBW}\xspace}
\newcommand{\hbwLONG}{{``Human Bodies in the Wild''}\xspace}

\newcommand{\colorNUMB}{black}
\newcommand{\hbwNumberSubjects}{{\color{\colorNUMB}35}\xspace}
\newcommand{\hbwNumberSubjectsM}{{\color{\colorNUMB}15}\xspace}
\newcommand{\hbwNumberSubjectsF}{{\color{\colorNUMB}20}\xspace}
\newcommand{\hbwNumberPhotos}{{\color{\colorNUMB}2543}\xspace}
\newcommand{\hbwNumberPhotosLab}{{\color{\colorNUMB}1,318}\xspace}
\newcommand{\hbwNumberPhotosWild}{{\color{\colorNUMB}1,225}\xspace}
\newcommand{\hbwNumberPhotosLabTHRESHperSubj}{{\color{\colorNUMB}111}\xspace}
\newcommand{\hbwNumberPhotosWildTHRESHperSubj}{{\color{\colorNUMB}126}\xspace}
\newcommand{\hbwNumberSPLITSsubj}{{\color{\colorNUMB}{10/25 subjects (6/14 female 4/11 male)}}\xspace}
\newcommand{\hbwNumberSPLITSimg}{{\color{\colorNUMB}{781/1,762 images (432/983 female 349/779 male)}}\xspace}

\newcommand{\bodytalk}{\mbox{BodyTalk}\xspace}

\newcommand{\bmi}{\mbox{BMI}\xspace}

\newcommand{\hmr}{\mbox{HMR}\xspace}
\newcommand{\scape}{\mbox{SCAPE}\xspace}
\newcommand{\spin}{\mbox{SPIN}\xspace}
\newcommand{\smplx}{\mbox{SMPL-X}\xspace}
\newcommand{\smplX}{\smplx}
\newcommand{\smplr}{{\smpl}R\xspace}

\newcommand{\groundtruth}{\mbox{ground-truth}\xspace}

\newcommand{\amt}{\mbox{AMT}\xspace}

\newcommand{\coco}{\mbox{COCO}\xspace}

\newcommand{\ghum}{\mbox{GHUM}\xspace}

\newcommand{\smpl}{\mbox{SMPL}\xspace}

\newcommand{\caesar}{\mbox{CAESAR}\xspace}
\newcommand{\resnet}{\mbox{ResNet50}\xspace}

\newcommand{\mocap}{\mbox{MoCap}\xspace}

\newcommand{\inthewild}{\mbox{in-the-wild}\xspace}

\newcommand{\nvertices}{10,475\xspace}
\newcommand{\nverticessmpl}{6,890\xspace}

\newcommand{\npointshd}{20k\xspace}

\newcommand{\supmat}{{\mbox{\textcolor{black}{Sup.~Mat.}}}\xspace}

\newcommand{\threedpw}{\mbox{3DPW}\xspace}

\newcommand{\mesh}{M}
\newcommand{\shape}{\bm{\beta}}

\newcommand{\pose}{\bm{\theta}}
\newcommand{\expression}{\bm{\psi}}

\renewcommand{\etal}{\mbox{et al.}\xspace}
\renewcommand{\ie}{\mbox{i.e.}\xspace}
\renewcommand{\eg}{\mbox{e.g.}\xspace}
\newcommand{\na}{n / a}

\newcommand{\colorRef}[1]{\textcolor{black}{#1}} %
\usepackage[capitalize]{cleveref}
\crefname{figure}{\colorRef{Fig.}}{\colorRef{Figs.}}
\Crefname{figure}{\colorRef{Figure}}{\colorRef{Figures}}
\crefname{section}{\colorRef{Sec.}}{\colorRef{Secs.}}
\Crefname{section}{\colorRef{Section}}{\colorRef{Sections}}
\Crefname{table}{\colorRef{Table}}{\colorRef{Tables}}
\crefname{table}{\colorRef{Tab.}}{\colorRef{Tabs.}}

\newcommand{\norm}[1]{\left\lVert#1\right\rVert}

\DeclareSymbolFont{matha}{OML}{txmi}{m}{it}%
\DeclareMathSymbol{\varv}{\mathord}{matha}{118}
\newcommand{\vtov}{\mbox{V2V}\xspace}
\newcommand{\vtovlong}{Vertex-to-Vertex\xspace}
\newcommand{\VtoV}{\vtov}

\newcommand{\vtovHD}{\mbox{$\text{P2P}_{20\text{K}}$}\xspace}

\newcommand{\mpjpe}{\mbox{MPJPE}\xspace}

\newcommand{\rgb}{\mbox{RGB}\xspace}

\newcommand{\height}{H}
\newcommand{\weight}{W}
\newcommand{\heightArg}{\height(\shape)}
\newcommand{\weightArg}{\weight(\shape)}
\newcommand{\chestCirc}{C_{\text{c}}}
\newcommand{\waistCirc}{C_{\text{w}}}
\newcommand{\hipsCirc}{C_{\text{h}}}
\newcommand{\chestCircArg}{\chestCirc(\shape)}
\newcommand{\waistCircArg}{\waistCirc(\shape)}
\newcommand{\hipsCircArg}{\hipsCirc(\shape)}

\newcommand{\sparseregressor}{\mathbf{H}}
\newcommand{\sparseregressorsmplx}{\sparseregressor_{\text{\smplx}}}
\newcommand{\sparseregressorsmpl}{\sparseregressor_{\text{\smpl}}}

\newcommand{\predattr}{A}

\newcommand{\cameraready}[1]{\textcolor{Fuchsia}{{#1}}\xspace}
\renewcommand{\cameraready}[1]{\textcolor{black}{{#1}}\xspace}

\newcommand{\myarraystretch}[0]{1.1}

\newcommand{\colorTERM}{blue}
\renewcommand{\colorTERM}{black}
\newcommand{\measurement}[0]{{\color{\colorTERM}anthropometric measurement}\xspace}
\newcommand{\measurements}[0]{{\color{\colorTERM}anthropometric measurements}\xspace}
\newcommand{\Measurements}[0]{{\color{\colorTERM}Anthropometric Measurements}\xspace}

\newcommand{\semanticshapeattribute}[0]{{\color{\colorTERM}semantic shape attribute}\xspace}
\newcommand{\semanticshapeattributes}[0]{{\color{\colorTERM}semantic shape attributes}\xspace}
\renewcommand{\semanticshapeattribute}[0]{{\color{\colorTERM}linguistic shape attribute}\xspace}
\renewcommand{\semanticshapeattributes}[0]{{\color{\colorTERM}linguistic shape attributes}\xspace}

\newcommand{\linguisticshapeattribute}[0]{{\color{\colorTERM}linguistic shape attribute}\xspace}
\newcommand{\linguisticattributes}[0]{{\color{\colorTERM}linguistic attributes}\xspace}
\newcommand{\linguisticshapeattributes}[0]{{\color{\colorTERM}linguistic shape attributes}\xspace}
\newcommand{\Linguisticshapeattributes}[0]{{\color{\colorTERM}Linguistic shape attributes}\xspace}

\newcommand{\LinguisticShapeAttributes}[0]{{\color{\colorTERM}Linguistic Shape Attributes}\xspace}

\newcommand{\scores}[0]{{scores}\xspace}

\newcommand{\linguisticattributescores}[0]{{\color{\colorTERM}linguistic attribute \scores}\xspace}
\newcommand{\linguisticshapeattributescores}[0]{{\color{\colorTERM}linguistic shape attribute \scores}\xspace}
\newcommand{\hulledgeindices}[0]{\mathcal{E}}

\begin{document}

\title{\ourTitle}

\author{%
Vasileios Choutas\textsuperscript{*1},
Lea M{\"u}ller\textsuperscript{*1},
Chun-Hao P. Huang\textsuperscript{1},
Siyu Tang\textsuperscript{2},
Dimitrios Tzionas\textsuperscript{1},
Michael J. Black\textsuperscript{1}\\
\textsuperscript{1}Max Planck Institute for Intelligent Systems, T{\"u}bingen, Germany
\quad
\textsuperscript{2}ETH Z{\"u}rich
\\
{\tt\small \{vchoutas, lea.mueller, paul.huang, stang, dtzionas, black\}@tuebingen.mpg.de}\\
{\footnotesize * Equal contribution, alphabetical order}
\vspace{-0.7em}
}

\newcommand{\teaserCaption}{
    Existing work 
    on \threeD human reconstruction from a \cameraready{color} image focuses mainly on \emph{pose}. %
    We present \modelname, a model that focuses on body \emph{shape} and learns to predict \cameraready{dense} \threeD 
    \cameraready{shape}
    from \cameraready{a color image, using} crowd-sourced 
    \emph{\linguisticshapeattributes}. 
    Even with this weak supervision, \modelname %
    \cameraready{outperforms} 
    the state of the art \cameraready{(SOTA)} %
    \cite{sengupta2021hierarchicalICCV} 
    on in-the-wild images with varied clothing.
}

\twocolumn[{
    \renewcommand\twocolumn[1][]{#1}
    \maketitle
    \centering
    \vspace{-1.0 em}
    \begin{minipage}{1.00\textwidth}
        \centering
        \includegraphics[width=1.00 \linewidth]{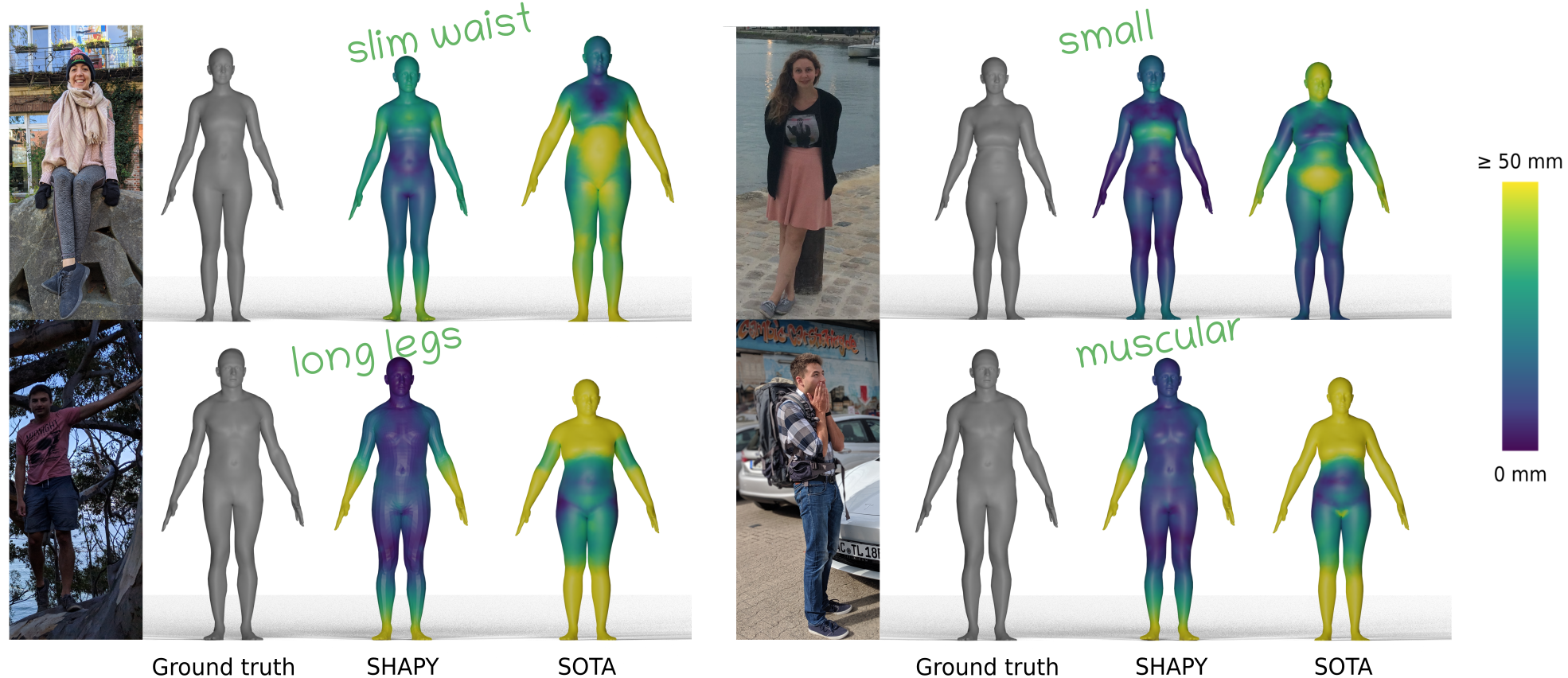}
    \end{minipage}
    \vspace{-0.3 em}
    \captionof{figure}{\teaserCaption}
    \label{fig:teaser}
    \vspace*{+01.85em}
}]

\begin{abstract}
\vspace{-1.0 em}
While methods that regress \threeD human meshes from images have progressed rapidly, the estimated body shapes often do not capture the true human shape. 
This is problematic since, for many applications, accurate body shape is as important as  pose. 
The key reason that body shape accuracy lags pose accuracy is the lack of data. 
While humans can label \twoD joints, and these constrain \threeD pose, it is not so easy to ``label" \threeD body shape. 
Since paired data with images and \threeD body shape are rare, we exploit two sources of
information: 
(1) we collect internet images of diverse \cameraready{``fashion'' models} together with a small set of 
\measurements;
(2) we collect 
\semanticshapeattributes 
for a wide range of  \threeD body meshes and the model images. 
Taken together, these datasets provide sufficient constraints to infer 
\cameraready{dense \threeD shape}.
We exploit
\cameraready{the 
\measurements 
and \linguisticshapeattributes}
in several novel ways to train a neural network, called \modelname, that regresses \threeD human pose and shape from an \rgb image.
We evaluate \modelname on public benchmarks, but note that they either lack significant body shape variation, ground-truth shape, or clothing variation.
Thus, we collect a new dataset for \cameraready{evaluating} \threeD human shape estimation, \cameraready{called \hbw}, containing photos of 
\cameraready{\hbwLONG} 
for \cameraready{which} we have \groundtruth \threeD body scans. 
On this new benchmark, \modelname significantly outperforms %
state-of-the-art methods on the task of \threeD body shape estimation. 
This is the first demonstration \cameraready{that \threeD body shape regression from images} can be trained from 
\cameraready{easy-to-obtain} 
\measurements 
and  
\semanticshapeattributes.
\cameraready{Our model and data are available at: \websiteURL} %
\end{abstract}

\section{Introduction}      \label{sec:intro}

The field of \threeD human pose and shape (\HPS) estimation is progressing rapidly and methods now regress accurate \threeD pose from a single image \cite{bogo2016keep,Joo2018_adam,kanazawa_2019_cvpr,VIBE:CVPR:2020,Kolotouros2019_spin,Pavlakos2019_smplifyx,xu2020ghum,pare,spec,pymaf}.
Unfortunately, less attention has been paid to body shape and many methods produce body shapes that clearly do not represent the person in the image (\cref{fig:teaser}, top right).
There are several reasons behind this.
Current evaluation datasets focus on pose and not shape. Training datasets of images with \threeD ground-truth shape are lacking.
Additionally, humans appear in images wearing clothing that obscures the body, making the problem challenging.
Finally, the fundamental scale ambiguity in \twoD images, makes 
\cameraready{\threeD shape} 
difficult to estimate.
For many applications, however, realistic body shape is critical.
These include AR/VR, apparel design, virtual try-on, and fitness.
To democratize avatars, it is important to represent and estimate all possible \threeD body shapes; \cameraready{we make a step in that direction.}

Note that commercial solutions to this problem require users to wear tight fitting clothing and capture multiple images or a video sequence using constrained poses.
In contrast, we tackle the unconstrained problem of \threeD body shape estimation in the wild from a single RGB image of a person in an arbitrary pose and standard clothing.

Most current approaches to \HPS estimation learn to regress a parametric \threeD body model like \smpl \cite{SMPL:2015} from images using \twoD joint locations as training data. 
Such joint locations are easy for human annotators to label in images.
\cameraready{Supervising} the training with joints, however, is not sufficient to learn shape since an infinite number of body shapes can share the same joints.
For example, consider someone who puts on weight.  Their body shape changes but their joints stay the same.
Several recent \cameraready{methods} employ additional \twoD cues, such as the silhouette, to provide additional shape cues \cite{sengupta2020straps, sengupta2021hierarchicalICCV}.
Silhouettes, however, are influenced by clothing and do not \TODO{provide} explicit \threeD supervision.
Synthetic approaches \cite{Liang_2019_ICCV}, on the other hand, drape SMPL \threeD bodies in virtual clothing and render them in images.
While this provides ground-truth \threeD shape, realistic synthesis of clothed humans is %
challenging, resulting in a domain gap.

To address these issues, we present \modelname, a new deep neural network that accurately regresses \threeD body shape and pose from a single \rgb image. 
To train \modelname, we first need to address the lack of paired training data with real images and ground-truth shape.
Without access to such data, we need alternatives that are easier to acquire, analogous to \twoD joints used in pose estimation.
To do so, we introduce two novel datasets and corresponding 
training methods.

\begin{figure}
    \centering
    \includegraphics[trim=000mm 000mm 000mm 000mm, clip=false, width=1.00 \linewidth]{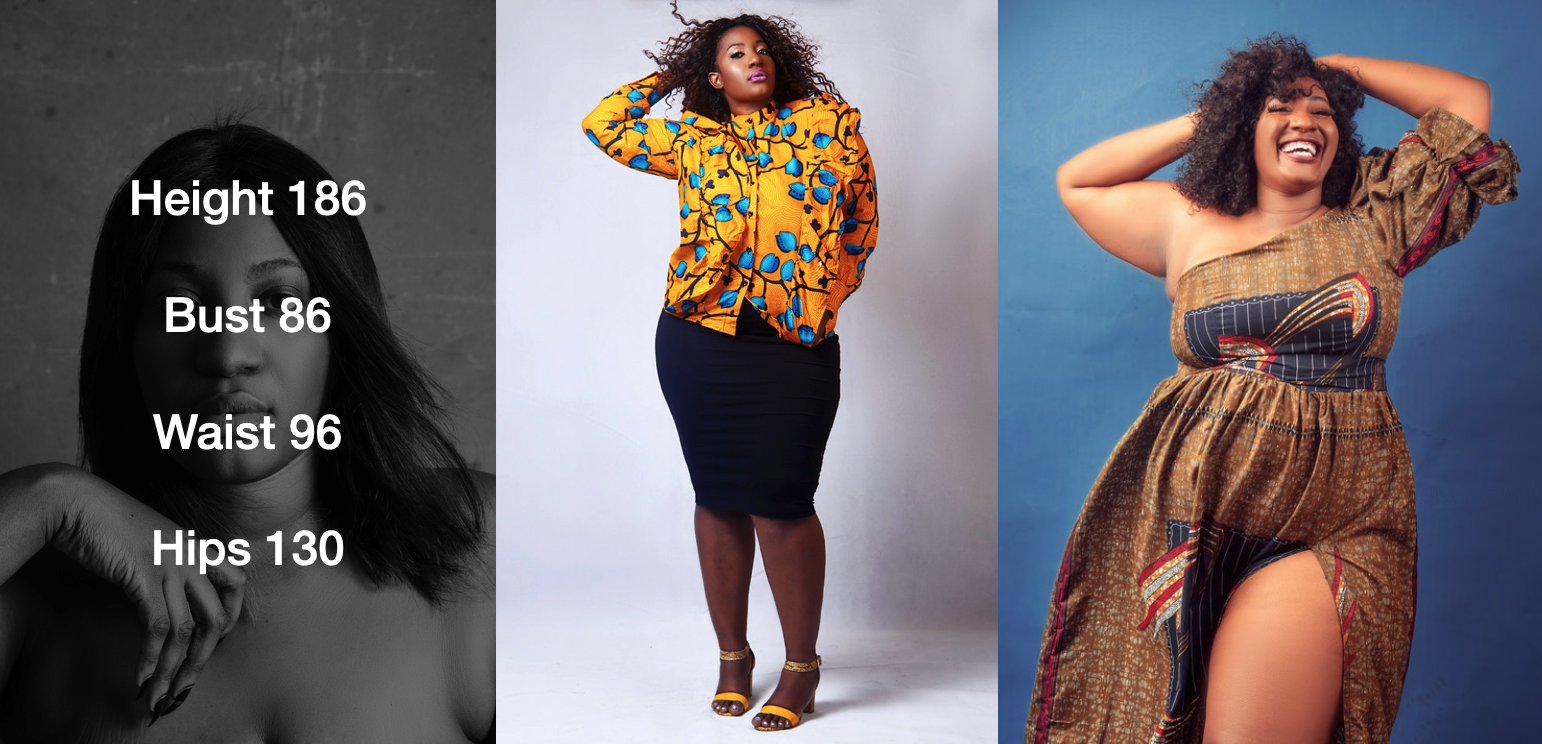}
    \caption{%
    \cameraready{%
        Model-agency websites
        contain multiple images of models together with %
        \measurements.  
        A wide range of body shapes are represented;
        example from
        \href{https://www.pexels.com/photo/calm-black-plus-size-woman-against-gray-wall-4986688}{\mbox{pexels.com}}.
        }
    }
    \label{fig:data_eg_metrics}
    \vspace{-0.5em}
\end{figure}

First, in lieu of full \threeD body scans, we use images of people with diverse body shapes for which we have 
\measurements
such as height \cameraready{as well as} chest, waist, and hip circumference.
While many \threeD human shapes can share the same measurements, they do constrain the space of possible shapes.
Additionally, these are important measurements for applications in clothing and health.
Accurate \measurements like these are difficult for individuals to take themselves but they are often captured for different applications.
Specifically, modeling agencies provide such information about their models; accuracy is a requirement for modeling clothing.
Thus, 
we collect a diverse set of such model images (with varied ethnicity, clothing, and body shape) with associated measurements; see \cref{fig:data_eg_metrics}.

\begin{figure}
    \centering
    \includegraphics[trim=000mm 000mm 000mm 000mm, clip=false, width=0.98 \linewidth]{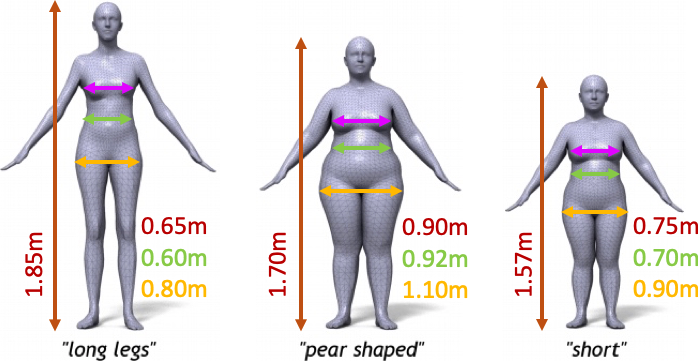}
    \caption{
        We crowd-source scores for \TODO{linguistic} 
        body-shape attributes \cite{Streuber:SIGGRAPH:2016} 
        \cameraready{and compute \measurements for  \caesar \cite{CAESAR} body meshes.}
        We also crowd-source 
        \linguisticshapeattributescores
        for model images, like those in Fig.~\ref{fig:data_eg_metrics} 
    }
    \label{fig:data_eg_words}
    \vspace{-0.50em}
\end{figure}

Since sparse \measurements do not fully constrain body shape, we exploit a novel approach and also use \emph{\semanticshapeattributes}.
Prior work has shown that people can rate images of others according to shape attributes such as ``short/tall'', ``long legs'' or ``pear shaped'' \cite{Streuber:SIGGRAPH:2016}; \cameraready{see}  \cref{fig:data_eg_words}.
Using the average \scores from several raters, Streuber et al.~\cite{Streuber:SIGGRAPH:2016} 
(\bodytalk) 
regress metrically accurate \threeD body shape.
This approach gives us a way to easily label images of people and use these labels to constrain \threeD shape.
To our knowledge, this sort of \semanticshapeattribute data has not previously been exploited to train a neural network to infer \cameraready{\threeD} body shape \cameraready{from images}.

We exploit these new datasets to train 
\modelname
with three novel \emph{losses}, 
which can be exploited by any \threeD human body reconstruction method:
(1) We define functions of the \smpl body mesh that return
a sparse set of \TODO{\measurements}.
When measurements are available for an image we use a loss that penalizes \cameraready{mesh measurements that differ from the \groundtruth (GT). } 
(2)     We learn a ``Shape to Attribute'' (\StoA) function that maps \threeD bodies to 
        \linguisticattributescores. 
        During training, we map meshes to attribute \scores and penalize differences from the GT \scores.
(3)     We similarly learn a function that maps ``Attributes to Shape'' (\AtoS).
We then penalize body shape parameters that deviate from the prediction.

We study each %
term
in detail to arrive at the final method.
Evaluation is challenging because existing benchmarks with GT shape either contain too few subjects \cite{vonMarcard2018} or
have limited clothing complexity and only pseudo-GT shape
\cite{sengupta2020straps}.
We fill this gap with a new dataset, named \hbwLONG (\hbw), that contains a ground-truth \threeD body scan and several \inthewild photos of \hbwNumberSubjects subjects, for a total of \hbwNumberPhotos photos.
Evaluation on this shows that \modelname estimates much more accurate \threeD shape. 

\smallskip
\noindent
\TODO{Models}, data and code are available at \websiteURL.

\section{Related Work}

\zheading{\threeD human pose and shape (HPS)} %
Methods that %
reconstruct \threeD human bodies from one or more \rgb images can be split into two broad categories: 
(1) \textbf{parametric methods} that %
\cameraready{predict parameters of}
a statistical \threeD body model, such as \scape \cite{anguelov2005scape}, \smpl~\cite{SMPL:2015}, \smplX~\cite{Pavlakos2019_smplifyx}, Adam \cite{Joo2018_adam}, \ghum \cite{xu2020ghum}, 
and
(2) \textbf{non-parametric methods} that predict a free-form representation of the human body
\cameraready{\cite{varol2018bodynet,saito2020pifuhd,Jafarian_2021_CVPR,xiu2022icon}.}
Parametric approaches lack \cameraready{details \wrt non-parametric ones, \eg, clothing or hair.} %
However, \cameraready{parametric models}~
\cameraready{disentangle the effects of identity and pose on the overall shape}.
\cameraready{Therefore, their parameters provide control for re-shaping and re-posing}. 
\cameraready{Moreover, pose can be factored out to bring meshes in a canonical pose; this is important for evaluating estimates of an 
individual's shape.} 
Finally, since
topology
\cameraready{is fixed}, meshes can be compared easily. 
\cameraready{For these reasons,} 
we use a \smplx body model.

\cameraready{Parametric methods
follow two main paradigms,} and are based on optimization or regression. 
\textbf{Optimization-based methods} \cite{balan2007detailed,bogo2016keep,guan_iccv_scape_2009, Pavlakos2019_smplifyx} search for model configurations that best explain image evidence, usually \twoD landmarks \cite{OpenPose_PAMI}, subject to model priors that \TODO{usually encourage parameters to be close to the mean of the model space.} 
Numerous methods penalize %
\cameraready{the discrepancy} 
between the projected and \groundtruth silhouettes \cite{MuVS_3DV_2017,lassner2017unite} to 
\cameraready{estimate} 
shape.
\cameraready{However,}
this needs special care to handle clothing \cite{Balan:ECCV}; 
without this, erroneous solutions \cameraready{emerge that} 
\TODO{``inflate'' %
body shape to explain the ``clothed'' silhouette.} 
\textbf{Regression-based methods} %
\cite{Choutas2020_expose,georgakis2020hierarchical,jiang2020multiperson,Kanazawa2018_hmr,Kolotouros2019_spin,Liang_2019_ICCV,VIBE:CVPR:2020,mueller2021tuch,zanfir2020weakly} 
\cameraready{are currently based on} deep neural networks that directly regress model parameters from image pixels.
Their training sets are a mixture of data captured in laboratory settings \cite{ionescu2013human36m,sigal_ijcv_10b}, with model parameters estimated from \mocap markers \cite{AMASS:ICCV:2019}, and \inthewild image collections, such as \coco \cite{lin2014coco},  that contain \twoD keypoint annotations. 
Optimization and regression can be combined, for example via in-the-\cameraready{network} model fitting \cite{Kolotouros2019_spin,mueller2021tuch}.

\zheading{Estimating \threeD body shape}
\cameraready{State-of-the-art methods} are effective for estimating \threeD pose, but \emph{struggle} with estimating \emph{body shape} under clothing. 
\TODO{There are several reasons for this.} 
\cameraready{First}, \twoD keypoints alone are not \cameraready{sufficient to fully constrain} \threeD body shape.
\cameraready{Second, shape priors address the lack of constraints}, but \TODO{bias solutions towards ``average'' shapes} \cite{bogo2016keep,Pavlakos2019_smplifyx,Kolotouros2019_spin,mueller2021tuch}. 
\cameraready{Third, datasets with \inthewild images have noisy \threeD bodies, recovered by} fitting  a model to \twoD keypoints \cite{bogo2016keep,Pavlakos2019_smplifyx}. 
\cameraready{Fourth, datasets captured in laboratory settings} have a small number of subjects, who do not represent the full spectrum of body shapes. 
\cameraready{Thus, there is a scarcity of images with known, \emph{accurate}, \threeD body shape.} 
Existing methods deal with this in two ways.

First, rendering 
\mbox{\emph{synthetic images}} is attractive since it gives automatic and precise \groundtruth annotation. 
This 
\cameraready{involves}
shaping, posing, dressing and texturing a \threeD body model \cite{Hoffmann:GCPR:2019,sengupta2020straps,sengupta2021probabilisticCVPR,varol17_surreal,weitz2021infiniteform}, \cameraready{then lighting it and rendering it in a scene}. 
Doing this realistically and with natural clothing is %
expensive, hence, current datasets suffer from a domain gap.
Alternative methods use artist-curated \threeD scans \cite{saito2019pifu,saito2020pifuhd,patel2020agora}, which are realistic but limited in %
variety.

Second,
\mbox{\emph{\twoD shape cues}} for \inthewild images, %
(body-part segmentation masks     \cite{omran2018neural,Ruegg:AAAI:2020,sai2021dsr}, silhouettes  \cite{agarwal_trigs_3d_poses,MuVS_3DV_2017,pavlakos2018learning})
are attractive, as these can be manually annotated or automatically detected \cite{gong2019graphonomy,He2020maskRCNN}. %
However, fitting to such %
\cameraready{cues}
often gives unrealistic body shapes, \TODO{by inflating the body to ``explain'' the clothing} ``baked'' into silhouettes and masks. 
 
Most related to our work is the work of Sengupta \etal \cite{sengupta2020straps,sengupta2021probabilisticCVPR,sengupta2021hierarchicalICCV}
who estimate body shape
using a probabilistic learning approach,
trained on \TODO{edge-filtered} synthetic images.
They evaluate on the \ssp dataset of real images with pseudo-GT \threeD bodies,
\cameraready{estimated by fitting \smpl to multiple video frames.}
\ssp is biased to people with tight-fitting clothing.
Their silhouette-based method works well on \ssp but does not generalize
to people in normal clothing, 
\cameraready{tending to over-estimate body shape; see \cref{fig:teaser}.} 
 
In contrast to previous work, \modelname is trained with \inthewild images paired with \linguisticshapeattributes, which are 
\cameraready{annotations} 
that can be easily crowd-sourced for \cameraready{weak} shape supervision.
We also go beyond \ssp to provide \hbw, a new dataset with %
in-the-wild images, varied clothing, and precise GT from \threeD scans.

\zheading{Shape, measurements and attributes}
Body shapes can be generated 
\cameraready{from}
\measurements
\cite{allen2003space,seo2003synthesizing,seo2003automatic}.
Tsoli \etal \cite{tsoliWACV14} register a body model to multiple high-resolution body scans to extract body measurements. 
The \virtualcaliper \cite{pujades2019virtual} allows users to build metrically accurate avatars of themselves using \cameraready{measurements or} VR game controllers.
\cameraready{ViBE} \cite{hsiao2020vibe} collects images, measurements (bust, waist, hip circumference, height) and the dress-size of models
from clothing websites to train a clothing recommendation network.
We draw inspiration from 
\cameraready{these approaches} 
for data collection and supervision.

Streuber \etal \cite{Streuber:SIGGRAPH:2016} \cameraready{learn \bodytalk}, a model that generates
\threeD
body shapes from \linguisticattributes.
\cameraready{For this, they select
attributes that describe human shape and
ask annotators to rate how much each attribute applies
to a body.}
They fit a linear model that maps attribute
ratings to \smpl shape parameters. 
Inspired by this, we collect attribute ratings for \caesar \cameraready{meshes \cite{CAESAR}}  and \inthewild data %
as proxy shape supervision to train a HPS regressor.
\cameraready{Unlike \bodytalk,
\modelname  automatically infers shape from images.}

\zheading{Anthropometry from images} 
Single-View metrology \cite{criminisi2000single}
estimates the height of a person in an image, using horizontal and vertical vanishing points and the height of a reference object.
G{\"u}nel \etal~\cite{gunel2019face} introduce the IMDB-23K dataset by gathering publicly available celebrity images and their height information.
Zhu \etal~\cite{SingleViewMetrology} use this dataset to learn to predict the height of people in images.
Dey \etal~\cite{Ratan2014} estimate the height of users in a photo collection by computing height differences between people in an image, creating a graph that links people across photos, and solving a maximum likelihood estimation problem. %
Bieler \etal \cite{Bieler_2019_ICCV} use gravity as a prior to convert pixel measurements extracted from a video to metric height.
\cameraready{These methods do not address body shape.}

\begin{figure}[!t]
    \centering
    \includegraphics[trim=000mm 000mm 000mm 000mm, clip=true, width=1.00\linewidth]{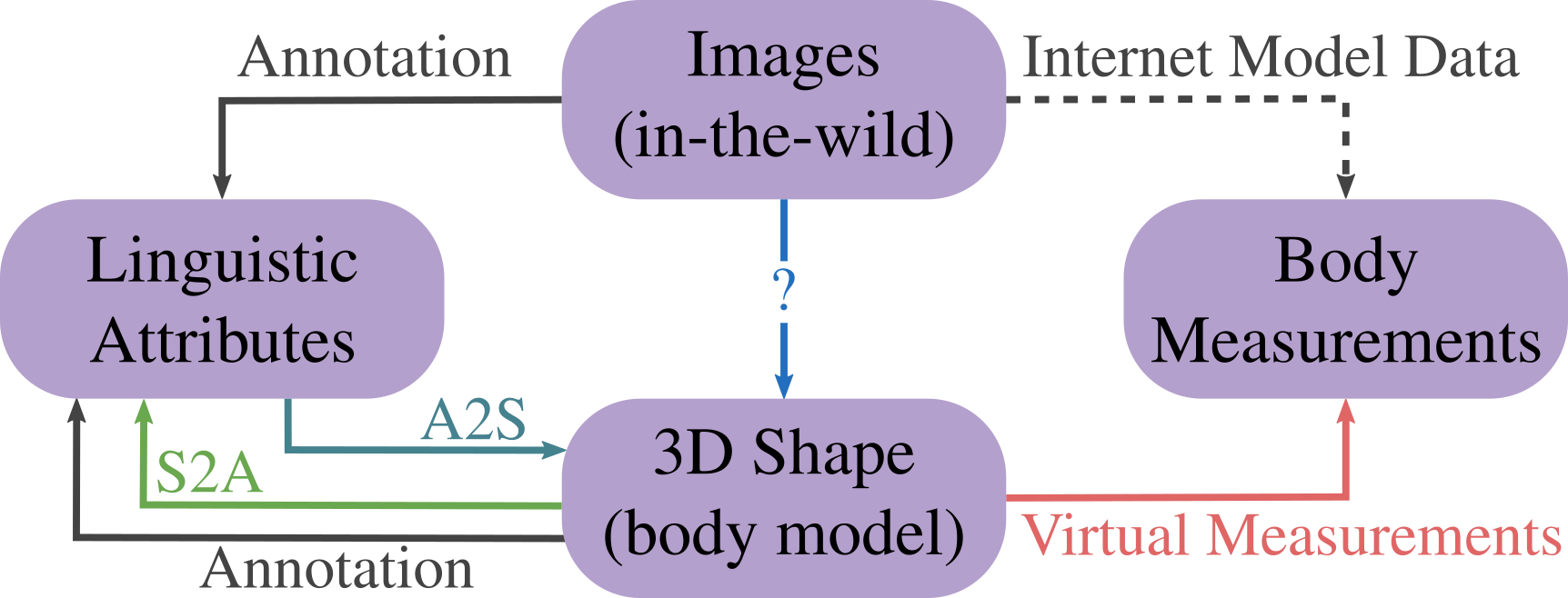}
    \caption{%
        \cameraready{%
        \TODO{Shape representations and data collection}. 
        Our goal is \threeD body shape estimation from \inthewild images. 
        Collecting data for direct supervision is difficult and does not scale.
        We explore two alternatives. %
        \qheading{\LinguisticShapeAttributes} 
        We annotate attributes (``A'') for
        \caesar meshes, for which we have accurate shape (``S'') parameters, and learn the 
        ``\AtoS'' and ``\StoA'' models, to map between these representations.
        Attribute annotations for %
        images can be easily crowd-sourced, making these scalable.
        \qheading{\Measurements} 
        We collect images with sparse body measurements
        from model-agency websites.
        A virtual measurement module \cite{pujades2019virtual}
        computes the measurements from \threeD meshes.
        \qheading{Training} %
        We combine these sources to learn a regressor  with weak supervision
        that infers \threeD shape from an image.
        }
    }
    \label{fig:representation_overview}
\end{figure}

\section{Representations \& Data for Body Shape}  \label{sec:shape_data_represent}

\cameraready{%
We %
use \linguisticshapeattributes and %
\measurements
as a connecting component 
between in-the-wild images and ground-truth body shapes;
see \cref{fig:representation_overview}. 
To that end, we annotate \linguisticshapeattributes for \threeD meshes and
in-the-wild
images, the latter from fashion-model agencies, labeled
via Amazon Mechanical Turk. %
}

\subsection{\smplX Body Model}   \label{sec:body_model} 
We use \smplx \cite{Pavlakos2019_smplifyx},
a differentiable \cameraready{model} that maps shape, $\shape$, pose, $\pose$,
and expression, $\expression$, parameters to a \threeD mesh, $\mesh$,
with $N=\nvertices$ vertices, $V$. %
The shape vector $\shape \in \mathbb{R}^{B}$ ($B\leq300$)
has coefficients of a low-dimensional PCA space. 
The vertices are posed with linear blend skinning with a learned rigged skeleton, $X \in \mathbb{R}^{55 \times 3}$.

\subsection{Model-Agency Images}      \label{sec:model_images}

Model agencies typically provide multiple color images of each model, \cameraready{in various poses, outfits, hairstyles, scenes, and with a varying camera framing}, together with 
\measurements
and clothing size.
We collect training data from multiple model-agency websites, focusing on under-represented body types, \cameraready{namely}:
\href{www.curve-models.com}{    \mbox{curve-models.com}},
\href{www.cocainemodels.com}{   \mbox{cocainemodels.com}},
\href{www.nemesismodels.com}{   \mbox{nemesismodels.com}},
\href{www.jayjay-models.de}{    \mbox{jayjay-models.de}},
\href{www.kultmodels.com}{      \mbox{kultmodels.com}},
\href{www.modelwerk.de}{        \mbox{modelwerk.de}},
\href{www.models1.co.uk}{       \mbox{models1.co.uk}}.
\href{showcast.de}{             \mbox{showcast.de}},
\href{www.the-models.de}{       \mbox{the-models.de}}, and
\href{www.ullamodels.com}{      \mbox{ullamodels.com}}.
In addition to photos, we store gender and four \measurements, \ie height, chest, waist and hip circumference, when available. 
\cameraready{%
To avoid 
having the same subject in both the training and test set}, we match model identities across websites to identify models that work for several agencies. 
For details, see \supmat

\begin{figure}[!t]
    \centering
    \includegraphics[trim=000mm 000mm 000mm 000mm, clip=true, width=1.00\linewidth]{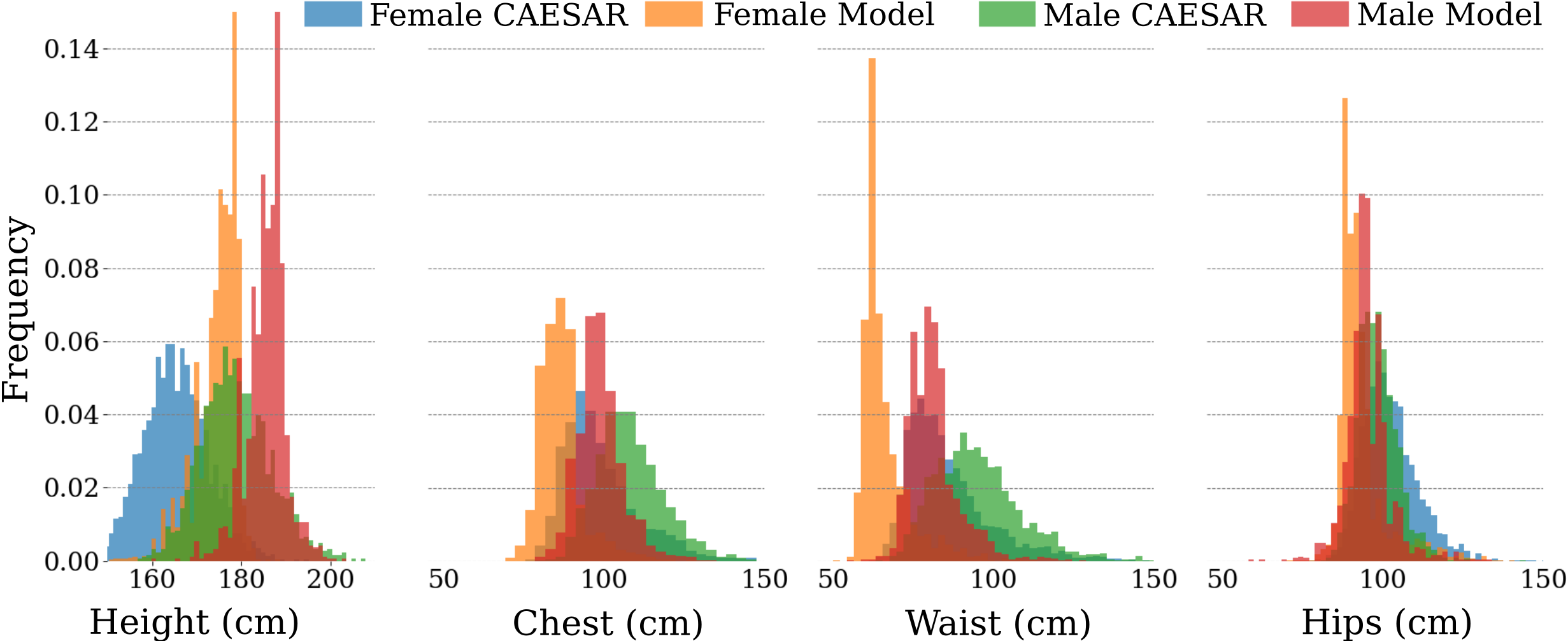}
    \caption{%
        Histogram of height and chest/waist/hips circumference for
        data from model-agency websites (\cref{sec:model_images}) and \caesar. %
        \cameraready{Model-agency data is diverse, yet not as much as \caesar data.}
    }
    \label{fig:agencies_histogram}
\end{figure}

After identity filtering, we have %
$94,620$ images of $4,419$ models along
with their 
\measurements. 
\cameraready{However, the distributions of these measurements,} %
shown in \cref{fig:agencies_histogram}, 
reveal a bias for ``fashion model'' body shapes, while other body types are under-represented %
\cameraready{in comparison} 
to \caesar~\cite{CAESAR}.
\cameraready{To enhance diversity in}
body-shapes \cameraready{and avoid strong biases and log tails}, 
we compute the quantized 
\twoD-distribution for height and weight and sample up to $3$ models per bin.
This results in $N=1,185$ models ($714$ females, $471$ males) and $20,635$ images.

\subsection{\LinguisticShapeAttributes} 
\label{sec:attributes}

Human body shape can be described by 
\linguisticshapeattributes 
\cite{hill2015exploring}. 
We draw inspiration from Streuber \etal~\cite{Streuber:SIGGRAPH:2016} who collect \scores for $30$ \linguisticattributes for $256$ \threeD body meshes, generated by sampling \smpl's shape space, to train a linear ``attribute to shape'' regressor. 
In contrast, we train a model that takes as input an image, instead of attributes, and outputs an accurate \threeD shape (and pose). 

We crowd-source \linguisticattributescores for a variety of body shapes, using 
\cameraready{images from the following sources:}

\qheading{\cameraready{Rendered \caesar images}}  \label{sec:caesar_data} 
We use \cameraready{\caesar \cite{CAESAR} bodies} 
to learn \cameraready{mappings} %
between \linguisticshapeattributes, 
\measurements, 
and \smplX shape parameters, $\shape$.
Specifically, we 
\cameraready{register a ``gendered''} 
\smplx model with $100$ shape components
to $1,700$ male and $2,102$ female \threeD scans,
\cameraready{pose}
all meshes in an \apose, %
and render \cameraready{synthetic images} %
with the same 
\cameraready{virtual} camera.

\qheading{Model-agency \cameraready{photos}}
Each annotator is shown $3$ %
body images per subject, sampled from the \cameraready{image} pool of \cref{sec:model_images}.

\zheading{\cameraready{Annotation}} %
To keep annotation tractable, 
we use $A=15$ \linguisticshapeattributes per gender \cameraready{(subset of \cameraready{\bodytalk's} \cite{Streuber:SIGGRAPH:2016} attributes)}; see  \cref{tab:selected_attributes}.
Each image is annotated by $\numattr = 15$ annotators on Amazon Mechanical Turk. %
Their task is to \emph{``indicate how strongly [they] agree or disagree that the [listed] words describe the shape of the [depicted] person's body''}; 
\cameraready{for an example, see \supmat} 
\cameraready{Annotations} 
range on a \cameraready{discrete} 5-level Likert scale from 1 (strongly disagree) to 5 (strongly agree). 
We get a rating matrix $\mathbf{A} \in \{1,2,3,4,5\}^{N \times A \times \numattr}$,
\cameraready{where $N$ is the number of subjects}. 
\cameraready{In the following}, $a_{ijk}$ denotes an element of $\mathbf{A}$. %

\begin{table}
\renewcommand{\arraystretch}{\myarraystretch} 
    \centering
    \footnotesize
    \begin{tabular}{c c||c|c}
        \multicolumn{2}{c||}{\bf Male \& Female}    & {\bf Male only}   & {\bf Female only} \\ \hline
        short           &   long neck               & skinny arms       & pear shaped       \\
        big             &   long legs               & average           & petite            \\
        tall            &   long torso              & rectangular       & slim waist        \\
        muscular        &   short arms              & delicate build    & large breasts     \\
                        &   broad shoulders         & soft body         & skinny legs       \\
                        &                           & masculine         & feminine          
    \end{tabular}
    \vspace{\vspaceTABaboveCaption}
    \caption{   
        \Linguisticshapeattributes \cameraready{for human bodies}. %
        Some attributes apply to both %
        \cameraready{genders}, 
        but others are 
        \cameraready{gender specific}. 
    }
    \label{tab:selected_attributes}
\end{table}

\section{Mapping Shape Representations} 
\label{sec:mappings}

In \cref{sec:shape_data_represent} we introduce three body-shape representations: %
(1)     \smplX's PCA shape space            (\cref{sec:body_model}), 
(2)     \measurements                       (\cref{sec:model_images}), and
(3)     \linguisticshapeattribute~\scores   (\cref{sec:attributes}). 
\cameraready{Here we} learn mappings between these, so that in \cref{sec:regressor} we \cameraready{can} define new losses for training body shape regressors using multiple data sources.

\subsection{Virtual Measurements (\StoM)}
\label{sec:virtual_measurements}

We obtain \measurements from a \threeD body \cameraready{mesh} in a \mbox{T-pose},
namely 
height, $\heightArg$, weight, $\weightArg$,
and chest, waist and hip circumferences, $\chestCircArg$, $\waistCircArg$,
and $\hipsCircArg$, respectively,
by following 
Wuhrer \etal~\cite{wuhrer2013estimating} and the
\virtualcaliper \cite{pujades2019virtual}.
For details on how we compute these measurements, see \supmat

\subsection{Attributes and \cameraready{\threeD} Shape}

\zheading{Attributes to Shape (\textbf{\AtoS})}
We predict \smplX shape coefficients from \linguisticattributescores
\cameraready{with a second-degree polynomial regression model.}
For each shape $\shape_i$, $i = 1 \dots \TODO{N}$, we create a feature vector,
\TODO{$\mathbf{x}_i^{\AtoS}$}, by averaging 
for \cameraready{each of the} $A$ attributes 
the corresponding \cameraready{$\numattr$} \scores:
\begin{equation}
\cameraready{
    \TODO{\TODO{\mathbf{x}}_i^{\AtoS}}    = [\bar{a}_{i,1}, \dots, \bar{a}_{i,A}]
    \text{,}    
    \quad\quad
    \bar{a}_{i,j}   = \frac{1}{\numattr} \sum_{k=1}^{\numattr} a_{ijk}
    \text{,}
}
\end{equation}
\cameraready{where 
$i$ is the shape        index (list of ``fashion'' or \caesar bodies), 
$j$ is the attribute    index, and 
$k$    the annotation   index.}

\noindent
We then define the full feature matrix for all $N$ shapes as:
\begin{equation}
    \cameraready{\mathbf{X}^{\AtoS}} =    [
                        \phi(\TODO{\mathbf{x}}_1^{\AtoS}),  \quad
                        \dots,               \quad
                        \phi(\TODO{\mathbf{x}}_N^{\AtoS}) 
                    ]^\top
                    \text{,}
\end{equation}
where $\phi(\mathbf{x}_i^{\AtoS})$ maps $\mathbf{x}_i$ to
\cameraready{2\textsuperscript{nd} order}
polynomial features.

\noindent
\cameraready{%
The target matrix $\lineartarget   = [\shape_1, \dots, \shape_N]^\top$ contains 
the shape parameters 
$\bm{\shape}_i = [\shape_{i,1}, \dots, \shape_{i,B}]^\top$. 
We compute 
the polynomial model's coefficients $\bm{\polynomialweights}$ 
via least-squares fitting:}
\begin{equation}
    \lineartarget = \mathbf{X\polynomialweights} + \epsilon
    \text{.}
\end{equation}
Empirically, the polynomial model performs better than several models that we evaluated; 
for details, see \supmat

\zheading{Shape to Attributes (\StoA)}
\cameraready{We} 
predict \linguisticattributescores,
\cameraready{$\predattr$}, 
from \smplX shape parameters, $\shape$.
Again, we  fit a second-degree polynomial regression model.
\cameraready{\StoA has ``swapped'' inputs and outputs \wrt \AtoS:}
\begin{align}
    \mathbf{x}_{i}^{\StoA} &= [\shape_{i,1}, \dots, \shape_{i,B}]
    \text{,} \\
    \mathbf{y}_i &= \cameraready{ [\bar{a}_{i,1}, \dots, \bar{a}_{i,A}]^\top }
    \text{.} 
\end{align}

\zheading{Attributes \& Measurements to Shape (\AFMtoS)}
Given a sparse set of 
\measurements, 
we predict 
\cameraready{\smplX shape parameters}, 
$\shape$.
\cameraready{The input vector is:}
\begin{equation}
    \mathbf{x}_{i}^{\colorFMtoS} = 
    [{\colorheight{h_{i}}}, {\colorweight{w_{i}}}, {\colorcirc{c_{c_i}, c_{w_i}, c_{h_i}}}]
    \text{,}
\end{equation}
where
\cameraready{$c_c, c_w, c_h$ is the chest, waist, and hip circumference, respectively, 
$h$ and $w$ are the height and weight, and}
\colorFMtoS~\cameraready{means} \textit{{\colorheight{H}}eight + {\colorweight{W}}eight + {\colorcirc{C}}ircumference to Shape}.
\cameraready{The regression target is} the \smplX shape parameters, $\mathbf{y}_i$.

When both \textit{{\colorattr{A}}ttributes} and measurements are available, we combine them 
for the \colorAFMtoS model \cameraready{with} input: %
\begin{align}
    \mathbf{x}_{i}^{\colorAFMtoS} = [{\colorattr{\bar{a}_{i,1}, \dots, \bar{a}_{i,A}}}, {\colorheight{h_{i}}}, {\colorweight{w_{i}}}, {\colorcirc{c_{c_i}, c_{w_i}, c_{h_i}}}]
    \text{.}
\end{align}

\cameraready{In practice,} 
depending on which measurements are available, we train and use different regressors.
Following the naming convention of \colorAFMtoS, these models are: \colorAHtoS, \colorAHWtoS, \colorACtoS, and \colorAHCtoS, as well as their equivalents without attribute input \colorHtoS, \colorHWtoS, \colorCtoS, and \colorHCtoS.
\cameraready{For an evaluation of the contribution of \linguisticshapeattributes on top of each \measurement, see  \supmat}

\zheading{Training Data}
To train the \AtoS and \StoA mappings we use \caesar data, for which we have 
\cameraready{\smplX shape parameters,} 
\measurements, and \linguisticattributescores.
\cameraready{We train separate gender-specific models.}

\section{\threeD Shape Regression from an Image}     \label{sec:regressor}

We 
\cameraready{present}
\modelname, a network that \cameraready{predicts}
\smplx parameters from an \rgb image with more accurate body shape than 
\cameraready{existing methods}. 
To improve the realism and accuracy of shape, we explore training losses based on
all shape representations
\cameraready{discussed above}, \ie, 
\smplX meshes (\cref{sec:body_model}),
\linguisticattributescores (\cref{sec:attributes}) and 
\cameraready{\measurements} (\cref{sec:virtual_measurements}).
In the following, symbols with/-out a hat are regressed/\groundtruth values. 

\cameraready{%
We convert %
shape $\hat{\shape}$
to height and circumferences values
$\{
    \hat \height,
    \hat \chestCirc, 
    \hat \waistCirc,
    \hat \hipsCirc \} =  $ $ \{
         \height    (\hat \shape),
         \chestCirc (\hat \shape),
         \waistCirc (\hat \shape),
         \hipsCirc  (\hat \shape)
 \}
$,
by applying our 
virtual measurement tool 
(\cref{sec:virtual_measurements})
to
the
mesh $\mesh(\hat{\shape})$ in the canonical T-pose.} 
\cameraready{%
We also convert %
shape 
$\hat{\shape}$ to \linguisticattributescores, with}
\TODO{$\hat{\predattr} = \text{\StoA}(\hat \shape)$}.

\begin{figure}
    \centering
    \includegraphics[width=0.85\linewidth]{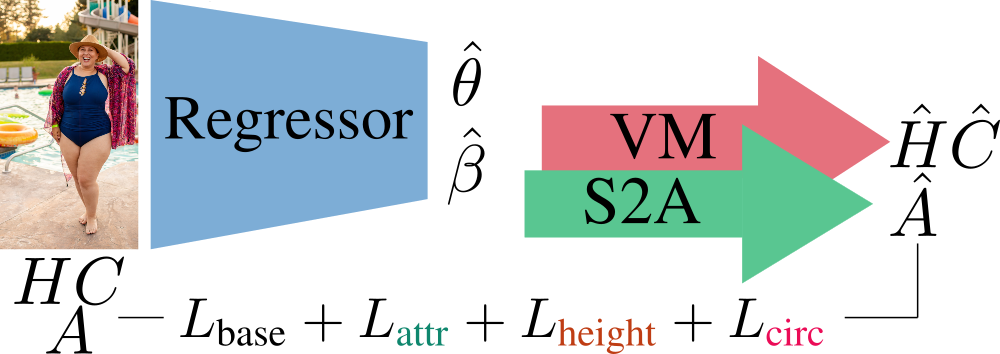}
    \caption{%
        \modelname~%
        first estimates shape, \cameraready{$\hat{\beta}$}, and pose, \cameraready{$\hat{\theta}$}. 
        \cameraready{Shape} 
        is \cameraready{used by:} %
        (1) our virtual \measurement \cameraready{(\StoM)} module to compute height, $\hat{H}$, and \cameraready{circumferences}, %
        $\hat{C}$, and 
        (2) our \StoA module to \cameraready{infer}~\linguisticattributescores, $\hat{\predattr}$.
        \cameraready{There are several \modelname variations}, \eg, 
        \shapyH uses only \StoM to \cameraready{infer} $\hat H$, while 
        \shapyHA uses \StoM to \cameraready{infer} $\hat H$ and \StoA to \cameraready{infer} $\hat{\predattr}$.
    }
    \label{fig:shapy_ah1}
\end{figure}

We train various \modelname versions %
\cameraready{with the following \cameraready{``\modelname losses''}, using %
either \linguisticshapeattributes, or \measurements, or both:}
{\small
\begin{align}
    L_\text{{\colorattr{attr}}}         &=   ||  \predattr - \hat{\predattr}||_2^2 \text{,}    
    \label{eq:attr_loss}                \\
    L_\text{{\colorheight{height}}}     &=   ||  \height - \hat H||_2^2  \text{,}
    \label{eq:height_loss}              \\
    L_\text{{\colorcirc{circ}}}         &= 
    \sum\nolimits_{i \in \{ c, w, h \} } || C_i - \hat{C}_i ||^2_2
    \label{eq:circ_loss}
\end{align}
}
These are optionally added to a base loss, $L_\text{\vanilla}$, \cameraready{defined below in ``training details''}.
The architecture of \modelname, with all optional components, is shown in \cref{fig:shapy_ah1}.
A suffix of color-coded letters describes which of the above losses
are used when training a model.
For example, \mbox{\modelname-\colorAH} 
denotes a model trained with the attribute and height losses,
{\ie}: 
\mbox{$L_\text{\modelname-\colorAHtoS} = L_\text{\vanilla} +
L_\text{{\colorattr{attr}}} + L_\text{{\colorheight{height}}}$.}

\smallskip
\qheading{Training Details}
We initialize \modelname with the \expose~\cite{Choutas2020_expose} network weights and use curated fits \cite{Choutas2020_expose}, H3.6M \cite{ionescu2013human36m}, the SPIN \cite{Kolotouros2019_spin} training data, and our model-agency dataset (\cref{sec:model_images}) for training.
In each batch, 50\% of the images are sampled from the model-agency images,
\cameraready{for which we ensure a} %
gender balance.
The 
\cameraready{``\modelname losses'' of \cref{eq:attr_loss,eq:height_loss,eq:circ_loss}} 
are 
applied only on the model-agency images. 
We use \cameraready{these on top of} a standard \cameraready{base loss:} %
\begin{align}
    L_\text{\vanilla} &= L_\text{pose} + L_\text{shape} \text{,}                      %
\end{align}
where 
$L_\text{joints}^{\text{\twoD}}$ and $L_\text{joints}^{\text{\threeD}}$ are \twoD and \threeD joint losses:
\begin{align}
    L_\text{pose}     &= L_\text{joints}^{\text{\twoD}} + L_\text{joints}^{\text{\threeD}} + L_{\pose} \text{,} \\
    L_\text{shape}    &= L_{\shape} + L_{\shape}^{\text{prior}} \text{,}
\end{align}
$L_{\pose}$             and 
$L_{\shape}$            are losses on pose and shape parameters, and 
$L_{\shape}^{\text{prior}}$ is \pixie's \cite{feng2021pixie} ``gendered'' shape prior.
All losses are L2, unless otherwise explicitly specified.
\cameraready{Losses on \smplx parameters are applied only on the 
pose data \mbox{\cite{ionescu2013human36m,Choutas2020_expose,Kolotouros2019_spin}}.}
For more implementation details, see \supmat

\section{Experiments}
\label{sec:experiments}

\subsection{Evaluation Datasets}     \label{sec:exp_datasets}

\qheading{\threeD Poses in the Wild (\threedpw) \cite{vonMarcard2018}}
\cameraready{%
We use this to evaluate \emph{pose} accuracy. 
This is widely used, but has only 5 test subjects, \ie, limited shape variation. 
For results, see \supmat}

\qheading{\ssplong (\ssp)\cite{sengupta2020straps}}
We use this to evaluate \threeD body \emph{shape} accuracy  from images. 
It has $62$ tightly-clothed subjects in $311$ \inthewild images from 
\mbox{Sports-1M} \cite{KarpathyCVPR14}, 
with \emph{pseudo} \groundtruth~\smpl meshes that we \TODO{convert} to \smplX for evaluation.

\qheading{Model Measurements Test Set (\mmts)}
We \cameraready{use this}
to evaluate 
\measurement accuracy,
as a proxy for body \emph{shape} accuracy.
To create \mmts, we withhold
2699/1514 images of 143/95 female/male identities from our 
model-agency
data, %
\cameraready{described in \cref{sec:model_images}}

\qheading{\caesar Meshes Test Set (\cmts)} 
We use \caesar to measure the accuracy of \smplX body shapes and \linguisticshapeattributes
\cameraready{for} 
the models
of \cref{sec:mappings}.
Specifically, we compute:
(1) errors for \smplx meshes estimated from \linguisticshapeattributes and/or \measurements by \AtoS and its variations, and
(2) errors for \linguisticshapeattributes estimated from \smplX meshes by \StoA.
To create an unseen mesh test set, we withhold 339 male and 410 female \caesar meshes from the crowd-sourced \caesar~ \linguisticshapeattributes, described in \cref{sec:caesar_data}.

\begin{figure}%
	\centering
    \includegraphics[width=1.00\linewidth]{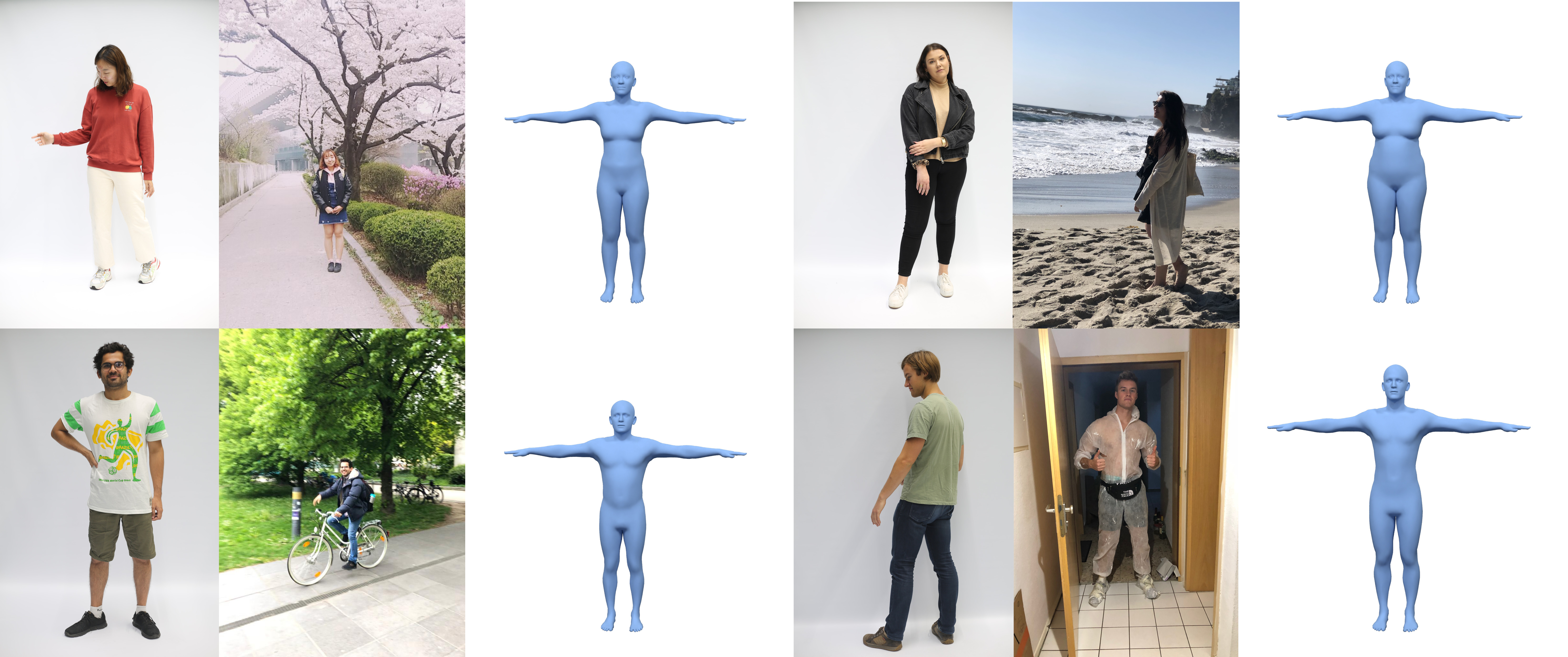}
	\caption{%
    ``Human Bodies in the Wild'' (\hbw)
    \cameraready{color images, taken in the lab and in the wild,
    and the \smplx ground-truth shape.}
	}
	\label{fig:hbw}
\end{figure}

\qheading{Human Bodies in the Wild (\hbw)}
The field is missing a dataset with
varied bodies, varied clothing, \inthewild images, and accurate \emph{\threeD shape ground truth}. 
We fill this gap by collecting a novel dataset, 
called
\mbox{\emph{``Human Bodies in the Wild'' (\hbw)}}, 
with three steps: %
\mbox{(1)     We collect} \cameraready{accurate} \threeD \cameraready{body} scans for \hbwNumberSubjects subjects (\hbwNumberSubjectsF female, \hbwNumberSubjectsM male), and register \cameraready{a ``gendered''}~\smplX model to these to recover \threeD \smplX ground-truth bodies \cameraready{\cite{Dyna}}. 
\mbox{(2)     We take photos} of each subject in ``photo-lab'' settings, \ie, in front of a white background with controlled lighting, and in various everyday outfits and ``fashion'' poses. 
\mbox{(3)     Subjects upload} full-body photos of themselves taken in the wild. %
For each subject we take up to \hbwNumberPhotosLabTHRESHperSubj photos in lab settings, and collect up to \hbwNumberPhotosWildTHRESHperSubj~\inthewild photos. 
In total, \hbw has \hbwNumberPhotos photos, \hbwNumberPhotosLab in the lab setting and  \hbwNumberPhotosWild in the wild. 
We split the data into a validation and a test set \cameraready{(val/test)} with 
\hbwNumberSPLITSsubj
and 
\hbwNumberSPLITSimg, 
respectively.
\Cref{fig:hbw} shows a few \hbw subjects, photos and their \smplX \groundtruth shapes.
All subjects gave prior written informed consent to participate in this study and to release the data. %
The study was reviewed by the 
ethics board 
of the \cameraready{University of T{\"ubingen}}, without objections.

\subsection{Evaluation Metrics} \label{sec:metrics}

\cameraready{We use standard accuracy metrics for \threeD body pose, but also introduce metrics specific to  \threeD body shape.} 

\qheading{\Measurements}
We report the mean absolute error in mm between \groundtruth and estimated measurements, computed 
as described in \cref{sec:virtual_measurements}. 
When weight \cameraready{is available}, %
we report the mean absolute error in kg. 

\qheading{\mpjpe and \VtoV metrics}
We
report in \supmat
the
mean per-joint point error (\mpjpe) and
mean vertex-to-vertex error (\vtov), when \smplx meshes are available. 
The prefix ``PA'' denotes metrics after Procrustes alignment.

\qheading{Mean point-to-point error (\vtovHD)}
\TODO{
\cameraready{\smplx has a highly non-uniform vertex distribution across the body, %
which negatively biases the mean vertex-to-vertex (\VtoV) error, when comparing estimated and ground-truth \smplX meshes.
To account for this, we evenly sample $20$K points on \smplX's surface,
and report the mean point-to-point %
(\vtovHD)
error.
For details, %
see \supmat}
}

\subsection{\cameraready{Shape-Representation Mappings}}       \label{sec:exp_shape_est}

We evaluate \cameraready{the models \AtoS and \StoA, which map between the various body} shape representations \cameraready{(\cref{sec:mappings})}.

\begin{table}
\renewcommand{\arraystretch}{\myarraystretch} 
    \centering
    \scriptsize
	\resizebox{\columnwidth}{!}{
	\setlength\tabcolsep{4pt}
        \begin{tabular}{m{0.5em}lllllll}
        \toprule
         & Method  & \vtovHD   & Height & Weight & Chest & Waist & Hips \\
         & -  & (mm)   & (mm) & (kg) & (mm) & (mm) & (mm) \\
         \midrule
            \multirow{9}*{\rotatebox{90}{\small Male subjects}}
            & \AtoS &  $11.1 \pm 5.2$ &   $29 \pm 21$ &  $5 \pm 4$ &  $30 \pm 22$ &  $32 \pm 24$ &  $28 \pm 21$ \\ 
            & \HtoS &  $12.1 \pm 6.1$ &   $5 \pm 4$ &  $11 \pm 11$ &  $81 \pm 66$ &  $102 \pm 87$ &  $40 \pm 33$ \\
            & \AHtoS &  $6.8 \pm 2.3$ &   $4 \pm 3$ &  $3 \pm 3$ &  $27 \pm 21$ &  $29 \pm 23$ &  $24 \pm 18$ \\ 
            & \HWtoS &  $8.1 \pm 2.7$ &   $5 \pm 4$ &  $1 \pm 1$ &  $24 \pm 17$ &  $26 \pm 20$ &  $21 \pm 18$ \\
            & \AHWtoS &  $6.3 \pm 2.1$ &    $4 \pm 3$ &  $1 \pm 1$ &  $19 \pm 15$ &  $19 \pm 14$ &  $20 \pm 16$ \\ 
            & \CtoS &  $19.7 \pm 11.1$ &   $59 \pm 47$ &  $9 \pm 8$ &  $55 \pm 41$ &  $63 \pm 49$ &  $37 \pm 28$ \\
            & \ACtoS &  $9.6 \pm 4.4$ &    $25 \pm 19$ &  $3 \pm 3$ &  $23 \pm 19$ &  $21 \pm 17$ &  $18 \pm 14$ \\ 
            & \HCtoS &  $7.7 \pm 2.6$ &   $5 \pm 4$ &  $2 \pm 2$ &  $28 \pm 23$ &  $18 \pm 15$ &  $13 \pm 11$ \\
            & \AHCtoS &  $6.0 \pm 2.0$ &  $4 \pm 3$ &  $2 \pm 2$ &  $21 \pm 17$ &  $17 \pm 14$ &  $13 \pm 10$ \\ 
            & \HWCtoS &  $7.3 \pm 2.6$ &   $5 \pm 4$ &  $1 \pm 1$ &  $20 \pm 15$ &  $14 \pm 12$ &  $13 \pm 11$ \\
            & \AHWCtoS &  $5.8 \pm 2.0$ &  $4 \pm 3$ &  $1 \pm 1$ &  $16 \pm 13$ &  $13 \pm 10$ &  $13 \pm 10$ \\

        \bottomrule
        \end{tabular}
    }
    \vspace{\vspaceTABaboveCaption}
    \caption{%
        Results of \AtoS~\cameraready{variants} on \cmts
        for male subjects, using 
        the male \smplx model. 
        For females, see \supmat
    }
    \label{tab:a2s_quant_result_cmts}
\end{table}

\qheading{\AtoS and its variations}
How well can we infer \threeD body shape from just \linguisticshapeattributes, \measurements, or %
\cameraready{both of these}
together?
In \cref{tab:a2s_quant_result_cmts}, we report reconstruction and measurement errors using many combinations of 
attributes      ({\colorattr{A}}), 
height          ({\colorheight{H}}), 
weight          ({\colorweight{W}}), and 
circumferences  ({\colorcirc{C}}).
Evaluation on \cmts data shows that attributes improve the overall shape prediction across the board. 
For example, height+attributes (\colorAHtoS) has a lower point-to-point error than height alone. 
The best performing model, \colorAHWCtoS, uses everything, with \vtovHD-errors of 
$5.8 \pm 2.0$ mm (males) and 
$6.2 \pm 2.4$ mm (females). 

\smallskip
\qheading{\StoA}
How well can we infer \linguisticshapeattributes from \threeD shape?
\cameraready{\StoA's accuracy on inferring} the attribute Likert score is
$75\%/69\%$ %
for males/females; details in \supmat

\begin{figure*}
    \centering
    \includegraphics[trim=050mm 020mm 050mm 020mm,clip=true, width=1.0\linewidth]{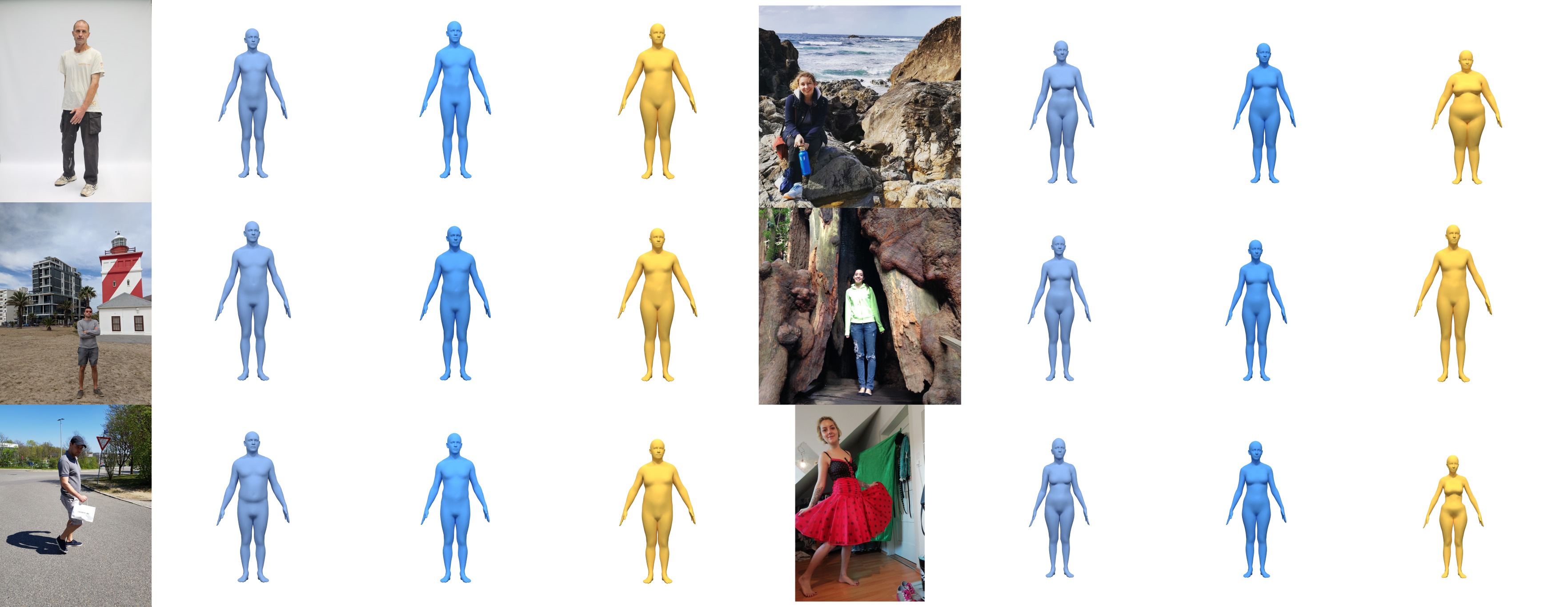}
	\caption{%
    Qualitative results from \hbw. 
    From left to right: \rgb,  ground-truth shape,
    \modelname and Sengupta \etal \cite{sengupta2021hierarchicalICCV}.
    \cameraready{For example, in the upper- and lower- right images, \modelname is less affected by pose variation and loose clothing.}
    }
\label{fig:resultimages}
\end{figure*}

\subsection{\cameraready{\threeD Shape from an Image}}

\cameraready{We evaluate all of our model's variations (see \cref{sec:regressor}) on the \hbw validation set and find, perhaps surprisingly, that \shapyA outperforms other variants}. 
\cameraready{We refer to this below (and \cref{fig:teaser}) simply as ``\modelname'' and report its performance} in \cref{tab:hbw} for \hbw, \cref{tab:measurements_test} for \mmts, and \cref{tab:straps} for \ssp. 
For images with \cameraready{natural and} varied clothing (\hbw, \mmts), \modelname significantly outperforms all other methods
(\cref{tab:hbw,tab:measurements_test}) 
using only weak \threeD shape supervision ({\colorattr{A}}ttributes). 
On these images, Sengupta \etal's method \cite{sengupta2021hierarchicalICCV} struggles with the \cameraready{natural} clothing.
In contrast, \cameraready{their method} is more accurate than \modelname on \ssp (\cref{tab:straps}), which has tight ``sports" clothing, in terms of \mbox{PVE-T-SC}, a scale-normalized metric used on this dataset. 
These results show that silhouettes are good for tight/minimal clothing and that \modelname struggles with high \bmi shapes due to the lack of such shapes in our training data; see \cref{fig:agencies_histogram}.
\cameraready{Note that, as \hbw}
has true ground-truth \threeD shape, it does not need \ssp's scaling for evaluation. %

\def\boxit#1{%
  \smash{\color{Fuchsia}\fboxrule=1pt\relax\fboxsep=2pt\relax%
  \llap{\rlap{\fbox{\vphantom{0}\makebox[#1]{}}}~}}\ignorespaces
}

\begin{table}
\renewcommand{\arraystretch}{\myarraystretch} 
\centering
\scriptsize
\resizebox{\columnwidth}{!}{
\begin{tabular}{lllllll}
    \toprule
    \multirow{1}{1em}{Method} %
    & Model & Height & Chest & Waist & Hips & \vtovHD \\
    \midrule
        \smplr
        \cite{madadi2020smplr}
        & \smpl 	& 	182 & 	267 &    309  &   305  &  69 \\
        \straps \cite{sengupta2020straps}                   & \smpl 	&  135 & 	167 & 	145 &    102 & 47 \\
        \spin   \cite{Kolotouros2019_spin}                  & \smpl  &  59  & 	92 &	78  &	101 &    29 \\
        \tuch \cite{mueller2021tuch}                        & \smpl  &	58  &	89 &	75  &	\textbf{57} &    26 \\
        Sengupta \etal \cite{sengupta2021hierarchicalICCV}  & \smpl  &	82  & 	133 &	107 &	63 &    32 \\
        \expose \cite{Choutas2020_expose}                   & \smplx  &	85  &	99 &	92  &	94 &    35 \\
        \modelname (ours)                                   & \smplx  &	\textbf{51}  &	\textbf{65} &	\textbf{69}  &	\textbf{57} &	\textbf{21} \\         
    \bottomrule
\end{tabular}
}
\vspace{\vspaceTABaboveCaption}
\caption{%
        Evaluation on the \hbw \cameraready{test set} in mm. 
        We compute \cameraready{the measurement and} point-to-point (\vtovHD) error 
        between predicted and ground-truth \smplX meshes. %
}
\label{tab:hbw}
\end{table}

\begin{table}
\renewcommand{\arraystretch}{\myarraystretch} 
    \centering
    \scriptsize
    \begin{tabular}{llllll}
        \toprule
        & \multicolumn{5}{c}{Mean absolute error (mm) $\downarrow$}                                           \\
        Method                              & Model     & Height    & Chest     & Waist     & Hips          \\ 
        \midrule  %
        Sengupta \etal \cite{sengupta2021hierarchicalICCV} & \smpl  & 84       & 186       & 263       & 142  \\
        \tuch \cite{mueller2021tuch}        & \smpl  &    82      &  92       &  129      &  91        \\
        \spin   \cite{Kolotouros2019_spin}  & \smpl     &  72    & 91     & 129      & 101      \\
        \straps \cite{sengupta2020straps}   & \smpl      &  207       &   278  &    326   &   145  \\ 
        \expose \cite{Choutas2020_expose}   & \smplx    & 107       & 107       & 136       & 92            \\
        \modelname (ours)                   & \smplx  & \textbf{71} & \textbf{64} & \textbf{98}  & \textbf{74}     \\
        \bottomrule
    \end{tabular}
    \vspace{\vspaceTABaboveCaption}
    \caption{
        Evaluation on \mmts
        .
        We report the mean absolute error between \groundtruth and estimated measurements.
    }
    \label{tab:measurements_test}
\end{table}

A key observation 
is that training with \linguisticshapeattributes alone is sufficient, \ie, without \measurements.
\cameraready{Importantly}, this opens up the possibility for \cameraready{significantly} larger %
data collections. %
For a study of how different measurements or attributes impact accuracy, see \supmat
\Cref{fig:resultimages} shows \modelname's qualitative results.

\vfill

\section{Conclusion}
\modelname is trained to regress more accurate human body shape from images than previous methods, without explicit \threeD shape supervision.
To achieve this,  we present two different ways to collect
proxy annotations for \threeD body shape for in-the-wild images. 
First, we collect sparse \measurements from 
\cameraready{online model-agency} data.
Second, we annotate images with \linguisticshapeattributes using crowd-sourcing.
We learn mappings between body shape, measurements, and attributes, enabling us to supervise a regressor using any combination of these.
To evaluate \modelname, we introduce a new shape estimation benchmark, the ``Human Bodies in the Wild'' (\hbw) dataset. \hbw has images of people in natural clothing and natural settings together with ground-truth \threeD shape from a body scanner.
\hbw is more challenging than existing shape benchmarks like \ssp, and \modelname significantly outperforms existing methods on this benchmark. 
We believe this work will open new  directions, since the idea of leveraging linguistic annotations to improve \threeD shape has many applications. 

\begin{table}[t]
\renewcommand{\arraystretch}{\myarraystretch} 
\centering
\scriptsize
\begin{tabular}{llll}
    \toprule
    Method &
    Model & PVE-T-SC & mIOU\\
    \midrule
    \hmr    \cite{Kanazawa2018_hmr}        & \smpl & 22.9    & 0.69    \\
    \spin   \cite{Kolotouros2019_spin}     & \smpl & 22.2    & 0.70    \\
    \straps \cite{sengupta2020straps}      & \smpl & 15.9    & \textbf{0.80}    \\
    Sengupta \etal \cite{sengupta2021hierarchicalICCV}  & \smpl &  \textbf{13.6} &   -     \\
    \modelname (ours)          & \smplx        & 19.2       & -       \\ 
    \bottomrule
\end{tabular}
\vspace{\vspaceTABaboveCaption}
\caption{
        Evaluation on the \ssp~\cameraready{test set} \cite{sengupta2020straps}. 
        We report the scaled mean vertex-to-vertex error in \tpose
        \cite{sengupta2020straps}, \cameraready{and mIOU}.
}
\label{tab:straps}
\end{table}

\pagebreak

\qheading{Limitations}
Our \cameraready{model-agency} training dataset (\cref{sec:model_images}) is not representative of the \cameraready{entire human} population and this
limits \modelname's %
ability to predict larger body shapes. To address this,
we need to find images of more diverse bodies together with \measurements
\cameraready{and \linguisticshapeattributes describing them.}

\qheading{Social impact}
Knowing the \threeD shape of a person has advantages, for example, in the clothing industry to avoid unnecessary returns. If used without consent, \threeD shape estimation may invade individuals' privacy.
As with all other \threeD pose and shape estimation methods, surveillance
and deep-fake creation is another important risk. 
\cameraready{Consequently, \modelname's license prohibits such uses.}

{%
\qheading{Acknowledgments}
This work was supported by the
Max Planck ETH Center for Learning Systems and the 
International Max Planck Research School for Intelligent Systems.
We thank Tsvetelina Alexiadis, Galina Henz, Claudia Gallatz, and Taylor McConnell for the data collection, and Markus H{\"o}schle for the camera setup. 
We thank Muhammed Kocabas, Nikos Athanasiou and Maria Alejandra Quiros-Ramirez for %
the insightful 
discussions.}

{\small
\qheading{Disclosure}
\href{https://files.is.tue.mpg.de/black/CoI_CVPR_2022.txt}{
     https://files.is.tue.mpg.de/black/{CoI\_CVPR\_2022.txt}}}

\pagebreak

\clearpage

{\small
    \bibliographystyle{ieee_fullname}
    \bibliography{00_BIB}
}

\clearpage

\begin{appendices}
    \renewcommand{\thefigure}{A.\arabic{figure}}
    \setcounter{figure}{0}
    \renewcommand{\thetable}{A.\arabic{table}}
    \setcounter{table}{0}

\section{\cameraready{Data Collection}}
\subsection{Model-Agency Identity Filtering}
\label{supmat:sec:model_images}

We collect internet data consisting of %
images and height/chest/waist/hips measurements, from model agency websites.
A ``fashion model'' can work for many agencies and their pictures can appear on multiple websites.
To create non-overlapping training, validation and test sets, we match model identities across websites.
To that end, we use \arcface~\cite{deng2018arcface} for face detection and \retinanet~\cite{Deng_2020_CVPR} to compute identity embeddings $E_i \in \mathbb{R}^{512}$ for each image.
For every pair of models $(q, t)$ with the same gender label, let $Q$, $T$ be the number of query and target model images and $\bm{E_Q} \in \mathbb{R}^{Q \times 512}$ and $\bm{E_T} \in \mathbb{R}^{T \times 512}$ the query and target embedding feature matrices.
We then compute the pairwise cosine similarity matrix $\mathcal{S} \in \mathbb{R}^{Q \times T}$ between all images in $\bm{E_Q}$ and $\bm{E_T}$, and the aggregate and average similarity:
\begin{align}
    \mathcal{S}_T(t) & = \frac{1}{Q}  \sum_q        \mathcal{S}(q, t), \\
    \mathcal{S}_{TQ} & = \frac{1}{QT} \sum_q \sum_t \mathcal{S}(q, t).
\end{align}
Each pair with $\mathcal{S}$ and $\mathcal{S}_T$ that has no element larger than the similarity threshold $\tau=0.3$ is ignored, as it contains dissimilar models.
Finally, we check if $\mathcal{S}_{TQ}$ is larger than $\tau$, and we keep a list of all pairs for which this holds true.

\subsection{Crowd-Sourced Linguistic Shape-Attributes}
To collect human ratings of how much a word describes a body shape, we conduct a
human intelligence task (HIT)
on Amazon Mechanical Turk (\amt). 
In this task, we show an image of a person along with $15$ different gender-specific attributes. 
We then ask participants to indicate how strongly they agree or disagree that the provided words describe the shape of this person's body. 
We arrange the rating buttons from strong disagreement to strong agreement with equal distances to create a $5$-point Likert scale. 
The rating choices are ``strongly disagree'' (score $1$), ``rather disagree'' (score $2$), ``average'' (score $3$), ``rather agree'' (score $4$), ``strongly agree'' (score $5$).

\cameraready{We ask multiple persons to rate each body and image, to “average out” the subjectivity of individual ratings \cite{Streuber:SIGGRAPH:2016}. Additionally, we compute the Pearson correlation between averaged attribute ratings and ground-truth measurements. Examples of highly correlated pairs are “Big / Weight”, and “Short / Height”.}

\begin{figure*}[t]
    \centering
    \resizebox{0.80\textwidth}{!}{%
        \includegraphics[width=\textwidth,height=\textheight,keepaspectratio=true,
            trim=000mm 005mm 010mm 000mm, clip=true]{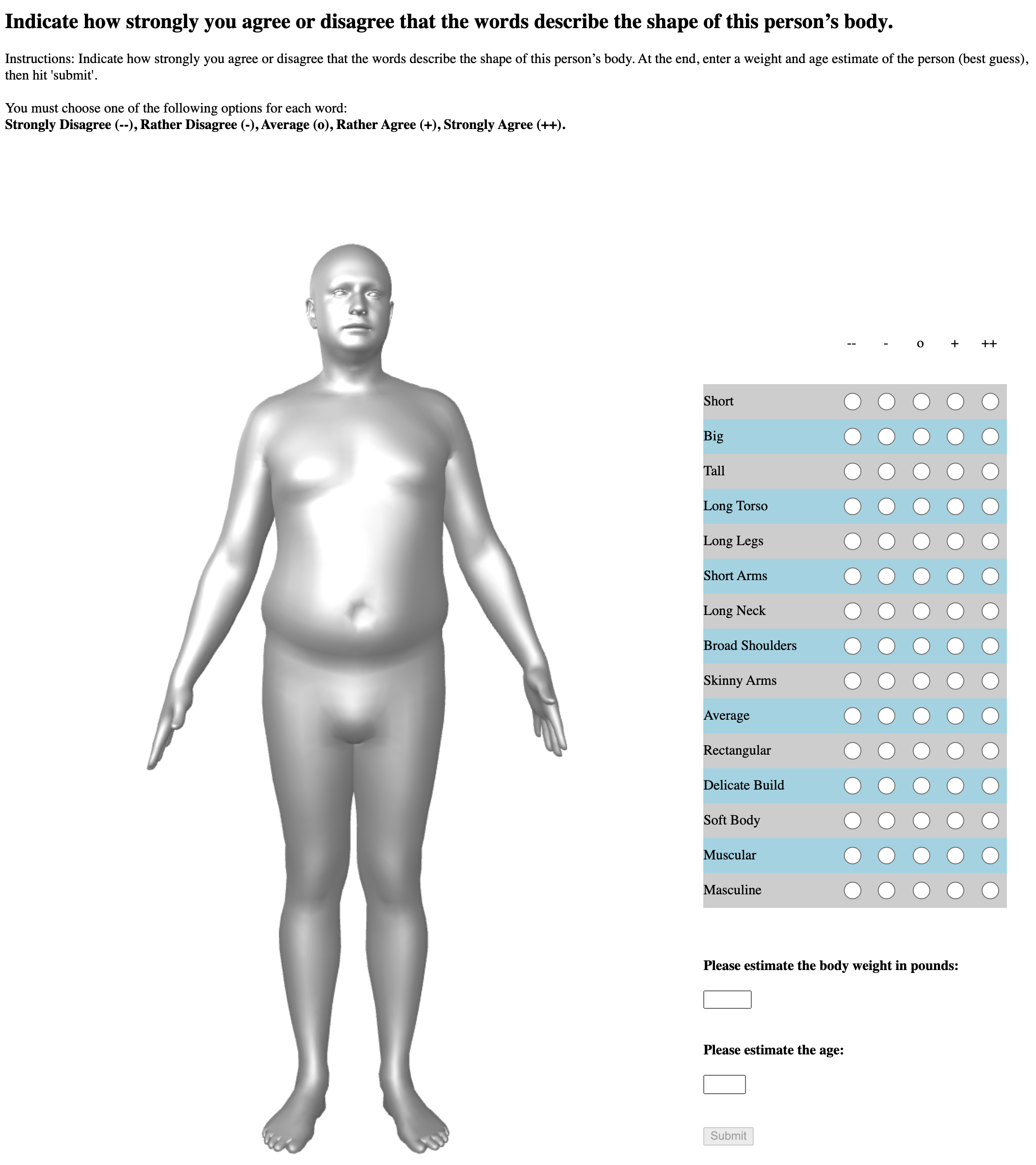}
    }
    \caption{%
        Layout of the \amt task for a male subject.
        \textbf{Left:}      the \threeD body mesh in A-pose.
        \textbf{Right:}     the attributes and ratings buttons.
    }
    \label{fig:amt_task}
\end{figure*}

The layout of our \caesar annotation task is visualized in \cref{fig:amt_task}. 
To ensure good rating quality, we have several qualification requirements per participant: 
submitting a minimum of $5000$ tasks on \amt and an \amt acceptance rate of $95\%$, as well as having a US residency and passing a language qualification test to ensure similar language skills and cultures across raters.

\section{Mapping Shape Representations}
\subsection{Shape to Anatomical Measurements (S2M)}
\label{supmat:sec:virtual_measurements}

\newcommand{\headvert}[0]{v_{\text{head}}}
\newcommand{\heelvertex}[0]{v_{\text{left heel}}}
\newcommand{\chestvertex}[0]{v_{\text{chest}}}
\newcommand{\waistvertex}[0]{v_{\text{waist}}}
\newcommand{\hipvertex}[0]{v_{\text{hip}}}

An important part of our project is the
computation of body measurements.
Following \virtualcaliper \cite{pujades2019virtual}, we present a method to compute anatomical measurements from a \threeD mesh in the canonical \tpose, \ie after ``undoing'' the effect of pose.
Specifically, we measure the height, $\heightArg$, weight, $\weightArg$, and the chest, waist and hip circumferences, $\chestCircArg$, $\waistCircArg$, and $\hipsCircArg$, respectively.
Let
$\headvert(\shape), \heelvertex(\shape), \chestvertex(\shape), \waistvertex(\shape), \hipvertex(\shape)$
be the
head,
left heel,
chest,
waist and
hip
vertices.
$\heightArg$ is computed as the difference in the vertical-axis ``Y'' coordinates between
the top of the head and the left heel:
$\heightArg = \lvert \headvert^y(\shape) - \heelvertex^y(\shape)\rvert$ .
To obtain $W(\shape)$ we multiply the mesh volume by 985 kg/m$^3$, which is the average human body density.
We compute circumference measurements using the method of
Wuhrer \etal~\cite{wuhrer2013estimating}.

\begin{figure}
    \centering%
    \includegraphics[trim=120mm 030mm 150mm 040mm, clip=true, width=1.00 \linewidth]{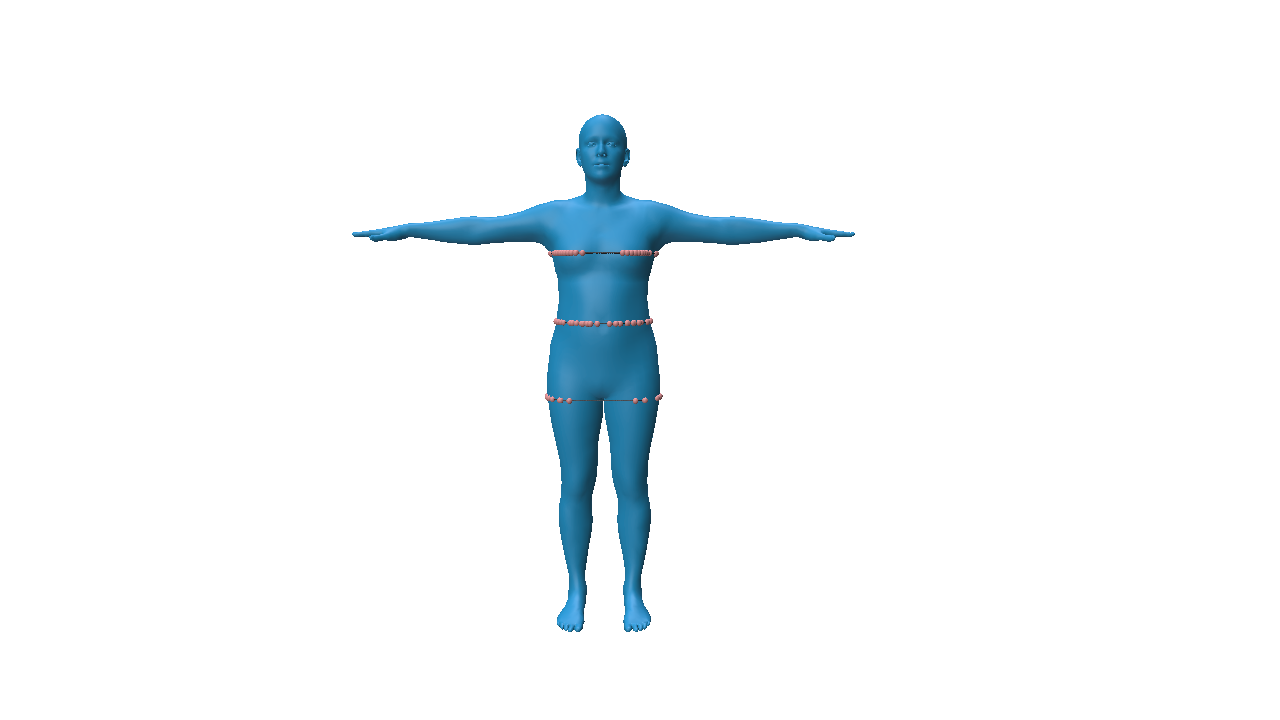}

    \vspace{-00.5 em}
    \caption{
        Automatic anatomical measurements on a \threeD mesh. 
        The red points lie on the intersection of planes at chest/waist/hip height with the mesh, 
        while their convex hull is shown with black lines.
    }
    \label{fig:measurements}
\end{figure}

Here, $T \in \mathbb{R}^{F \times 3 \times 3}$
, where $F=20,908$ is the number of triangles in the \smplx mesh,
denotes ``shaped'' vertices of all triangles of the mesh $\mesh(\shape, \pose)$
;  we drop expressions, $\expression$, which are not used in this work.
Let us explain this using the chest circumference $\chestCircArg$ as an example.
We form a plane $P$ with normal $\bm{n}=(0, 1, 0)$
that crosses the point $\chestvertex(\shape)$.
Then, let
$\mathcal{I} = \{\bm{p}_i\}_{i=1}^{N}$
be the set of points of $P$ that intersect the body mesh (red points in \cref{fig:measurements}).
We store their barycentric coordinates $(\mathtt{u}_i, \mathtt{v}_i, \mathtt{w}_i)$ and the corresponding body-triangle index $t_i$.
Let $\mathcal{H}$ be the convex hull of $\mathcal{I}$ (black lines in \cref{fig:measurements}),
and $\hulledgeindices$
the set of edge indices of $\mathcal{H}$.
$\chestCircArg$ is equal to the length of the convex hull:
\begin{equation}
    \chestCircArg = \sum_{(i, j) \in \hulledgeindices}\norm{
        \begin{pmatrix}
            \mathtt{u}_i \\
            \mathtt{v}_i \\
            \mathtt{w}_i
        \end{pmatrix}^\top T_{t_i} -
        \begin{pmatrix}
            \mathtt{u}_j \\
            \mathtt{v}_j \\
            \mathtt{w}_j
        \end{pmatrix}^\top T_{t_j}
    }_2,
    \label{eq:circumference}
\end{equation}
where $i,j$ are point indices for line segments of $\hulledgeindices$. %
The process is the same for the waist and hips, but the intersection plane
is computed using $\waistvertex, \hipvertex$.
All of $\heightArg, \weightArg, \chestCircArg, \waistCircArg, \hipsCircArg$ are differentiable functions of body shape parameters, $\shape$.

\cameraready{Note that \smplx knows the height distribution of humans and acts as a strong prior in shape estimation. Given the ground-truth height of a person (in meter), $\heightArg$ can be used to directly supervise height and overcome scale ambiguity.}

\subsection{Mapping Attributes to Shape (\AtoS)}
We introduce \AtoS, a model that maps the input attribute ratings to shape components $\shape$ as output.
We compare
a $2^{\text{nd}}$ degree polynomial model
with a linear regression model and a multi-layer perceptron (MLP),
\cameraready{%
    using the \vtovlong (\vtov)
    error metric between predicted and ground-truth \smplx meshes,
    and report results in \cref{tab:AtoS_Comparison_Models}.
} When using only
attributes as input (\AtoS),
the polynomial model of degree $d=2$ achieves the best performance.
Adding height and weight to the input vector requires a small modification,
namely using the cubic root of the weight and converting the height from (m) to (cm).
\cameraready{We}.
With these additions,
the $2^{\text{nd}}$ degree polynomial achieves the best performance.
\begin{table}
\renewcommand{\arraystretch}{1.2} 
\centering
\scriptsize
\begin{tabular}{llllll}
    \toprule
    Model             & Input                           & \multicolumn{2}{c}{\vtov mean $\pm$ std}  \\
                      &                                 & Females              & Males                       \\
    \midrule
    Mean Shape        &                                & 18.01            $\pm$ 8.73            & 19.24            $\pm$ 10.36           \\ 
    Linear Regression & A                               & 10.83            $\pm$ 4.77            & 10.43            $\pm$ 4.63            \\
    Polynomial (d=2)  & A                               & 10.58   $\pm$ 4.67            & 10.25   $\pm$ 4.48   \\
    MLP             & A                               & 10.73            $\pm$ 4.62   & 10.33            $\pm$ 4.57            \\ 
    \midrule
    Linear Regression & A+H+W                             & 7.00             $\pm$ 2.59            & 6.56             $\pm$ 2.21   \\
    Polynomial (d=2)  & A+H+W                             & 7.31             $\pm$ 2.56            & 6.71             $\pm$ 2.21            \\
    MLP             & A+H+W                             & 7.03             $\pm$ 2.6             & 6.68             $\pm$ 2.24            \\
    Linear Regression & A+H+$\sqrt[3]{W}$                         & 6.97             $\pm$ 2.58            & 6.54             $\pm$ 2.22            \\
    Polynomial (d=2)  & A+H+ $\sqrt[3]{W}$                         & \textbf{6.88}    $\pm$ \textbf{2.55}   & \textbf{6.49}    $\pm$ \textbf{2.20}   \\
    \bottomrule
\end{tabular}
\caption{Comparison of models for \AtoS and  \AHWtoS regression.}
\label{tab:AtoS_Comparison_Models}
\end{table}

\subsection{Images to Attributes (\ItoA)}
\label{supmat:subsec:attrs_from_images}

\cameraready{%
    We briefly experimented with models that learn to predict
    attribute scores from images (\ItoA). This attribute predictor
    is implemented using a \resnet for feature extraction from the input
    images, followed by one MLP per gender for attribute score prediction.
    To quantify the model's performance, we use the attribute classification
    metric described in the main paper. \ItoA
    achieves $60.7$ / $69.3 \%$ (fe-/male) of correctly predicted attributes,
    while our \StoA achieves $68.8$ / $76 \%$  on \caesar.
    Our explanation for this result is that it is hard for the \ItoA
    model to learn to correctly
    predict attributes independent of subject pose.
    Our approach works better, because it decomposes \threeD human estimation into predicting pose and
    shape.
    Networks are good at estimating pose even without
    GT shape \cite{li2021hybrik}.
    ``\modelname’s losses''
    affect only the shape branch. To minimize these losses,
    the network has to learn to correctly predict shape irrespective
    of pose variations.
}

\section{\modelname - \threeD Shape Regression from Images}
\qheading{Implementation details:}
To train \modelname,
each batch of training images contains 
$50\%$ images collected from model agency websites and 
$50\%$ images from \expose{'s} \cite{Choutas2020_expose} training set.
Note that the overall number of images of males and females in our collected model data differs significantly; images of female models are many more.
Therefore, we randomly sample a subset of female images so that, eventually, we get an equal number of male and female images.
We also use the BMI of each subject, when available, as a sampling weight for images. 
In this way, subjects with higher BMI are selected more often, due to their smaller number,
to avoid biasing the model towards the average BMI of the dataset.
Our pipeline is implemented in PyTorch \cite{pytorch} and we use the Adam \cite{adam} optimizer with a learning rate of $1e-4$.
\cameraready{We tune the weights of each loss term with
grid search on the \mmts and \hbw validation sets.}
Using a batch size of $48$, \modelname achieves the best performance on the \hbw validation set after 80k steps.

\section{Experiments}

\subsection{Metrics} \label{supmat:sec:metrics}

\begin{figure}[t!]
    \centering%
    \includegraphics[trim=118mm 019mm 112mm 015mm,clip=true, width=1.0\linewidth]{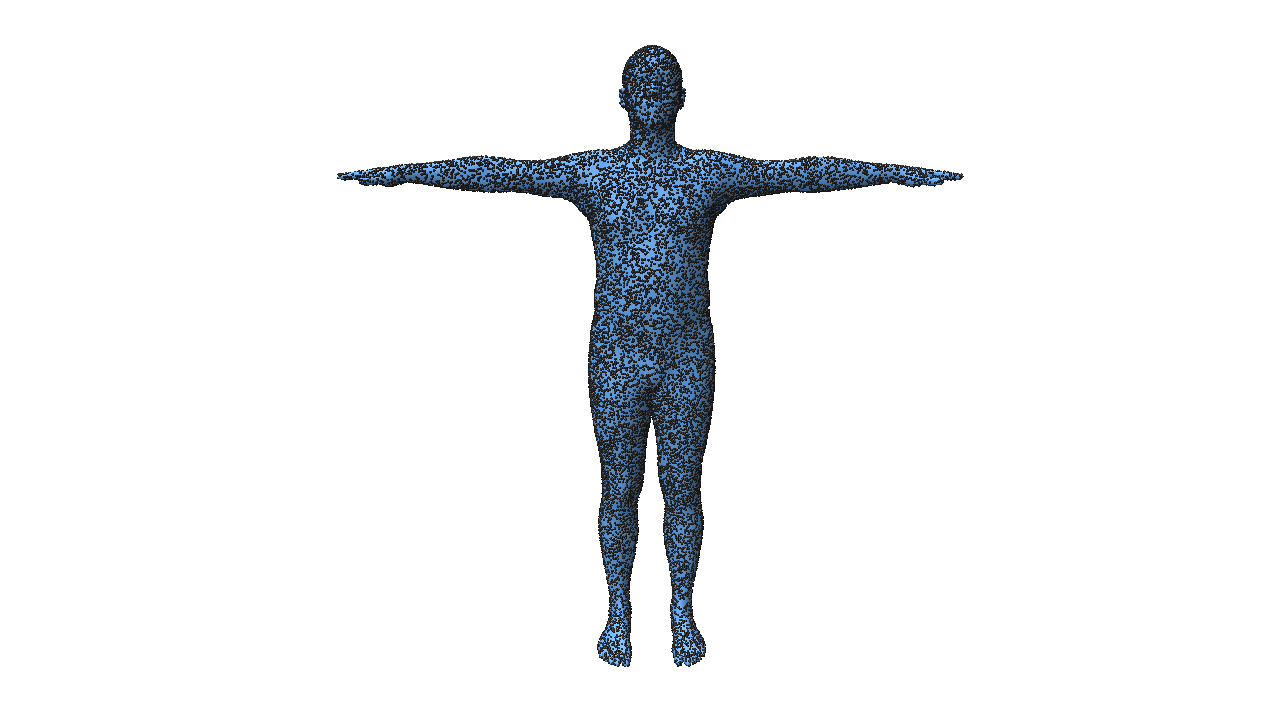}
    \caption{%
        The 20K body mesh surface points (in black) used to evaluated body shape
        estimation accuracy.}
    \label{fig:mesh_surface_points}
\end{figure}

\qheading{\vtovHD}
\smplx has more than half of its vertices on the head.
Consequently, computing an error based on vertices
overemphasizes
the importance of the head.
To remove this bias, we also report the mean distance between $P=\npointshd$ mesh surface points; see \cref{fig:mesh_surface_points} for a visualization on the \groundtruth and estimated meshes.
For this, we uniformly sample the \smplx template mesh and
compute a sparse matrix $\sparseregressorsmplx \in \mathbb{R}^{P \times N}$
that regresses the mesh surface points from \smplx vertices $V$,
as $\mathbf{P} = \sparseregressorsmplx V$.

To use this metric in a mesh with different topology, \eg \smpl,
we simply need to compute the corresponding $\sparseregressorsmpl$.
For this, we align the \smpl model to the \smplx template mesh.
For each point sampled from the \smplx mesh surface,
we find the closest point on the aligned \smpl mesh surface.
To obtain the \smpl mesh surface points from \smpl vertices,
we again compute a sparse matrix, $\sparseregressorsmpl \in \mathbb{R}^{P \times \nverticessmpl}$.
The distance between the \smplx and \smpl mesh surface points on the template meshes is $0.073$ mm,
which is negligible.

Given two meshes $M_1$ and $M_2$ of topology $T_1$ and $T_2$ we obtain the mesh surface points $P_1 = \mathbf{H}_{T_1} U_1$ and $P_2 = \mathbf{H}_{T_2} U_2$, where $U_1$ and $U_2$ denote the vertices of the shaped zero posed (t-pose) meshes.
To compute the \vtovHD error we correct for translation $t = \bar P_2 - \bar P_1$ and define $$ \vtovHD(U_1, U_2) = || \mathbf{H}_{T_1} U_1 + t - \mathbf{H}_{T_2} U_2 ||_2^2 \text{.}$$

\begin{table}[t!]
    \renewcommand{\arraystretch}{\myarraystretch}
    \centering
    \scriptsize
    \begin{tabular}{lllll}
        \toprule
        \multicolumn{5}{c}{Mean absolute error (mm) $\downarrow$}                                     \\
        \midrule   %
        Method                                & Height      & Chest       & Waist       & Hips        \\

        \modelname-\heightVar                 & \textbf{52} & 113         & 172         & 108         \\
        \modelname-\heightVar\attrVar         & 60          & \textbf{64} & \textbf{96} & \textbf{77} \\ %
        \modelname-\circVar                   & 119         & 66          & \textbf{70} & 70          \\
        \modelname-\circVar\attrVar           & \textbf{74} & \textbf{60} & 82          & \textbf{69} \\ %
        \modelname-\heightVar\circVar         & \textbf{54} & 62          & \textbf{72} & \textbf{69} \\
        \modelname-\heightVar\circVar\attrVar & 57          & \textbf{61} & 85          & 73          \\
        \bottomrule
    \end{tabular}
    \vspace{\vspaceTABaboveCaption}
    \caption{
        Leave-one-out evaluation on \mmts.}
    \label{tab:mmts_leave_one_out}
\end{table}
\begin{table}[t!]
    \renewcommand{\arraystretch}{\myarraystretch}
    \centering
    \scriptsize
    \begin{tabular}{llllll}
        \toprule
        \multicolumn{6}{c}{Mean absolute error (mm) $\downarrow$}                                                   \\
        \midrule
        Method                                & Height      & Chest       & Waist       & Hips        & \vtovHD     \\

        \modelname-\heightVar                 & 54          & 90          & 77          & \textbf{54} & 22          \\
        \modelname-\heightVar\attrVar         & \textbf{49} & \textbf{62} & \textbf{71} & 58          & \textbf{20} \\ %
        \modelname-\circVar                   & 72          & 65          & \textbf{77} & 60          & 26          \\
        \modelname-\circVar\attrVar           & \textbf{54} & \textbf{69} & 78          & \textbf{58} & \textbf{22}
        \\ %
        \modelname-\heightVar\circVar         & 53          & \textbf{61} & 77          & 55          & 23          \\
        \modelname-\heightVar\circVar\attrVar & \textbf{47} & 66          & \textbf{75} & \textbf{52} & \textbf{20} \\

        \bottomrule
    \end{tabular}
    \vspace{\vspaceTABaboveCaption}
    \caption{%
        Leave-one-out evaluation on the \hbw test set.}
    \label{tab:hbw_leave_one_out}
\end{table}
\begin{table}[t!]
    \renewcommand{\arraystretch}{\myarraystretch}
    \centering
    \scriptsize
    \resizebox{\columnwidth}{!}{
        \begin{tabular}{lllll}
            \toprule
            \multirow{2}{*}{Attribute} & \multicolumn{2}{c|}{Male} & \multicolumn{2}{c}{Female}                               \\ \cline{2-5}
                                       & MAE $\pm$ SD              & CCP                        & MAE $\pm$ SD    & CCP       \\ \hline
            Big                        & $0.25 \pm 0.18$           & $71.68\%$                  & $0.31 \pm 0.23$ & $70.00\%$ \\
            Broad Shoulders            & $0.26 \pm 0.20$           & $73.75\%$                  & $0.33 \pm 0.24$ & $63.90\%$ \\
            Long Legs                  & $0.23 \pm 0.17$           & $81.12\%$                  & $0.43 \pm 0.33$ & $58.05\%$ \\
            Long Neck                  & $0.27 \pm 0.21$           & $73.75\%$                  & $0.29 \pm 0.21$ & $69.51\%$ \\
            Long Torso                 & $0.27 \pm 0.20$           & $70.80\%$                  & $0.36 \pm 0.27$ & $62.68\%$ \\
            Muscular                   & $0.31 \pm 0.24$           & $69.03\%$                  & $0.26 \pm 0.21$ & $73.17\%$ \\
            Short                      & $0.28 \pm 0.22$           & $72.27\%$                  & $0.27 \pm 0.21$ & $67.56\%$ \\
            Short Arms                 & $0.20 \pm 0.15$           & $84.07\%$                  & $0.27 \pm 0.22$ & $72.20\%$ \\
            Tall                       & $0.27 \pm 0.22$           & $70.80\%$                  & $0.30 \pm 0.23$ & $70.98\%$ \\
            Average                    & $0.27 \pm 0.19$           & $78.76\%$                  & \na             & \na       \\
            Delicate Build             & $0.21 \pm 0.16$           & $78.17\%$                  & \na             & \na       \\
            Masculine                  & $0.23 \pm 0.18$           & $78.17\%$                  & \na             & \na       \\
            Rectangular                & $0.27 \pm 0.20$           & $80.24\%$                  & \na             & \na       \\
            Skinny Arms                & $0.25 \pm 0.19$           & $76.40\%$                  & \na             & \na       \\
            Soft Body                  & $0.32 \pm 0.23$           & $68.14\%$                  & \na             & \na       \\
            Large Breasts              & \na                       & \na                        & $0.31 \pm 0.23$ & $72.93\%$ \\
            Pear Shaped                & \na                       & \na                        & $0.32 \pm 0.22$ & $64.39\%$ \\
            Petite                     & \na                       & \na                        & $0.40 \pm 0.30$ & $61.95\%$ \\
            Skinny Legs                & \na                       & \na                        & $0.25 \pm 0.18$ & $81.22\%$ \\
            Slim Waist                 & \na                       & \na                        & $0.30 \pm 0.23$ & $71.71\%$ \\
            Feminine                   & \na                       & \na                        & $0.26 \pm 0.20$ & $73.41\%$ \\
            \bottomrule
        \end{tabular}
    }
    \vspace{\vspaceTABaboveCaption}
    \caption{%
        S2A evaluation. We report mean, standard deviation and percentage of correctly predicted classes per attribute on \cmts test set.
    }
    \label{tab:s2a_quant_result}
\end{table}

\subsection{Shape Estimation}

\qheading{\AtoS and its variations}
For completeness, \Cref{supmat:tab:a2s_quant_result_cmts} shows the results of the female \AtoS models in addition to the male ones. The male results are also presented in the main manuscript.
Note that attributes improve shape reconstruction across the board.
For example, in terms of \vtovHD,
\colorAHtoS
is
better than just
\colorHtoS
,
\colorAHWtoS
is
better than just
\colorHWtoS
.
It should be emphasized that even when many measurements are used as input features,
\ie height, weight, and chest/waist/hip circumference, adding attributes still
improves the shape estimate, \eg \mbox{%
    {\colorheight{H}}{\colorweight{W}}{\colorcirc{C}}2S
} vs. \mbox{{\colorattr{A}}{\colorheight{H}}{\colorweight{W}}{\colorcirc{C}}2S}.

\begin{table}
    \renewcommand{\arraystretch}{\myarraystretch}
    \centering
    \scriptsize
    \resizebox{\columnwidth}{!}{
        \setlength\tabcolsep{4pt}
        \begin{tabular}{llllllll}
            \toprule
             & Method   & \vtovHD         & Height      & Weight      & Chest       & Waist        & Hips        \\
             & -        & (mm)            & (mm)        & (kg)        & (mm)        & (mm)         & (mm)        \\
            \midrule
            \multirow{9}*{\rotatebox{90}{female}}
             & \AtoS    & $10.9 \pm 5.2$  & $27 \pm 21$ & $5 \pm 5$   & $30 \pm 26$ & $32 \pm 31$  & $28 \pm 22$ \\
             & \HtoS    & $12.8 \pm 7.0$  & $5 \pm 5$   & $12 \pm 11$ & $93 \pm 72$ & $101 \pm 88$ & $60 \pm 52$ \\
             & \AHtoS   & $7.2 \pm 2.8$   & $4 \pm 3$   & $3 \pm 4$   & $27 \pm 23$ & $29 \pm 28$  & $23 \pm 19$ \\
             & \HWtoS   & $7.9 \pm 3.2$   & $5 \pm 5$   & $1 \pm 1$   & $25 \pm 22$ & $22 \pm 18$  & $26 \pm 25$ \\
             & \AHWtoS  & $6.4 \pm 2.5$   & $4 \pm 3$   & $1 \pm 1$   & $14 \pm 12$ & $14 \pm 12$  & $17 \pm 14$ \\
             & \CtoS    & $19.5 \pm 10.8$ & $58 \pm 46$ & $8 \pm 6$   & $54 \pm 36$ & $57 \pm 42$  & $47 \pm 36$ \\
             & \ACtoS   & $9.6 \pm 4.3$   & $24 \pm 18$ & $3 \pm 2$   & $18 \pm 15$ & $19 \pm 16$  & $19 \pm 14$ \\
             & \HCtoS   & $7.3 \pm 2.8$   & $5 \pm 5$   & $2 \pm 2$   & $19 \pm 16$ & $16 \pm 14$  & $15 \pm 13$ \\
             & \AHCtoS  & $6.3 \pm 2.4$   & $4 \pm 3$   & $1 \pm 1$   & $15 \pm 12$ & $14 \pm 12$  & $14 \pm 12$ \\
             & \HWCtoS  & $7.2 \pm 2.9$   & $5 \pm 5$   & $1 \pm 1$   & $14 \pm 12$ & $13 \pm 11$  & $14 \pm 12$ \\
             & \AHWCtoS & $6.2 \pm 2.4$   & $4 \pm 3$   & $1 \pm 1$   & $11 \pm 9$  & $12 \pm 10$  & $13 \pm 11$ \\ \hline

            \multirow{9}*{\rotatebox{90}{male}}
             & \AtoS    & $11.1 \pm 5.2$  & $29 \pm 21$ & $5 \pm 4$   & $30 \pm 22$ & $32 \pm 24$  & $28 \pm 21$ \\
             & \HtoS    & $12.1 \pm 6.1$  & $5 \pm 4$   & $11 \pm 11$ & $81 \pm 66$ & $102 \pm 87$ & $40 \pm 33$ \\
             & \AHtoS   & $6.8 \pm 2.3$   & $4 \pm 3$   & $3 \pm 3$   & $27 \pm 21$ & $29 \pm 23$  & $24 \pm 18$ \\
             & \HWtoS   & $8.1 \pm 2.7$   & $5 \pm 4$   & $1 \pm 1$   & $24 \pm 17$ & $26 \pm 20$  & $21 \pm 18$ \\
             & \AHWtoS  & $6.3 \pm 2.1$   & $4 \pm 3$   & $1 \pm 1$   & $19 \pm 15$ & $19 \pm 14$  & $20 \pm 16$ \\
             & \CtoS    & $19.7 \pm 11.1$ & $59 \pm 47$ & $9 \pm 8$   & $55 \pm 41$ & $63 \pm 49$  & $37 \pm 28$ \\
             & \ACtoS   & $9.6 \pm 4.4$   & $25 \pm 19$ & $3 \pm 3$   & $23 \pm 19$ & $21 \pm 17$  & $18 \pm 14$ \\
             & \HCtoS   & $7.7 \pm 2.6$   & $5 \pm 4$   & $2 \pm 2$   & $28 \pm 23$ & $18 \pm 15$  & $13 \pm 11$ \\
             & \AHCtoS  & $6.0 \pm 2.0$   & $4 \pm 3$   & $2 \pm 2$   & $21 \pm 17$ & $17 \pm 14$  & $13 \pm 10$ \\
             & \HWCtoS  & $7.3 \pm 2.6$   & $5 \pm 4$   & $1 \pm 1$   & $20 \pm 15$ & $14 \pm 12$  & $13 \pm 11$ \\
             & \AHWCtoS & $5.8 \pm 2.0$   & $4 \pm 3$   & $1 \pm 1$   & $16 \pm 13$ & $13 \pm 10$  & $13 \pm 10$ \\

            \bottomrule
        \end{tabular}
    }
    \vspace{\vspaceTABaboveCaption}
    \caption{
        Results of \AtoS and its variations on \cmts test set, in mm or kg. Trained with gender-specific \smplx model.
    }
    \label{supmat:tab:a2s_quant_result_cmts}
\end{table}

\qheading{Attribute/Measurement ablation}
To investigate the extent to which attributes can replace ground truth measurements
in network training, we train \modelname's variations in a leave-one-out manner:
{\modelname-\colorheight{H}} uses only height and \modelname-{\colorcirc{C}}
only hip/waist/chest circumference.
We compare these models with \modelname-{\colorattr{A}}{\colorheight{H}} and
\modelname-{\colorattr{A}}{\colorcirc{C}},
which use attributes in addition to height and circumference measurements,
respectively. For completeness, we also evaluate
\modelname-{\colorheight{H}}{\colorcirc{C}}
and \modelname-{\colorattr{A}}{\colorheight{H}}{\colorcirc{C}},
which use all measurements; the latter also uses attributes.
The results are reported in \cref{tab:mmts_leave_one_out} (\mmts) and \cref{tab:hbw_leave_one_out} (\hbw).
\TODO{The tables show that attributes are an adequate replacement for measurements.
For example, in \cref{tab:mmts_leave_one_out}, the height
(\modelname-{\colorcirc{C}} vs.~\modelname-{\colorcirc{C}}{\colorattr{A}})
and circumference errors
(\modelname-{\colorheight{H}} vs.~\modelname-\colorAH
) are reduced significantly
when attributes are taken into account. On \hbw, the \vtovHD errors are equal or lower,
when attribute information is used, see \cref{tab:hbw_leave_one_out}.
Surprisingly, seeing attributes improves the height error in all three variations.
This suggests that training on model images introduces a bias that \AtoS antagonizes.
}

\qheading{\StoA}
\Cref{tab:s2a_quant_result} shows the results of \StoA in detail. \TODO{All attributes are classified correctly with an accuracy of at least $58.05\%$ (females) and $68.14\%$ (males). The probability of randomly guessing the correct class is 20\%.}

\qheading{\colorAHWCtoS and \colorAHWC noise}
\cameraready{%
    To evaluate \colorAHWCtoS's robustness to noise in the input,
    we fit \colorAHWCtoS using the per-rater scores instead
    of the average score. The \vtovHD $\downarrow$ error only increases
    by $1.0$ mm to $6.8$ when using the per-rater scores.
}
\subsection{Pose evaluation}
\label{sec:pose_eval}

\qheading{\threeD Poses in the Wild (\threedpw) \cite{vonMarcard2018}} 
This dataset is mainly useful for evaluating 
body \emph{pose} accuracy since it contains few subjects and limited body shape variation.
The test set contains a limited set of $5$ subjects in indoor/outdoor videos with everyday clothing. \TODO{All subjects were scanned to obtain their ground-truth body shape}. The body poses are pseudo \groundtruth~\smpl fits, recovered from images and IMUs. We convert pose and shape to \smplX for evaluation.

We evaluate \modelname on \threedpw to report pose estimation accuracy (\cref{tab:3dpw}). \modelname's pose accuracy is slightly behind \expose which also uses \smplX.  \lookat{\modelname's performance is better than \hmr \cite{Kanazawa2018_hmr} and \straps \cite{sengupta2020straps}. However, \modelname does not outperform recent pose estimation methods, \eg HybrIK \cite{li2021hybrik}. We assume that \modelname's pose estimation accuracy on \threedpw can be improved by (1) adding data from the \threedpw training set (similar to Sengupta \etal \cite{sengupta2021hierarchicalICCV} who sample poses from \threedpw training set) and (2) creating pseudo ground-truth fits for the model data.}

\begin{table}
\renewcommand{\arraystretch}{1.2} 
\centering
\scriptsize
\begin{tabular}{llll}
    \toprule
    & Model & MPJPE & PA-MPJPE \\
    \midrule
    \hmr    \cite{Kanazawa2018_hmr}                     & \smpl     & 130      & 81.3        \\
	\spin \cite{Kolotouros2019_spin}                    & \smpl     & 96.9     & 59.2        \\
	TUCH \cite{mueller2021tuch}                         & \smpl     & 84.9     & 55.5        \\
	EFT \cite{joo2020eft}                               & \smpl     & -        & 54.2        \\
    \hybrik \cite{li2021hybrik}                         & \smpl     & \textbf{80.0}     & \textbf{48.8}        \\
    \straps \cite{sengupta2020straps}*                   & \smpl     & -        & 66.8        \\
    Sengupta \etal \cite{sengupta2021probabilisticCVPR}* & \smpl     & -        & 61.0        \\
    Sengupta \etal \cite{sengupta2021hierarchicalICCV}*  & \smpl     & 84.9     & 53.6        \\
	\expose \cite{Choutas2020_expose}                   & \smplx    & 93.4     & 60.7        \\
    \modelname (ours)                                   & \smplx    & 95.2     & 62.6         \\ 
    \noalign{\smallskip}
    \bottomrule
\end{tabular}
\vspace{\vspaceTABaboveCaption}
\caption{Evaluation on \threedpw \cite{vonMarcard2018}. * uses body poses sampled from the \threedpw training set for training.
}
\label{tab:3dpw}
\end{table}

\subsection{Qualitative Results}
\begin{figure*}[th]
    \centering
    \includegraphics[width=0.240\textwidth,trim={0cm, 0cm, 0cm, 3cm}, clip]{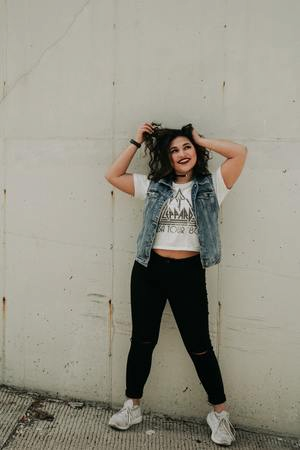}%
    \includegraphics[width=0.240\textwidth,trim={0cm, 0cm, 0cm, 3cm}, clip]{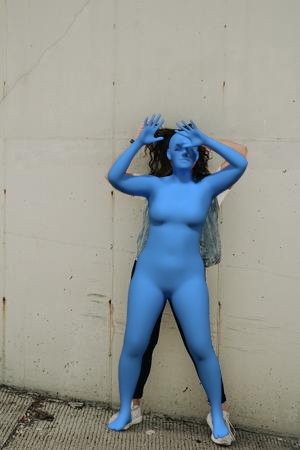}%
    \includegraphics[width=0.260\textwidth,trim={110mm, 050mm, 120mm, 130mm}, clip=true]{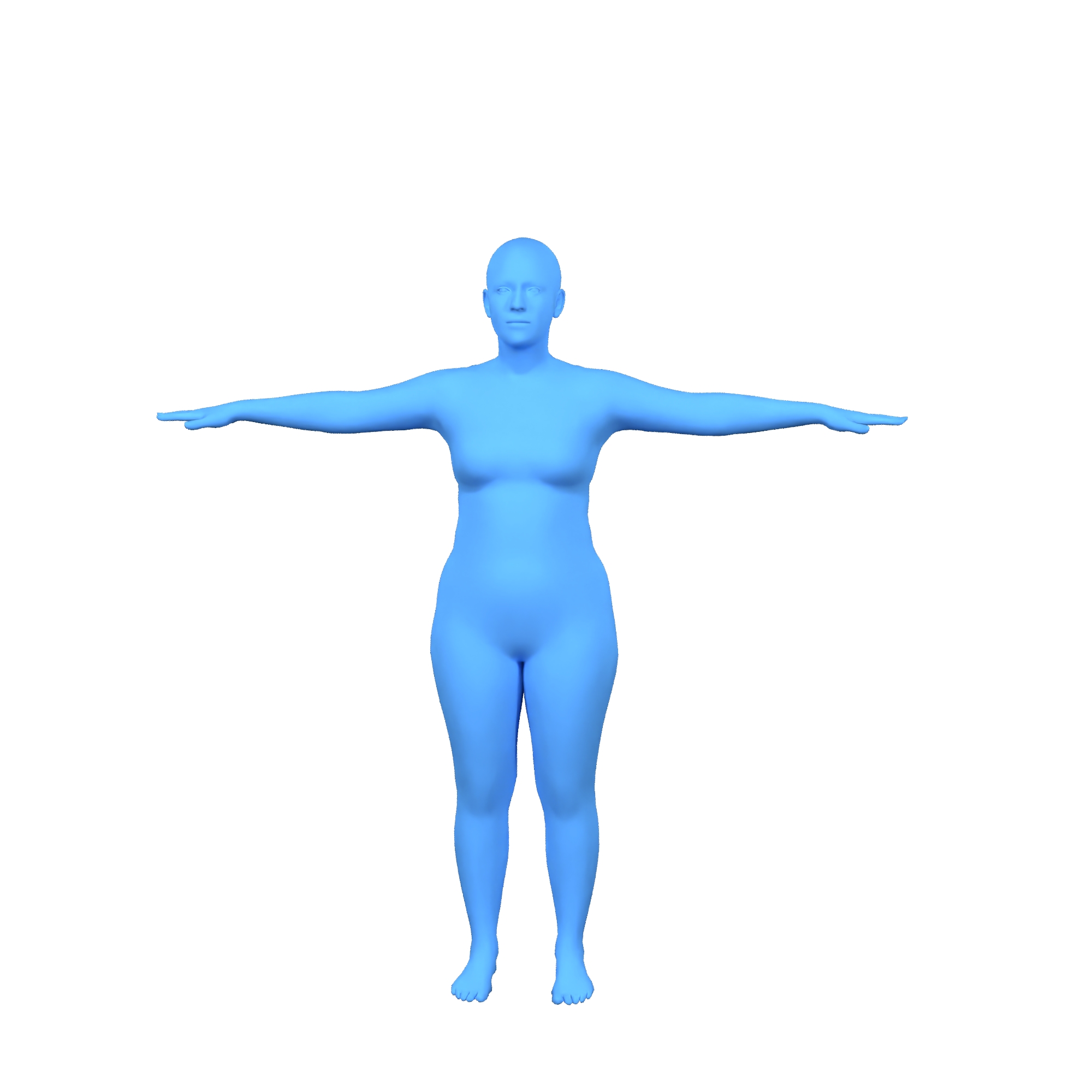}%
    \includegraphics[width=0.260\textwidth,trim={110mm, 050mm, 120mm, 130mm}, clip=true]{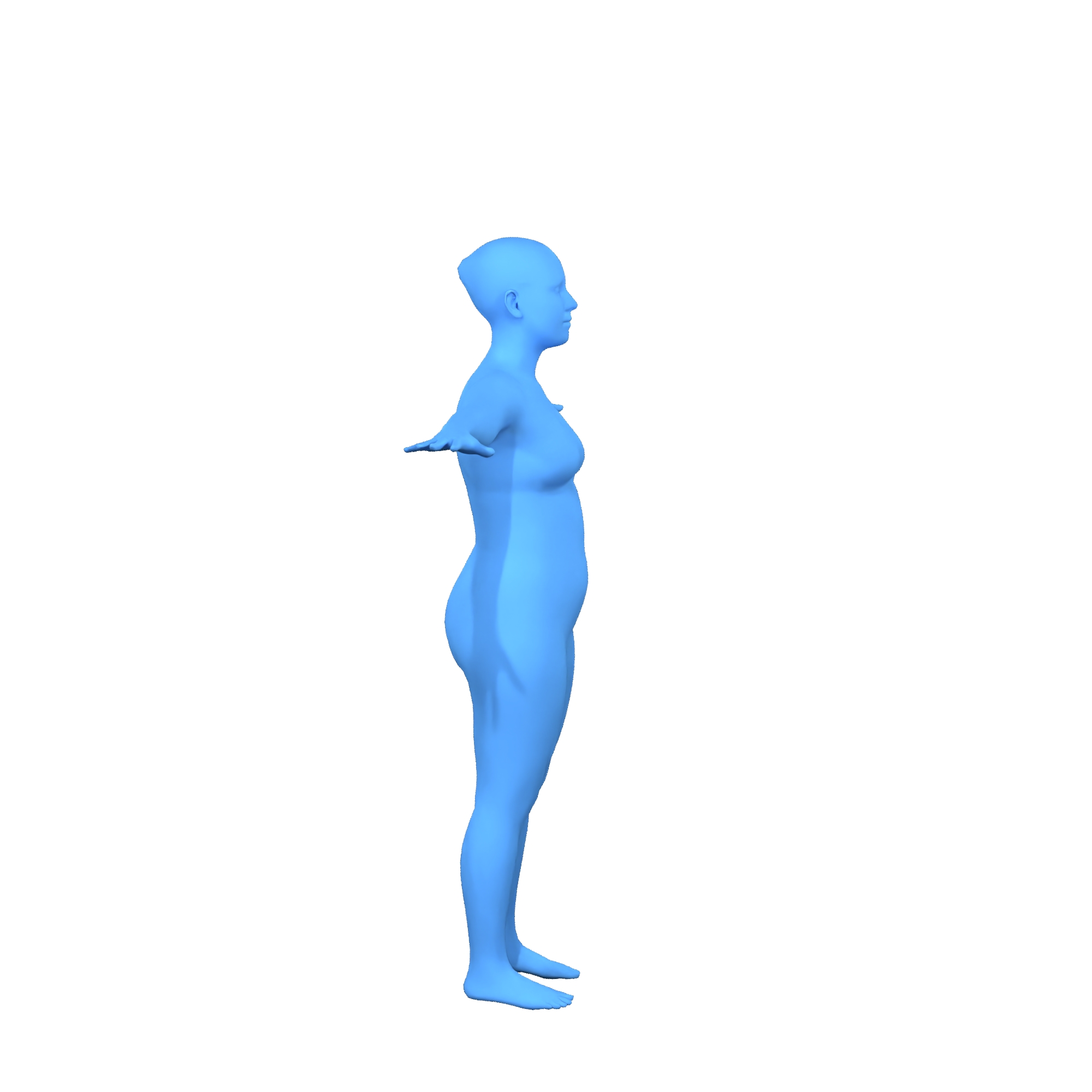}
    
    \includegraphics[width=0.240\textwidth,trim={0cm, 0cm, 0cm, 0cm}, clip]{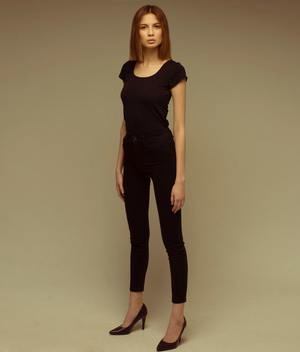}%
    \includegraphics[width=0.240\textwidth,trim={0cm, 0cm, 0cm, 0cm}, clip]{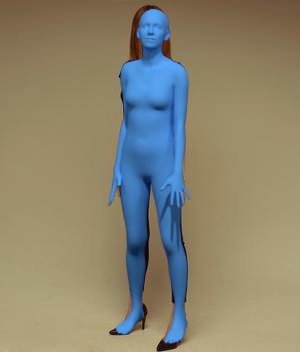}%
    \includegraphics[width=0.260\textwidth,trim={110mm, 050mm, 120mm, 130mm}, clip=true]{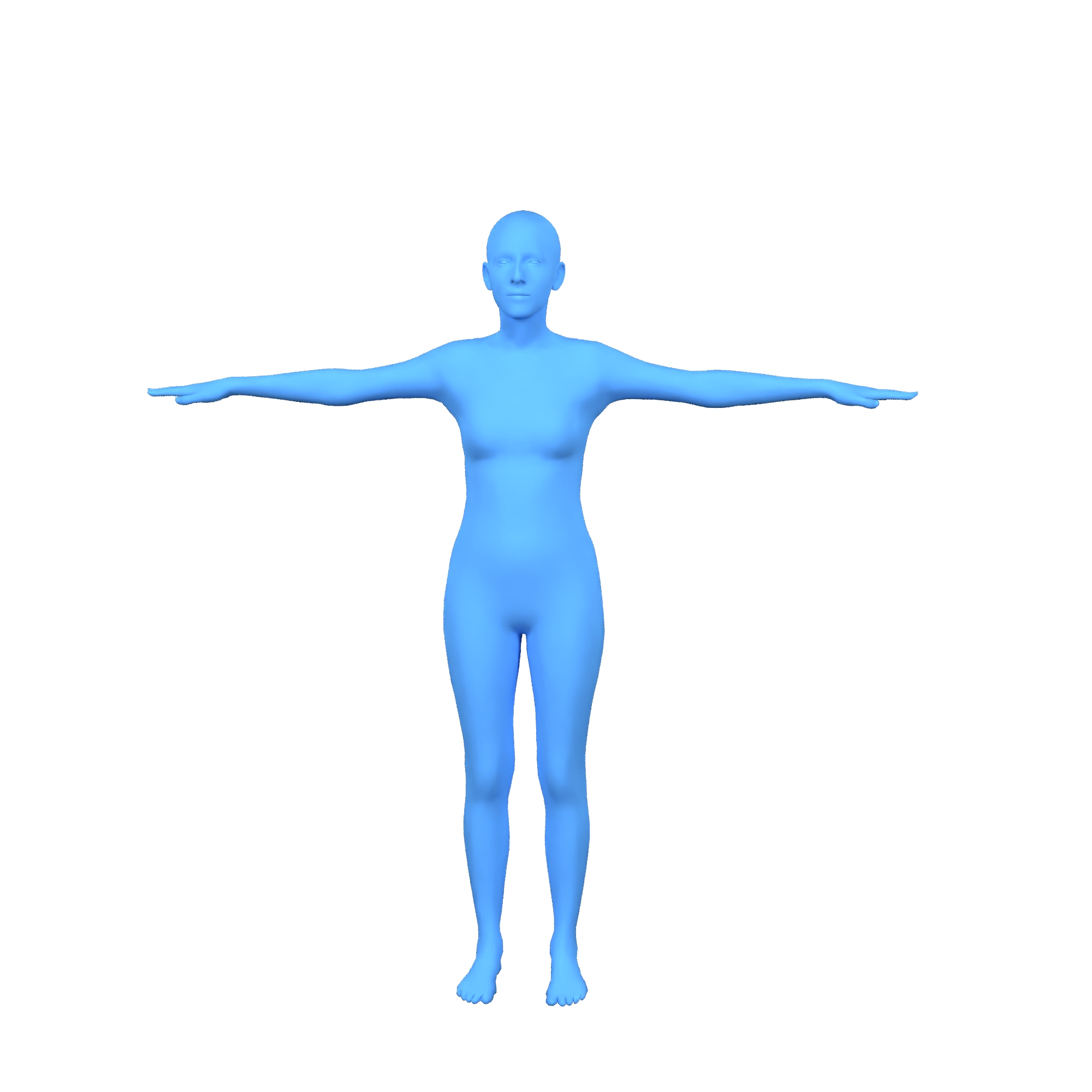}%
    \includegraphics[width=0.260\textwidth,trim={110mm, 050mm, 120mm, 130mm}, clip=true]{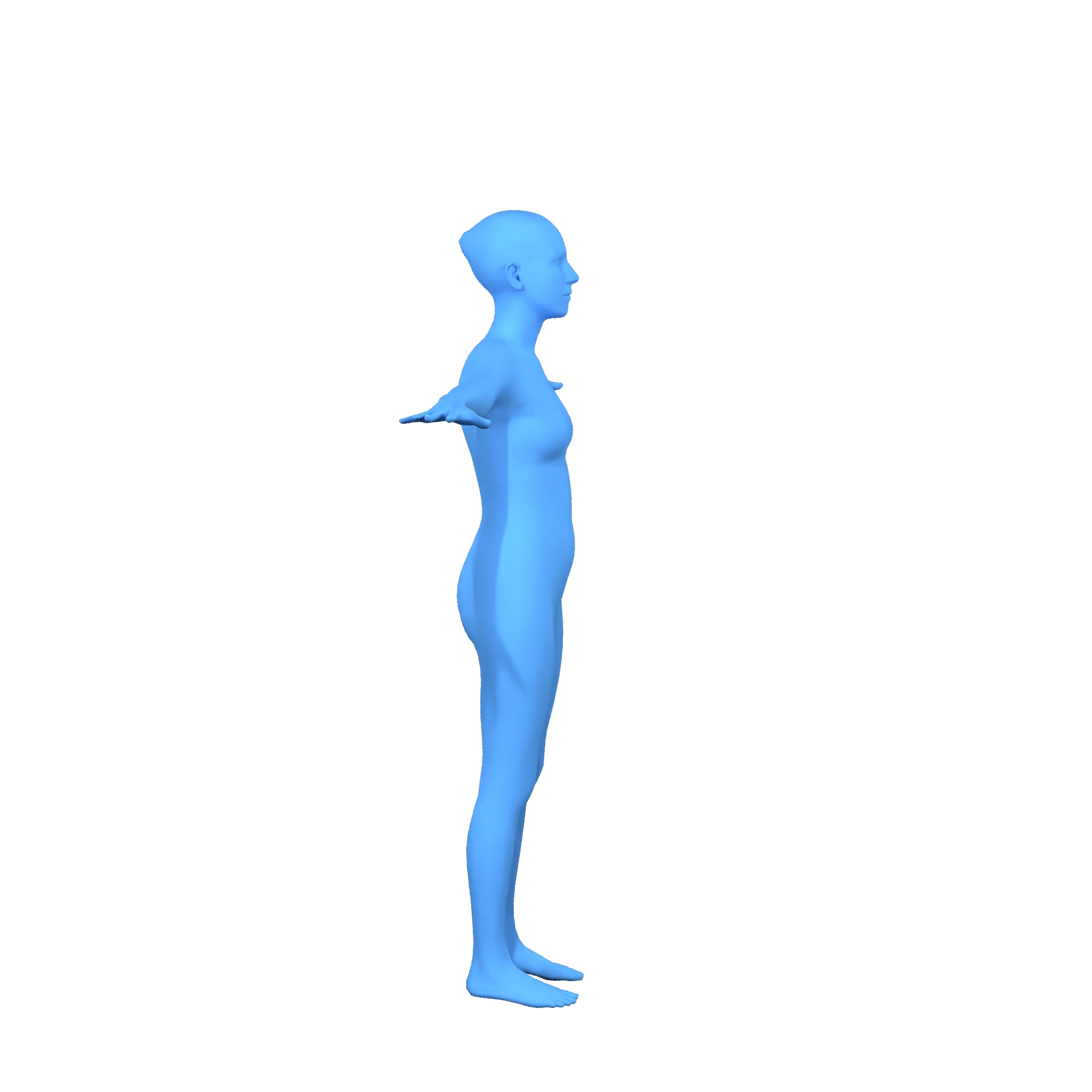}%
                
    \caption{Qualitative results of \modelname predictions for female bodies. (Cont.) }
    \label{fig:shapy_qual_pose_female}
\end{figure*}

\begin{figure*}
    \centering
        \includegraphics[width=0.250\linewidth,trim={0cm, 0cm, 0cm, 0cm}, clip]{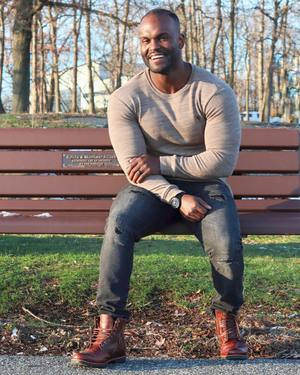}%
        \includegraphics[width=0.250\linewidth,trim={0cm, 0cm, 0cm, 0cm}, clip]{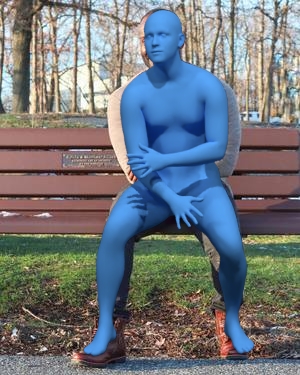}%
        \includegraphics[width=0.250\linewidth,trim={100mm, 050mm, 120mm, 110mm}, clip=true]{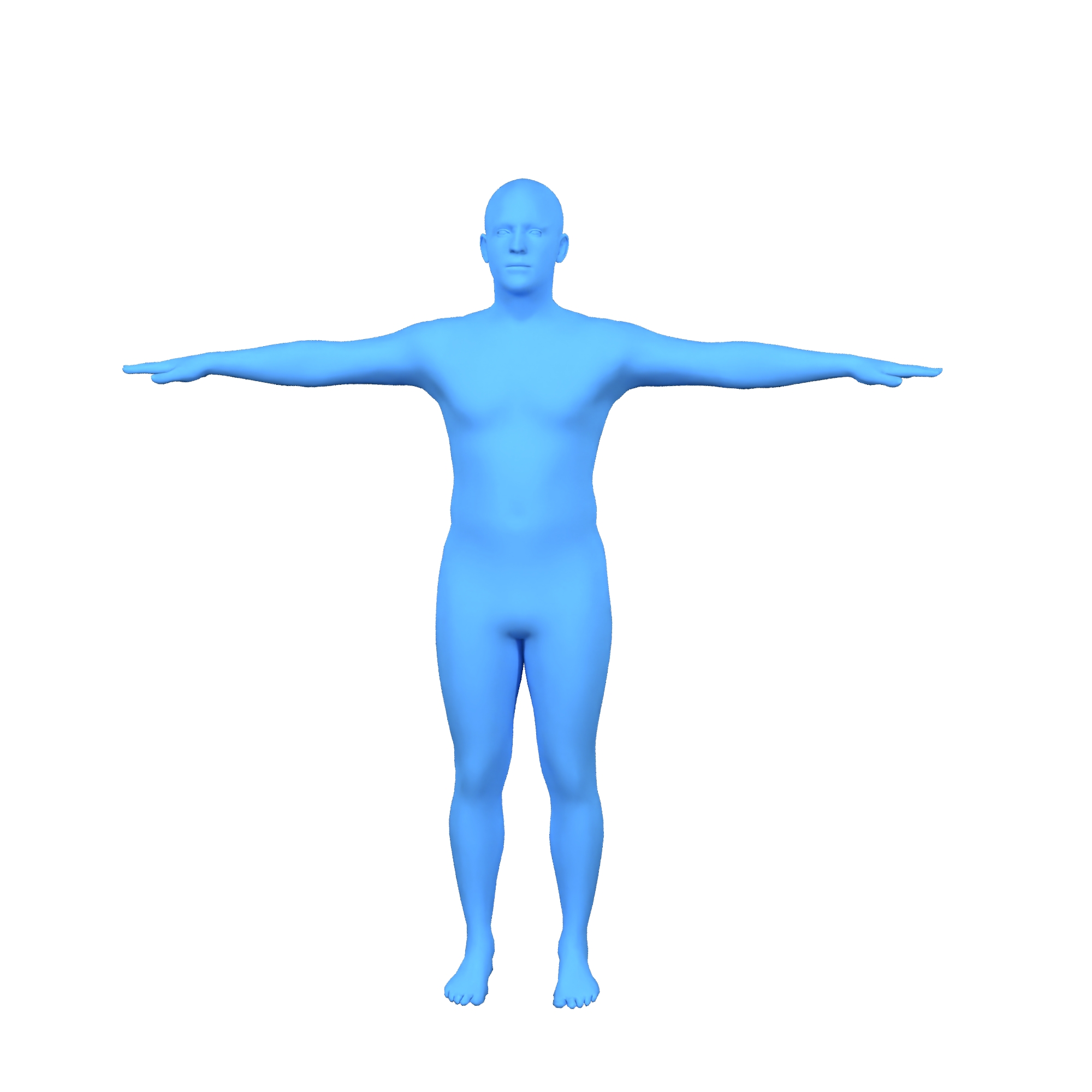}%
        \includegraphics[width=0.250\linewidth,trim={100mm, 050mm, 120mm, 110mm}, clip=true]{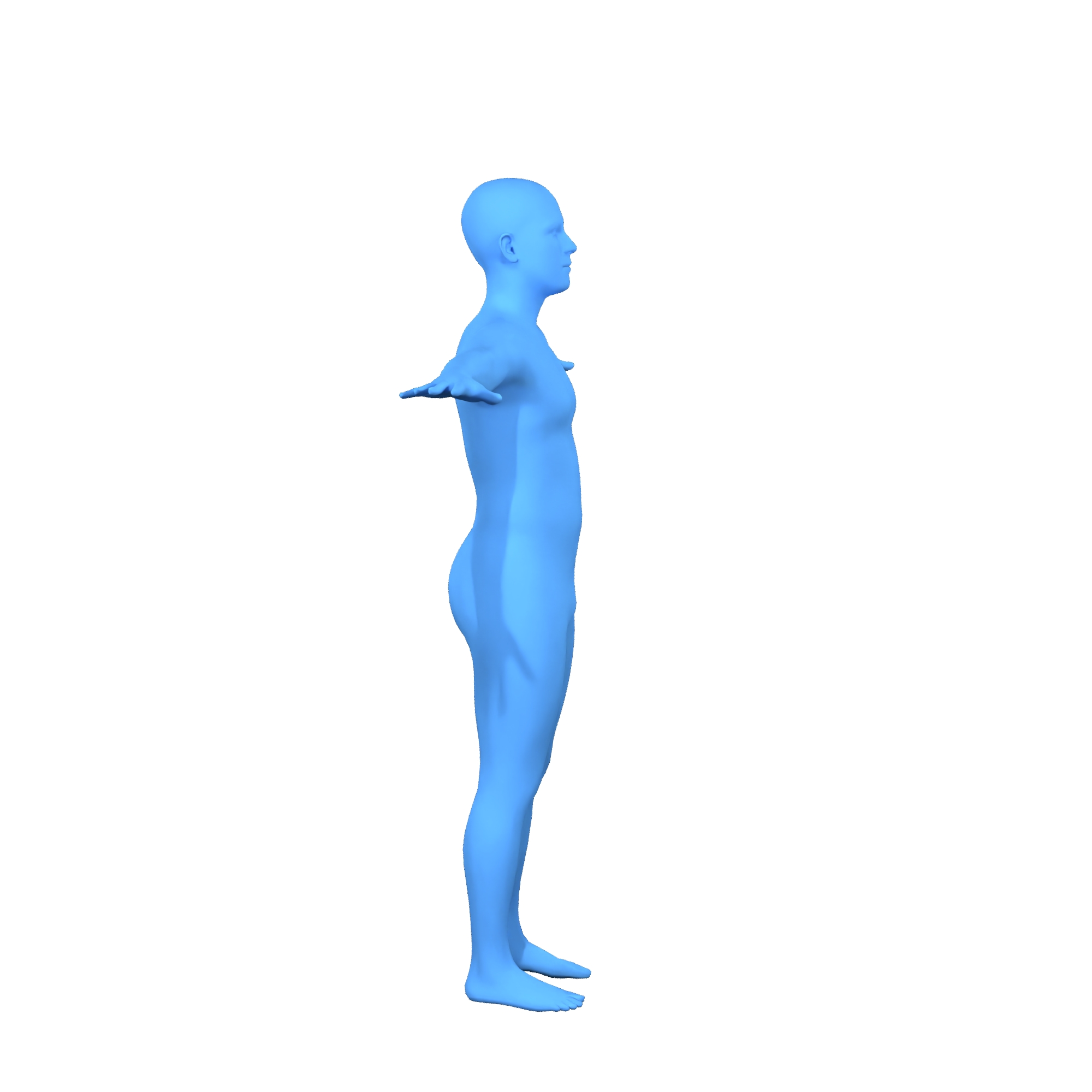}
        
        \includegraphics[width=0.240\linewidth,trim={0cm, 0cm, 0cm, 030mm}, clip]{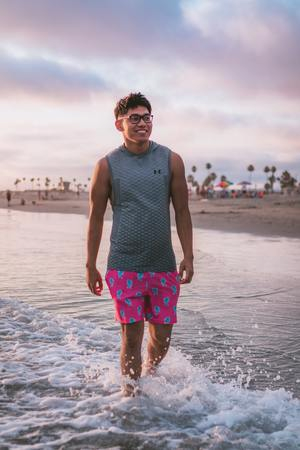}%
        \includegraphics[width=0.240\linewidth,trim={0cm, 0cm, 0cm, 030mm}, clip]{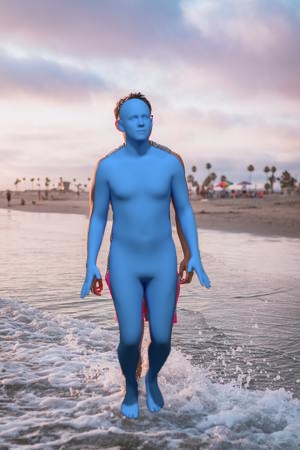}%
        \includegraphics[width=0.260\linewidth,trim={100mm, 050mm, 120mm, 110mm}, clip=true]{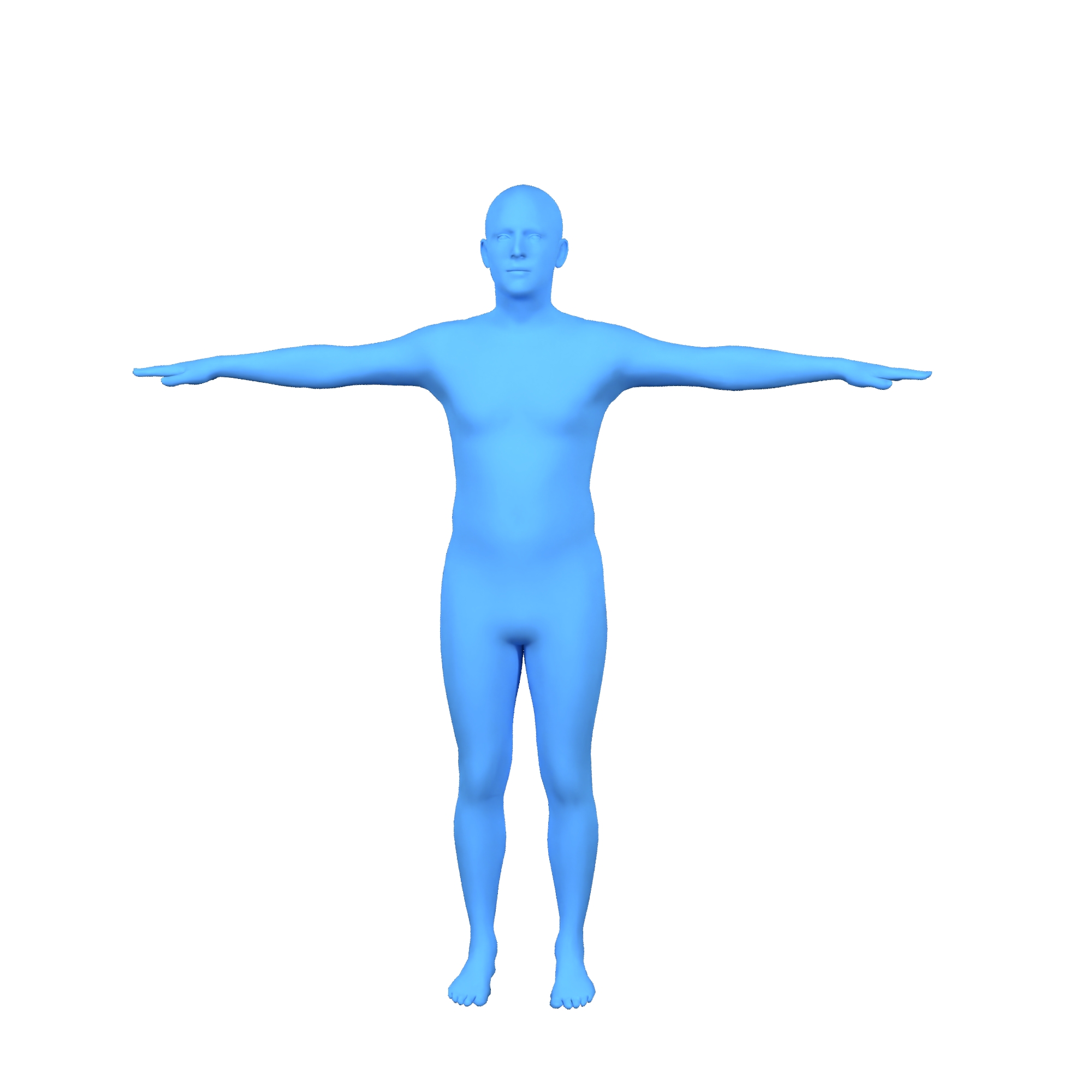}%
        \includegraphics[width=0.260\linewidth,trim={100mm, 050mm, 120mm, 110mm}, clip=true]{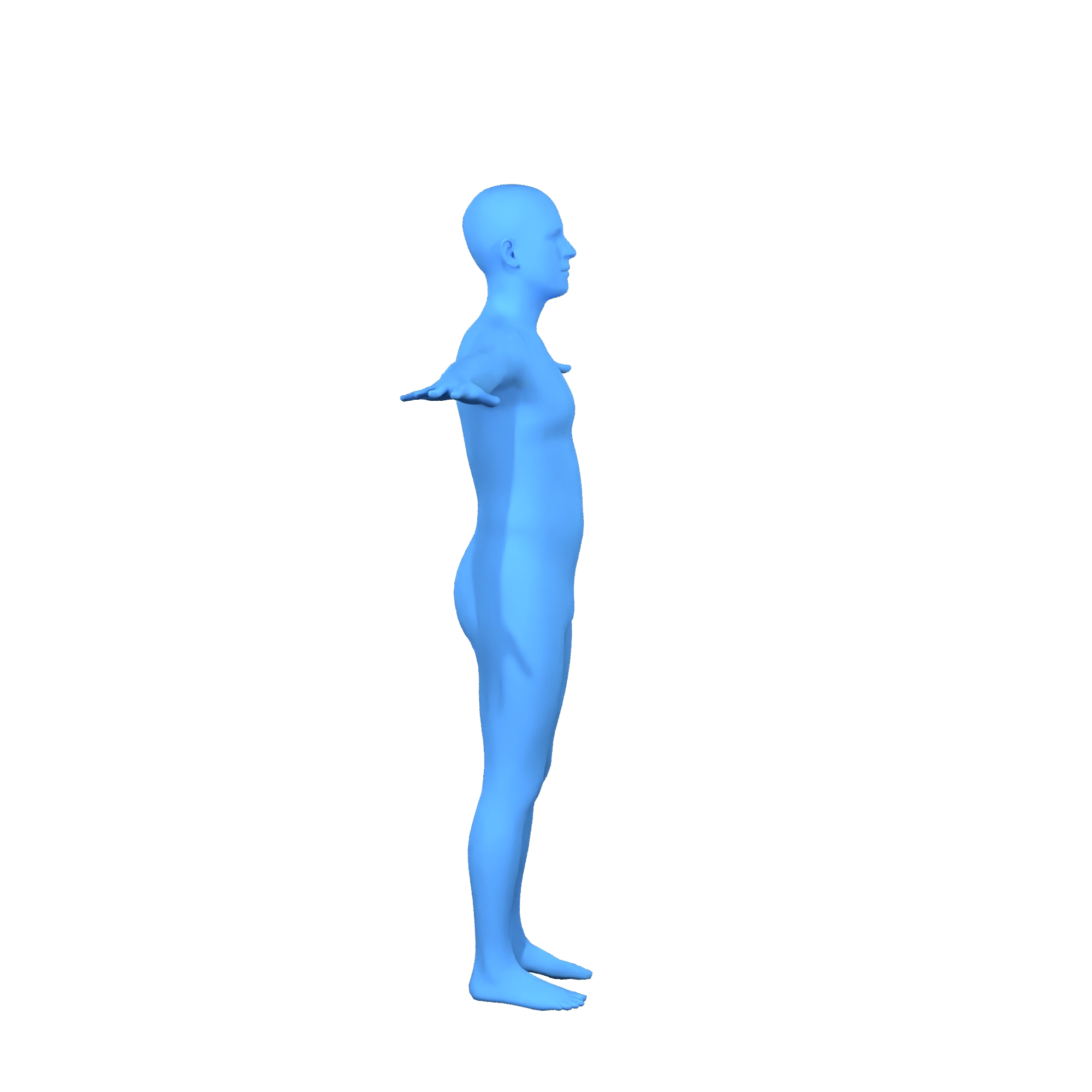}
    
    \vspace{-0.5em}
    \caption{Qualitative results of \modelname predictions for male bodies (Cont.) .}
    \label{fig:shapy_qual_pose_male}
\end{figure*}

\begin{figure*}[t]
    \centering
    \includegraphics[width=1.00\linewidth]{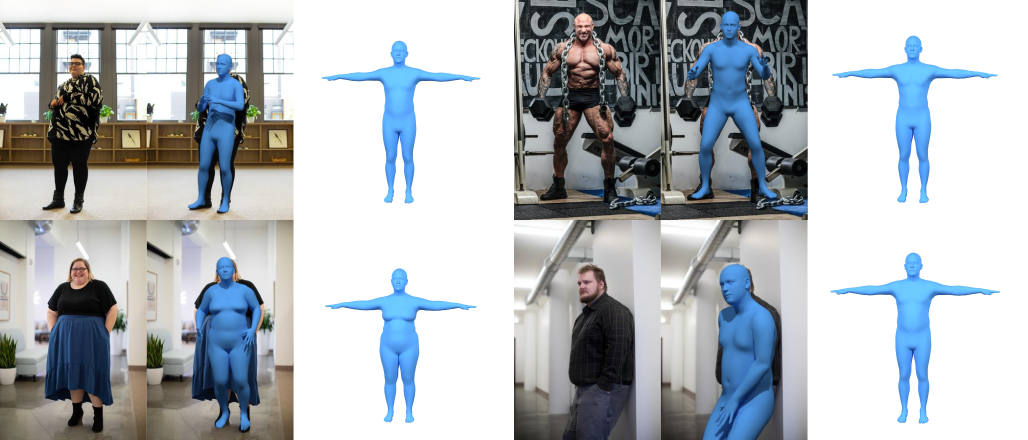}
    \caption{Failure cases. In the first example (upper left)
    \cameraready{the weight is underestimated}.
    Other failure cases of \modelname are
    muscular bodies (upper right) and 
    body shapes with high \bmi (second row).}
    \label{fig:failur_cases}
\end{figure*}

We show additional qualitative results in \cref{fig:shapy_qual_pose_female} and \cref{fig:shapy_qual_pose_male}. Failure cases are shown in \cref{fig:failur_cases}.
To deal with high-BMI bodies, we need to expand the set of training images and add additional shape attributes that are descriptive for high-BMI shapes.
Muscle definition on highly muscular bodies is not well represented by \smplx, nor do our attributes capture this.
The \modelname approach, however, could be used to capture this with a suitable
body model and more appropriate attributes.

\begin{figure*}[th]
    \centering
    \includegraphics[width=0.220\textwidth,trim={0cm, 1cm, 0cm, 2cm}, clip]{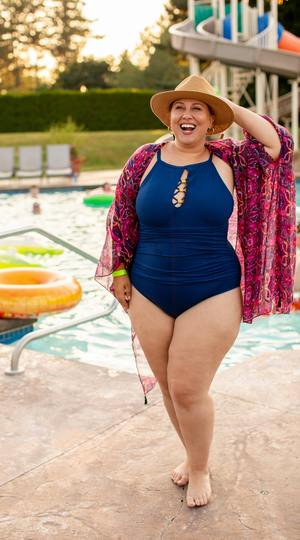}%
    \includegraphics[width=0.220\textwidth,trim={0cm, 1cm, 0cm, 2cm}, clip]{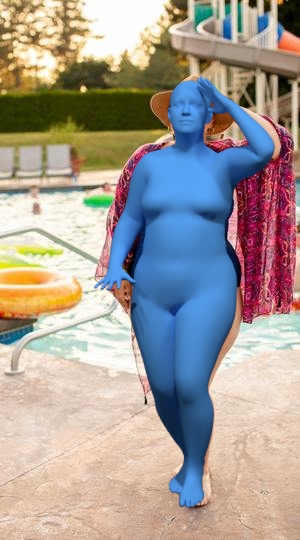}%
    \includegraphics[width=0.280\textwidth,trim={110mm, 050mm, 120mm, 150mm}, clip=true]{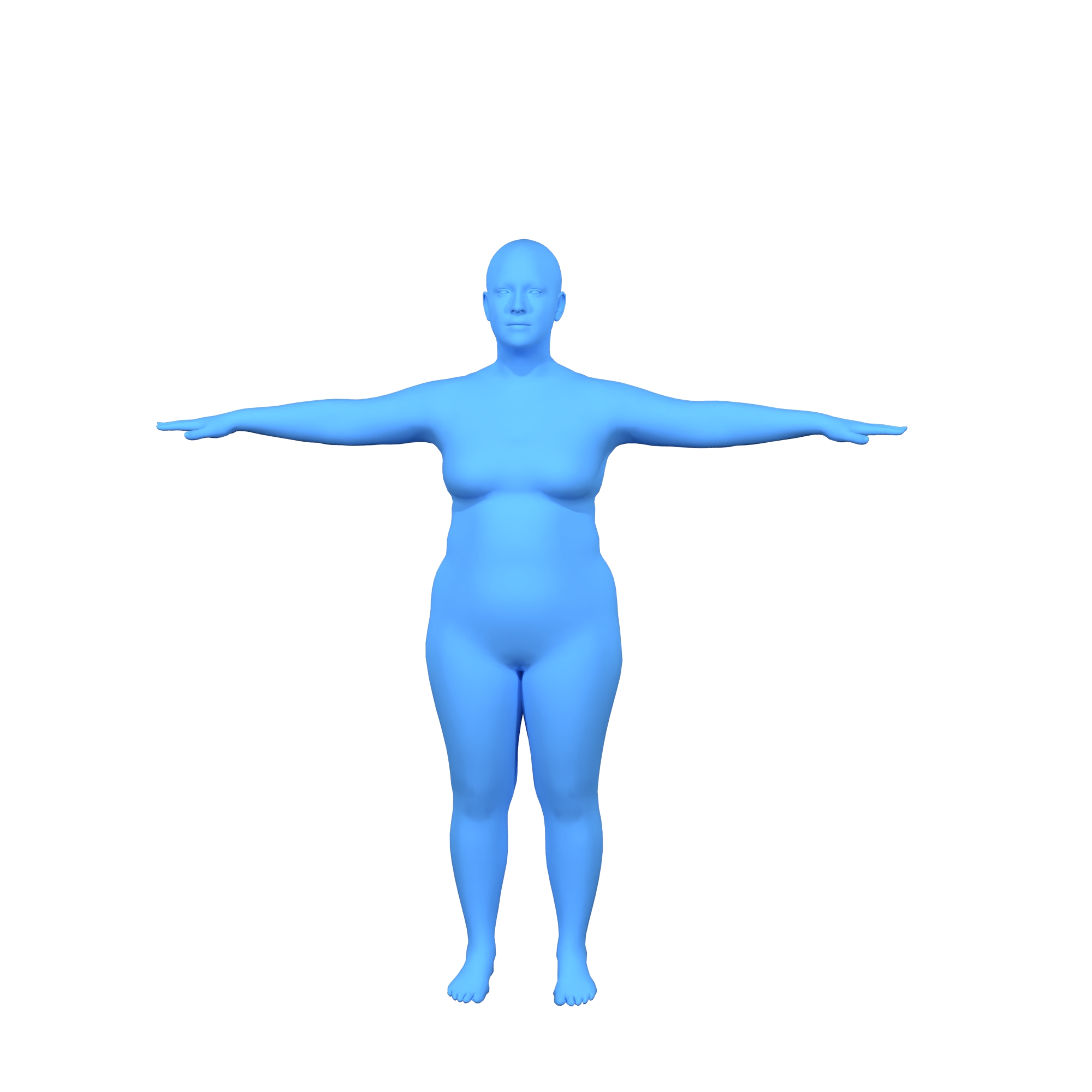}%
    \includegraphics[width=0.280\textwidth,trim={110mm, 050mm, 120mm, 150mm}, clip=true]{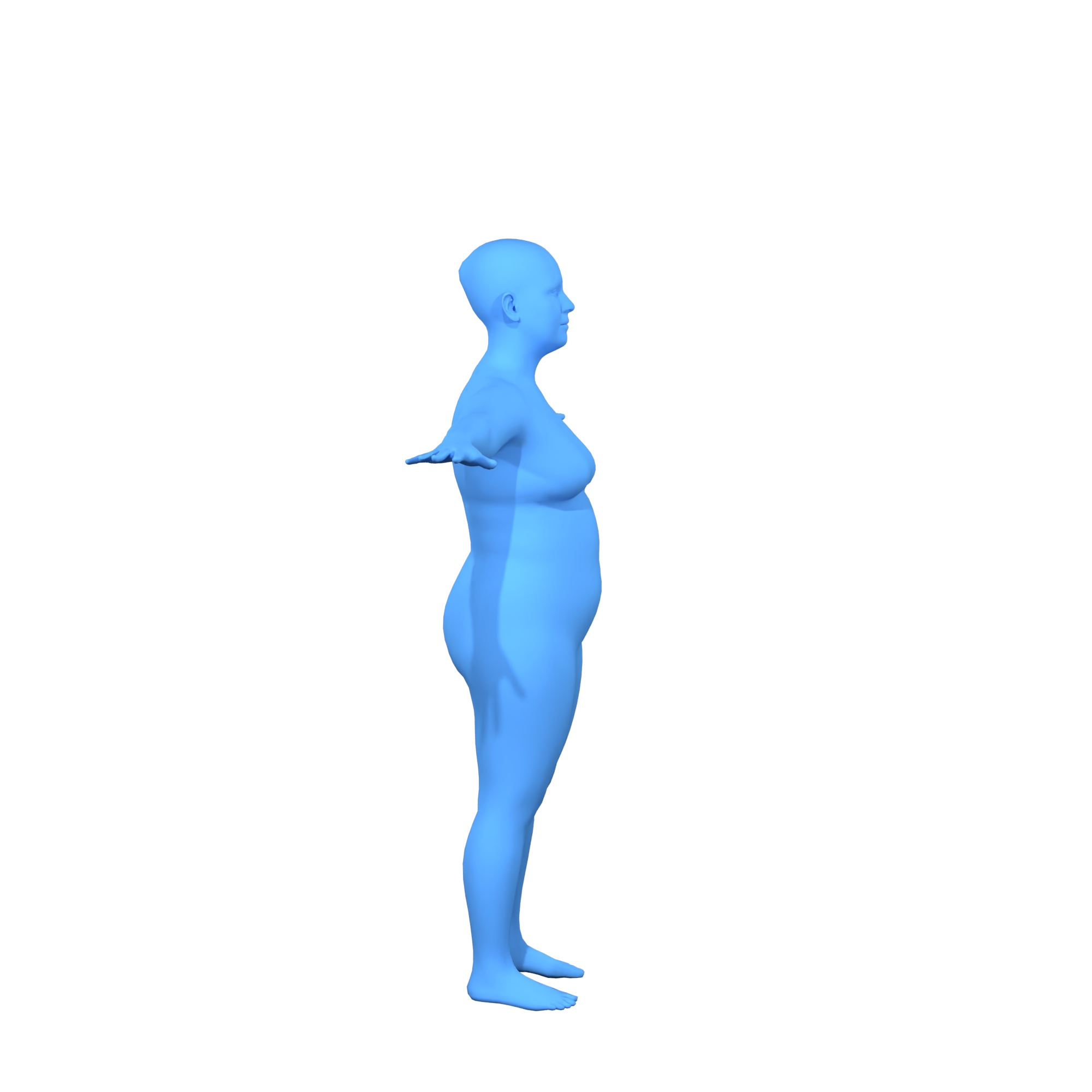}
    
    \includegraphics[width=0.220\textwidth,trim={0cm, 0cm, 0cm, 0cm}, clip]{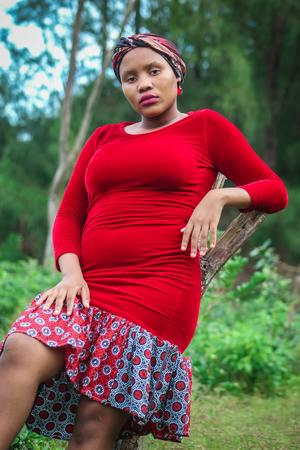}%
    \includegraphics[width=0.220\textwidth,trim={0cm, 0cm, 0cm, 0cm}, clip]{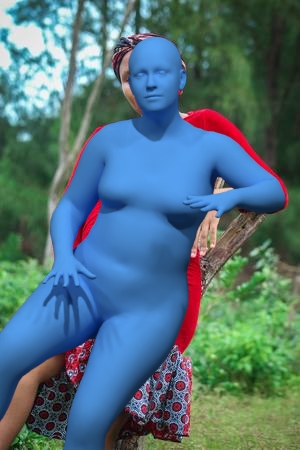}%
    \includegraphics[width=0.280\textwidth,trim={110mm, 050mm, 120mm, 130mm}, clip=true]{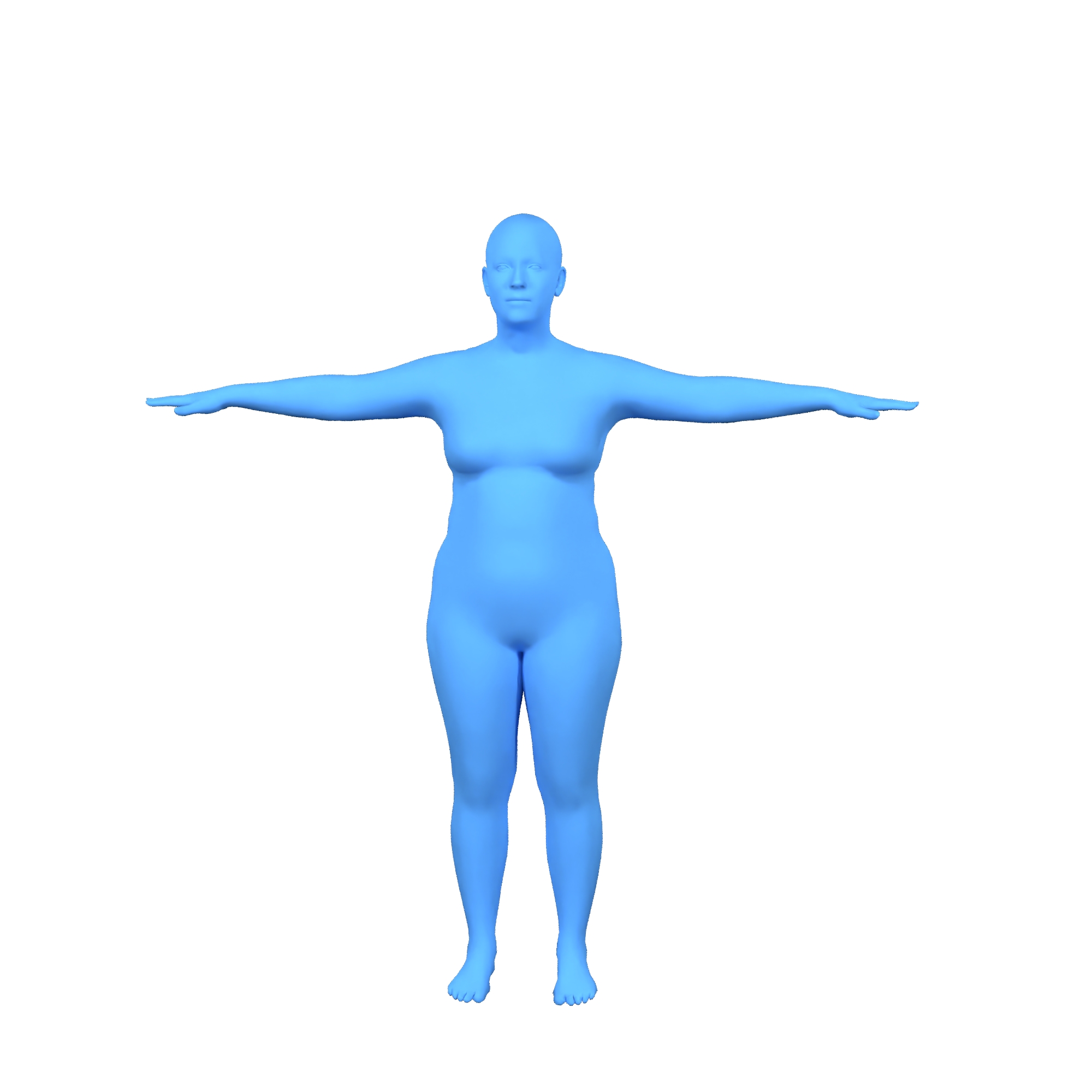}%
    \includegraphics[width=0.280\textwidth,trim={110mm, 050mm, 120mm, 130mm}, clip=true]{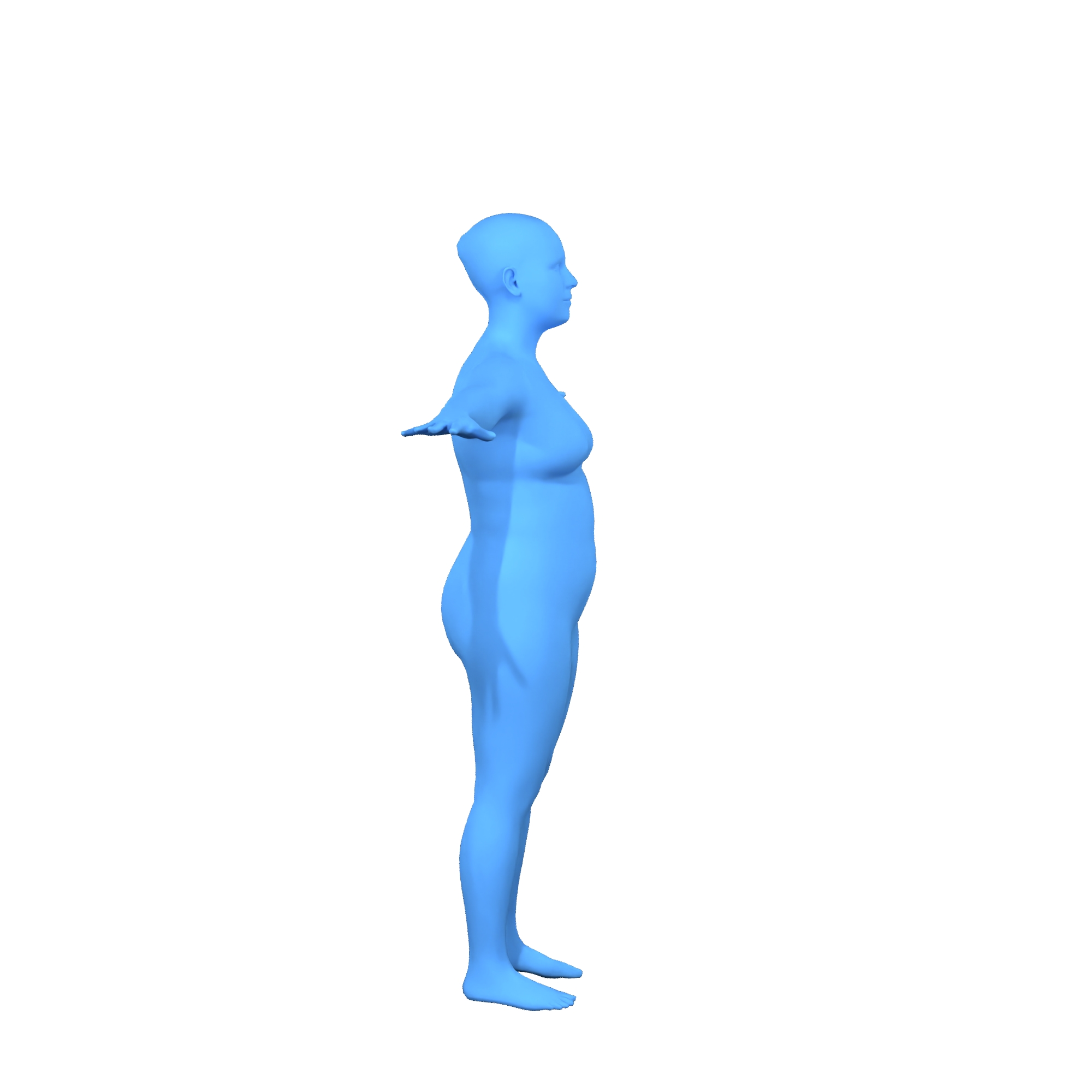}
    
    \includegraphics[width=0.250\textwidth,trim={0cm, 2cm, 0cm, 0cm}, clip]{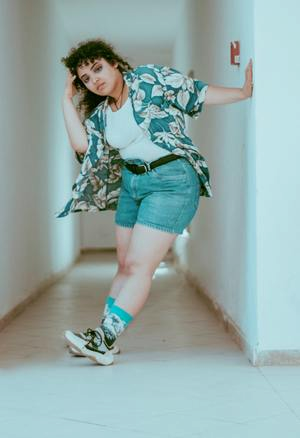}%
    \includegraphics[width=0.250\textwidth,trim={0cm, 2cm, 0cm, 0cm}, clip]{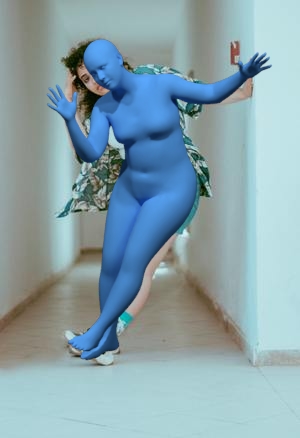}%
    \includegraphics[width=0.250\textwidth,trim={110mm, 050mm, 120mm, 130mm}, clip=true]{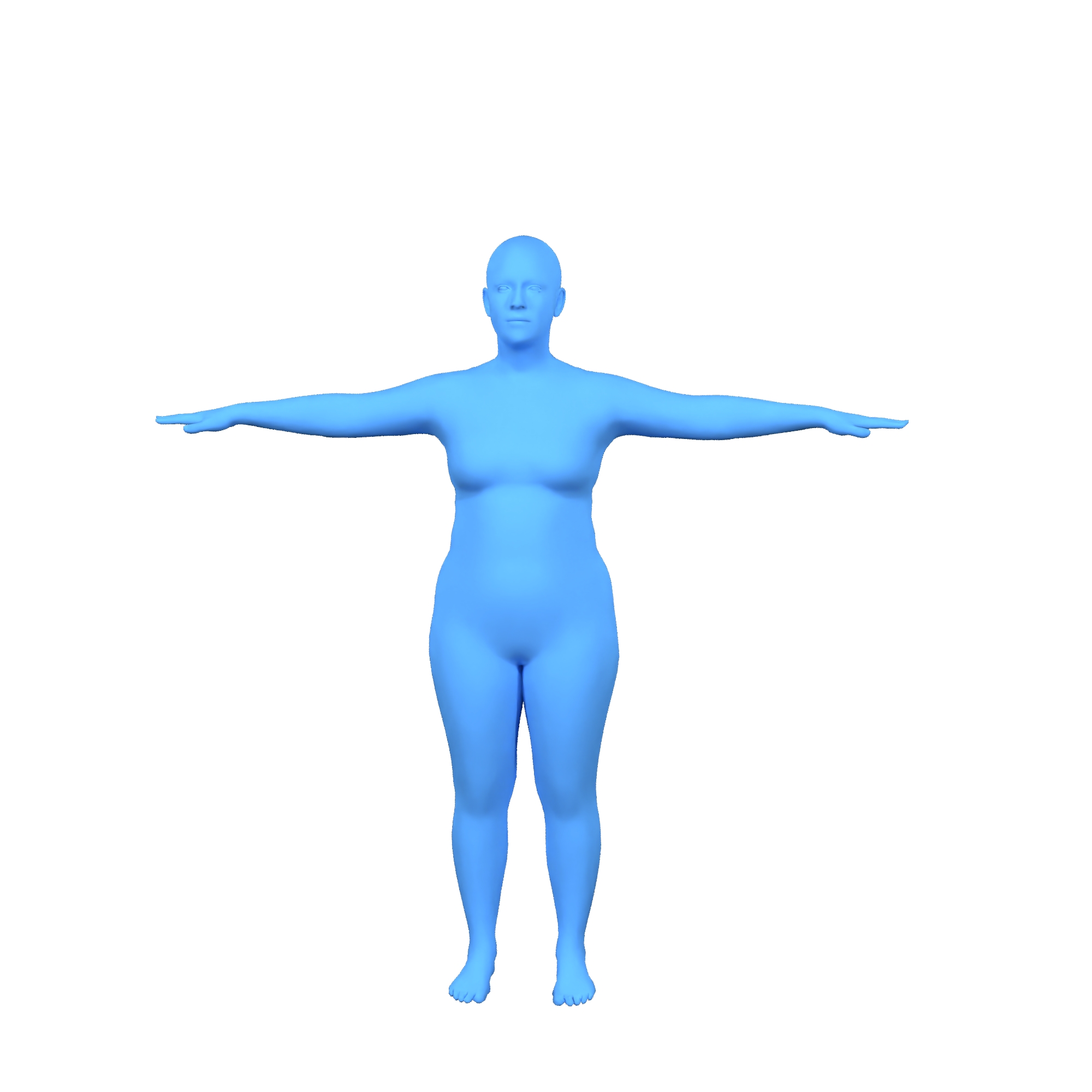}%
    \includegraphics[width=0.250\textwidth,trim={110mm, 050mm, 120mm, 130mm}, clip=true]{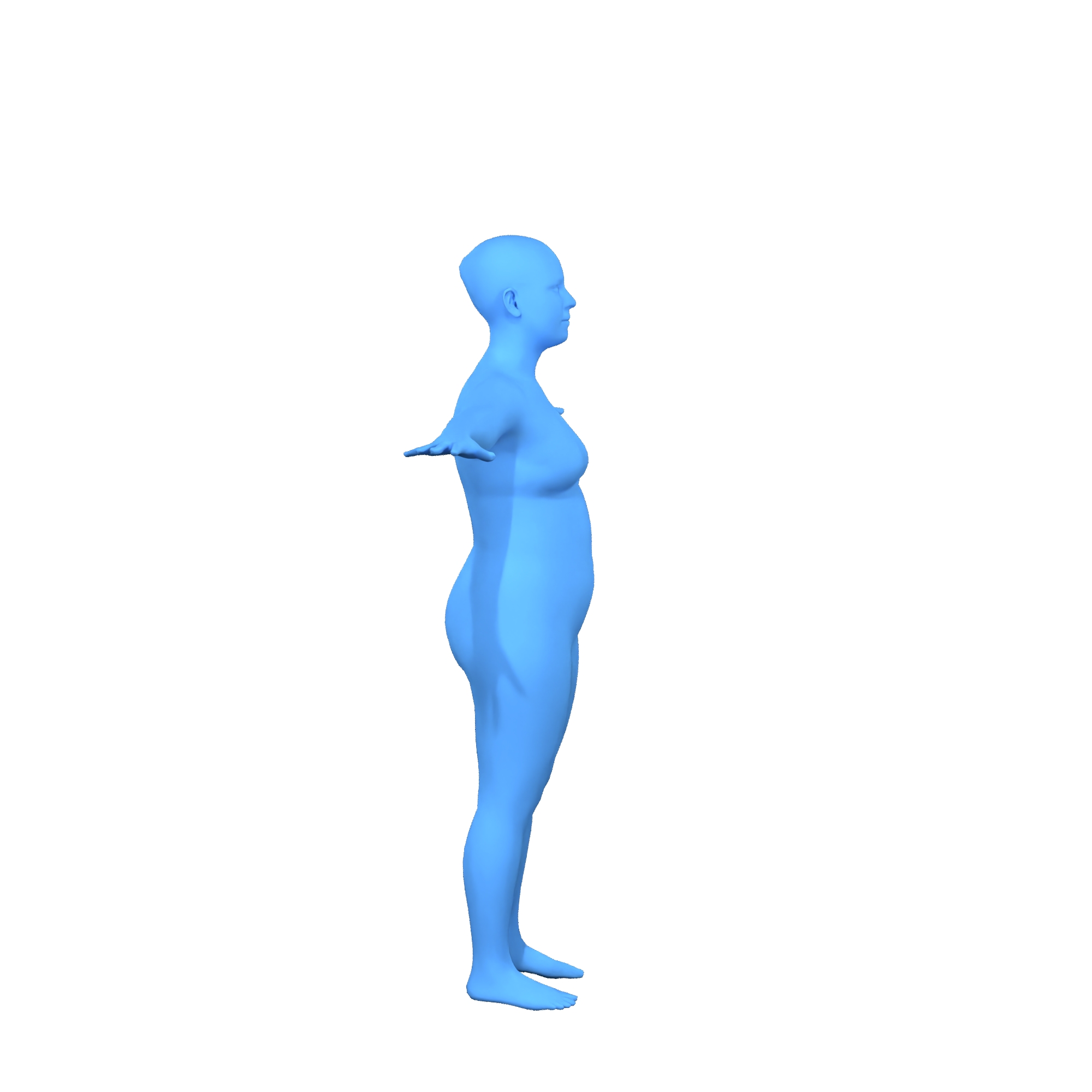}
    
    \caption{Qualitative results of \modelname predictions for female bodies. }
    
\end{figure*}

\begin{figure*}
    \centering
        \includegraphics[width=0.240\linewidth,trim={0cm, 0cm, 0cm, 0cm}, clip]{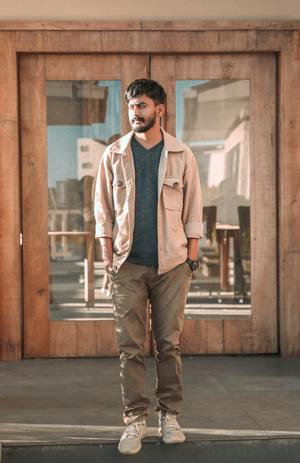}%
        \includegraphics[width=0.240\linewidth,trim={0cm, 0cm, 0cm, 0cm}, clip]{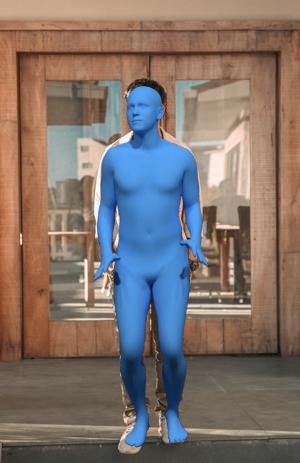}%
        \includegraphics[width=0.260\linewidth,trim={090mm, 050mm, 120mm, 110mm}, clip=true]{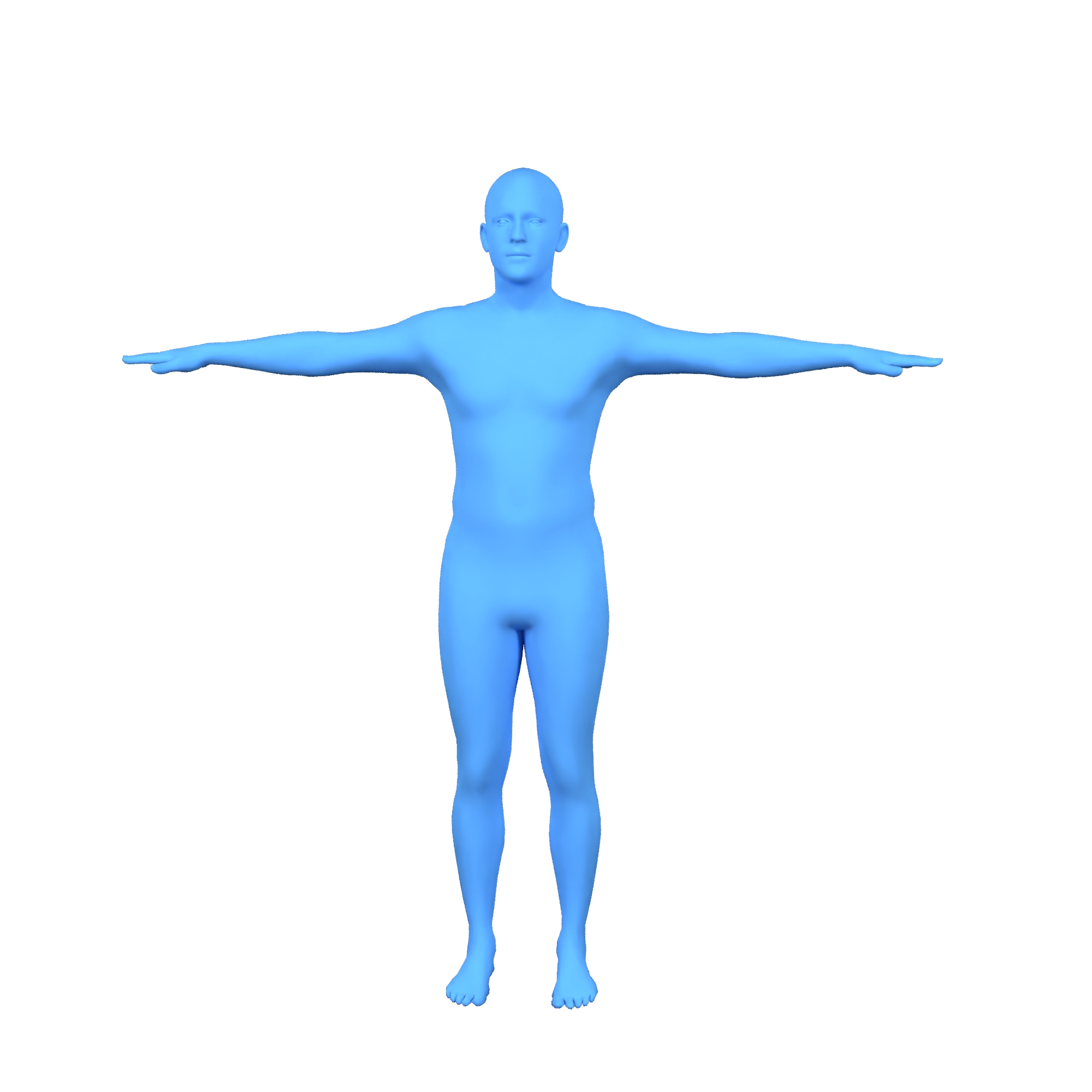}%
        \includegraphics[width=0.260\linewidth,trim={110mm, 050mm, 120mm, 110mm}, clip=true]{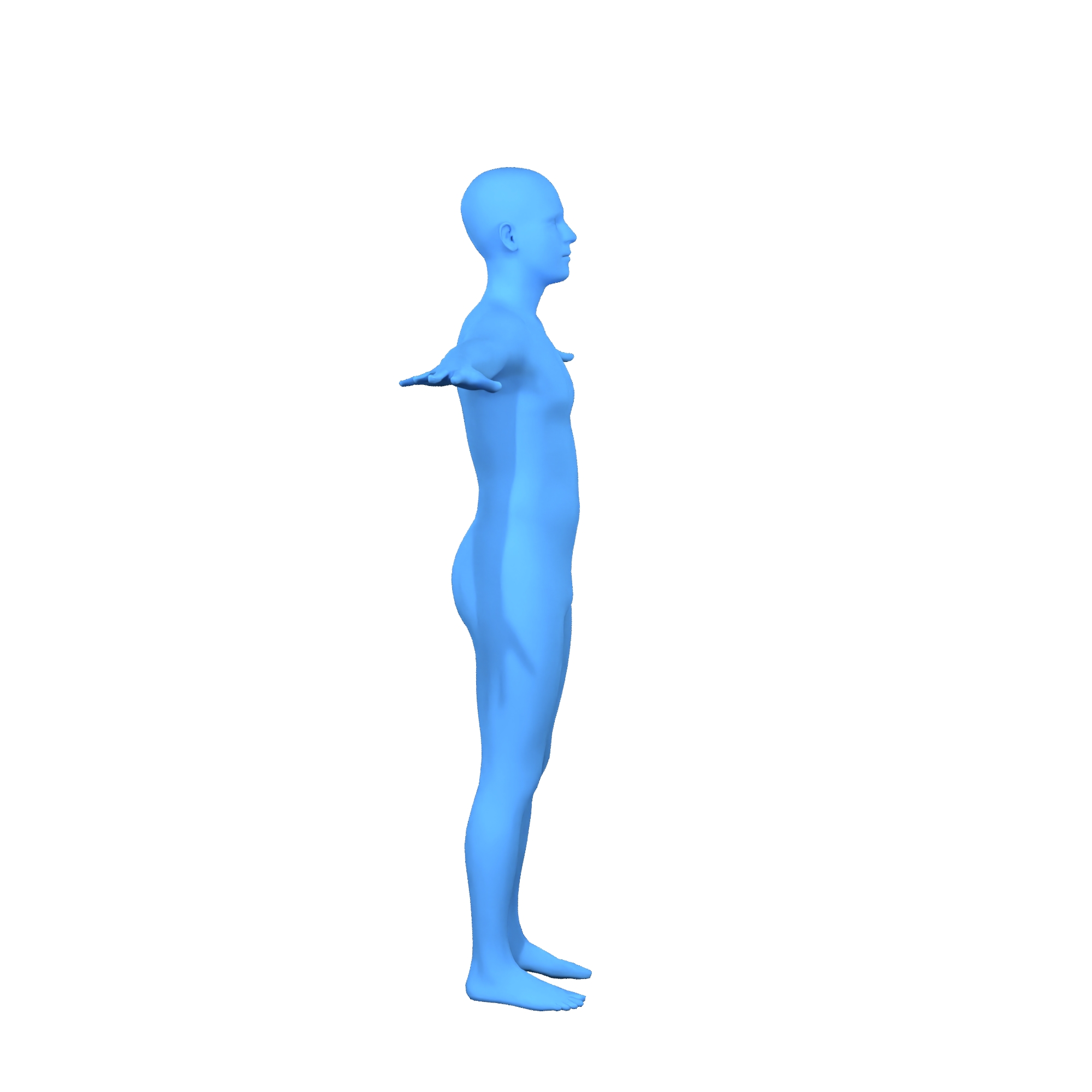}
        
        \includegraphics[width=0.240\linewidth,trim={5cm, 0cm, 0cm, 0cm}, clip]{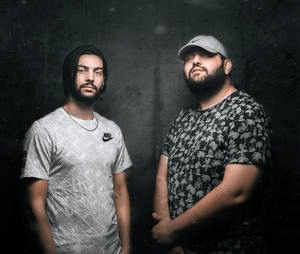}%
        \includegraphics[width=0.240\linewidth,trim={5cm, 0cm, 0cm, 0cm}, clip]{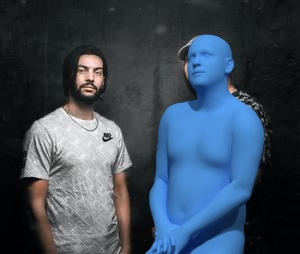}%
        \includegraphics[width=0.260\linewidth,trim={090mm, 050mm, 120mm, 110mm}, clip=true]{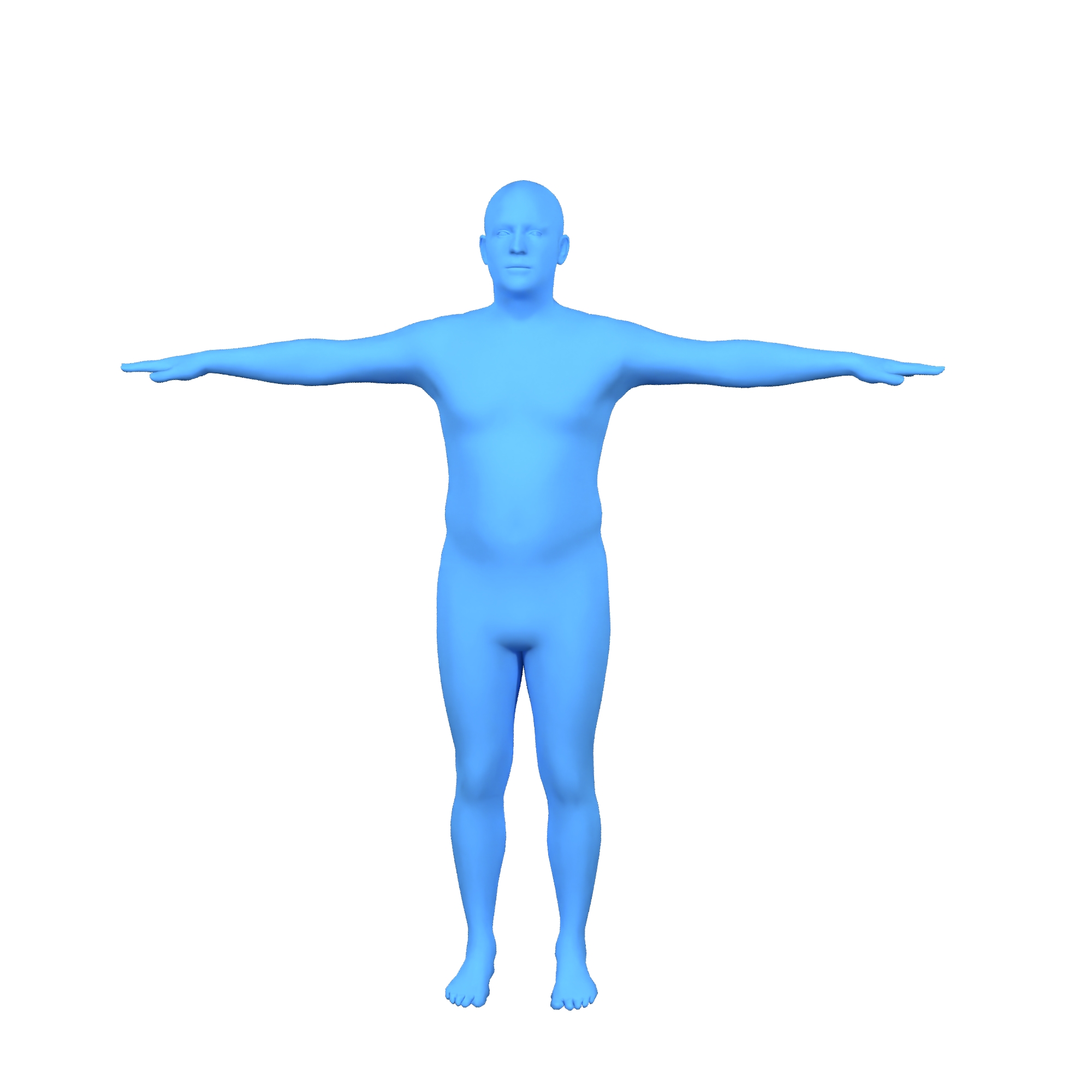}%
        \includegraphics[width=0.260\linewidth,trim={110mm, 050mm, 120mm, 110mm}, clip=true]{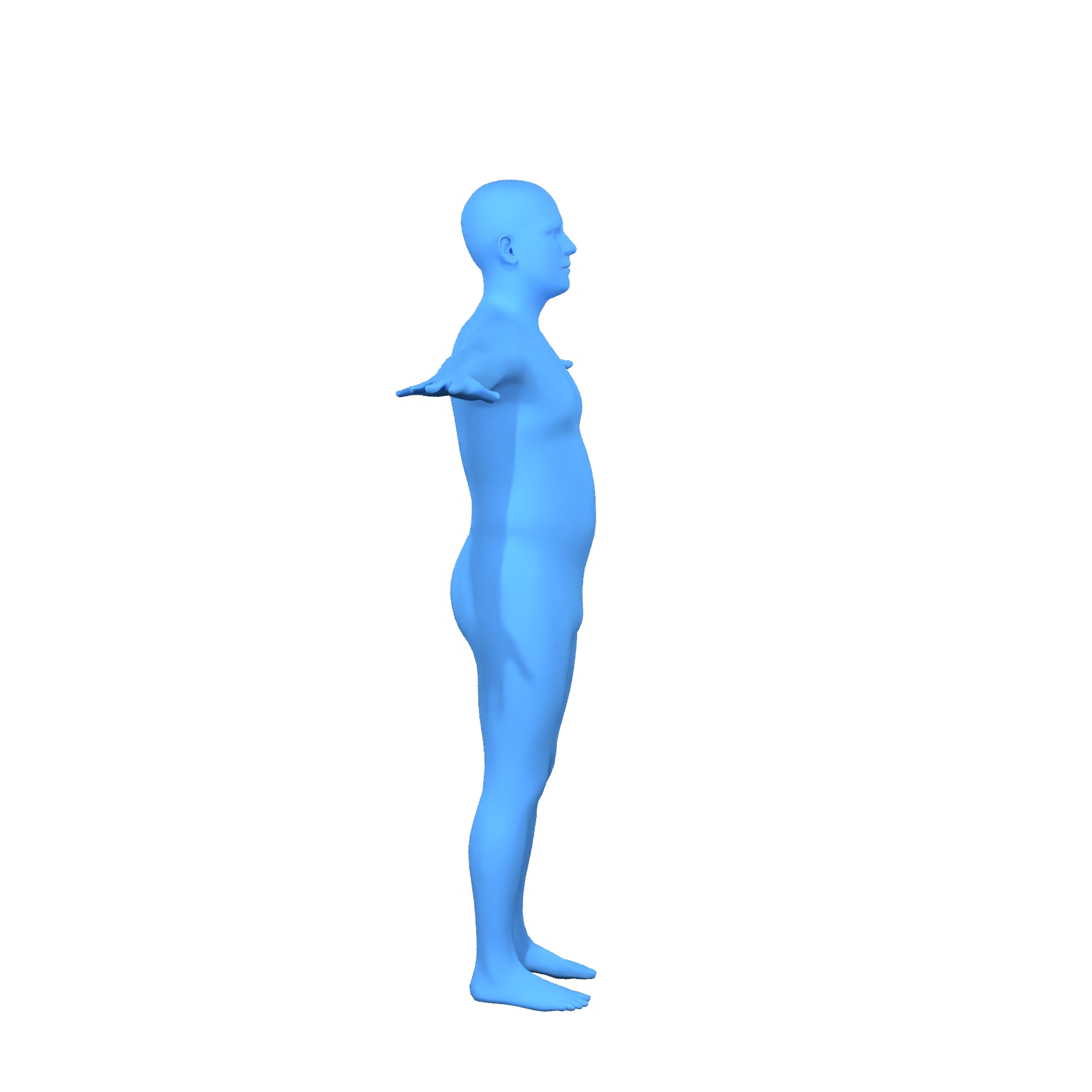}
        
        \includegraphics[width=0.240\linewidth,trim={0cm, 0cm, 0cm, 0cm}, clip]{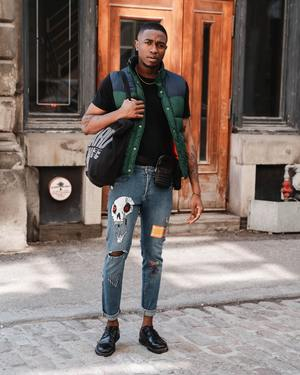}%
        \includegraphics[width=0.240\linewidth,trim={0cm, 0cm, 0cm, 0cm}, clip]{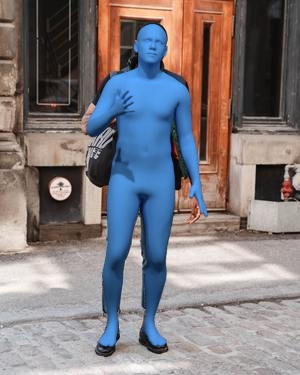}%
        \includegraphics[width=0.260\linewidth,trim={100mm, 050mm, 120mm, 110mm}, clip]{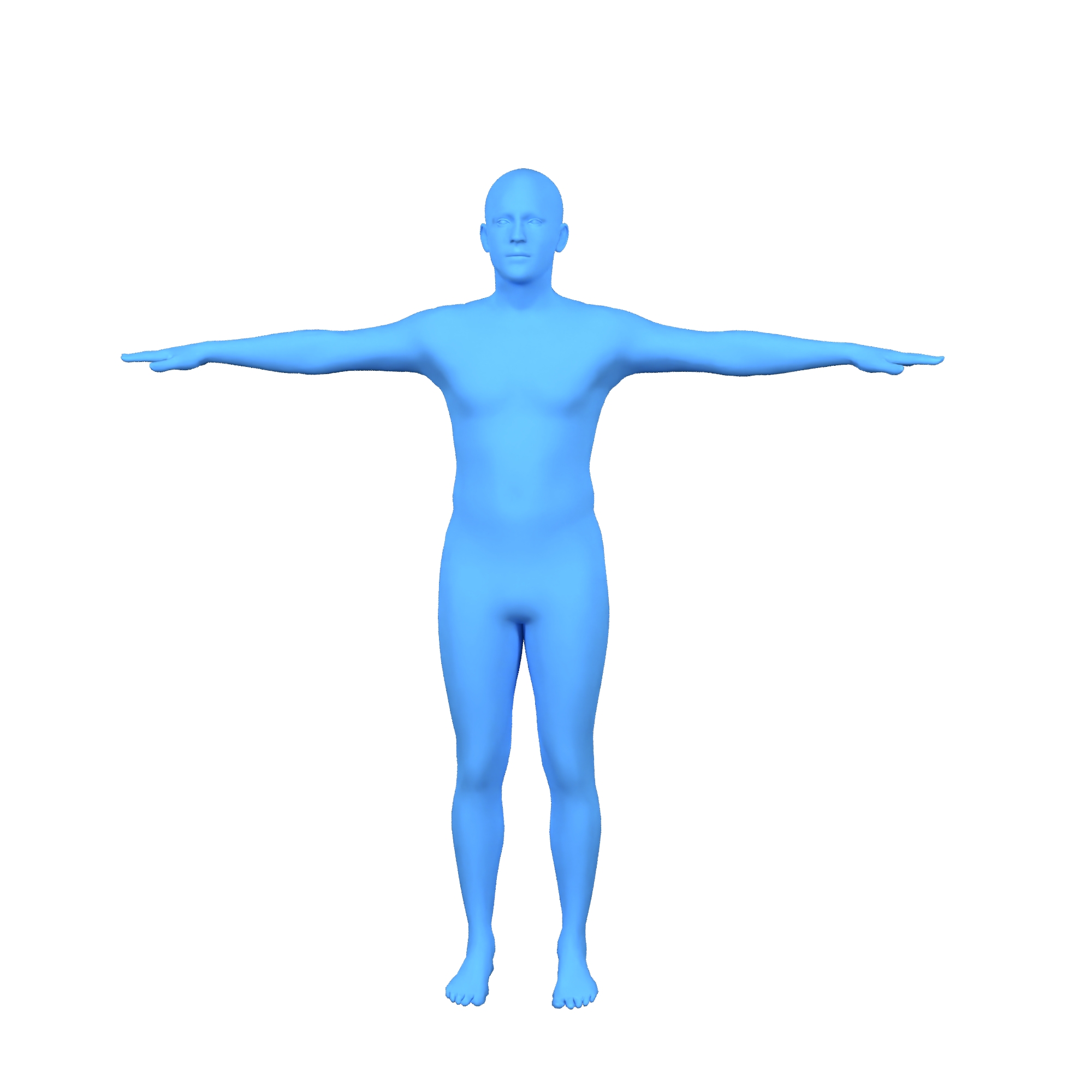}%
        \includegraphics[width=0.260\linewidth,trim={100mm, 050mm, 120mm, 110mm}, clip]{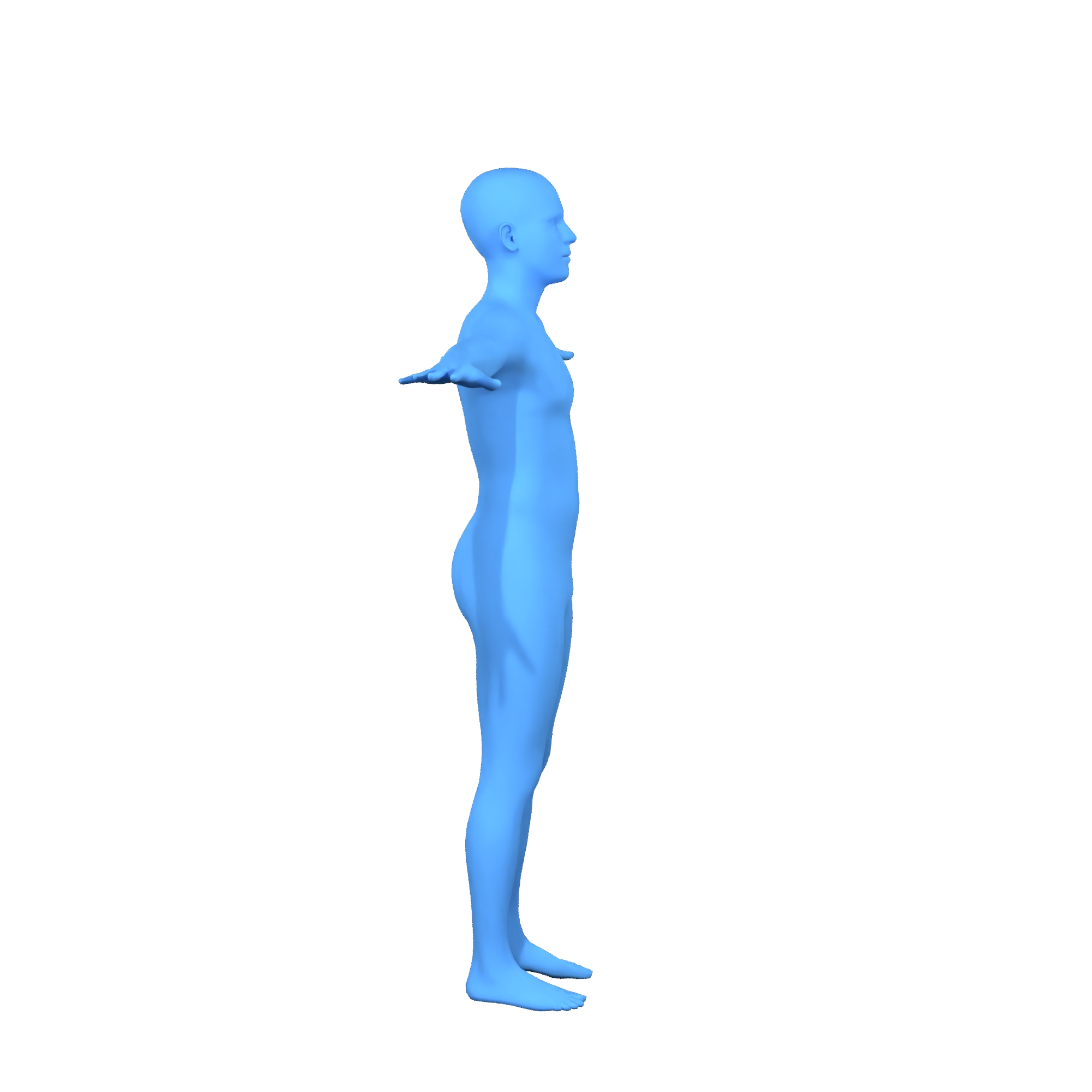}
        
    \caption{Qualitative results of \modelname predictions for male bodies.}
\end{figure*}

\end{appendices}

\end{document}